\documentclass{article}

\usepackage{microtype}
\usepackage{graphicx}
\usepackage{subfigure}
\usepackage{booktabs} 

\usepackage{hyperref}

\usepackage[accepted]{icml2023}

\usepackage{amsmath}
\usepackage{amssymb}
\usepackage{mathtools}
\usepackage{amsthm}
\DeclareMathOperator{\tr}{tr}
\DeclareMathOperator*{\argmax}{arg\,max} 
\usepackage[capitalize,noabbrev]{cleveref}

\theoremstyle{plain}

\theoremstyle{definition}

\theoremstyle{remark}

\usepackage[disable,textsize=tiny]{todonotes}

\usepackage{xspace}
\usepackage{comment}

\icmltitlerunning{Streaming Active Learning with Deep Neural Networks}

\begin{document}

\twocolumn[
\icmltitle{Streaming Active Learning with Deep Neural Networks}




\begin{icmlauthorlist}
\icmlauthor{Akanksha Saran}{msr}
\icmlauthor{Safoora Yousefi}{bing}
\icmlauthor{Akshay Krishnamurthy}{msr}
\icmlauthor{John Langford}{msr}
\icmlauthor{Jordan T. Ash}{msr}
\end{icmlauthorlist}

\icmlaffiliation{msr}{Microsoft Research NYC}
\icmlaffiliation{bing}{Microsoft Bing}

\icmlcorrespondingauthor{Akanksha Saran}{akankshasaran@utexas.edu}

\icmlkeywords{Machine Learning, Active Learning, Streaming Active Learning, Batch Active Learning}

\vskip 0.3in
]



\printAffiliationsAndNotice{}

\begin{abstract}
Active learning is perhaps most naturally posed as an online learning problem. However, prior active learning approaches with deep neural networks assume offline access to the entire dataset ahead of time. This paper proposes VeSSAL, a new algorithm for batch active learning with deep neural networks in streaming settings, which samples groups of points to query for labels at the moment they are encountered. Our approach trades off between uncertainty and diversity of queried samples to match a desired query rate without requiring any hand-tuned hyperparameters. Altogether, we expand the applicability of deep neural networks to realistic active learning scenarios, such as applications relevant to HCI and large, fractured datasets.
\end{abstract}

\section{Introduction}

Active learning considers a supervised learning situation where unlabeled data are abundant, but acquiring labels is expensive~\citep{S10, D11}. One example of this might be classifying underlying disorders from histological images, where obtaining labels involves querying medical experts. Another might be predicting drug efficacy, where labels corresponding to candidate molecules could require clinical trials or intensive computational experiments. In these settings, we typically want to carefully consider what samples to request labels for, and to obtain labels for data that are maximally useful for progressing the performance of the model.

Active learning is a classic problem in machine learning, with traditional approaches typically considering the convex and well-specified regime~\citep{S10, D11,H14}. Much recent interest in active learning has turned to the neural network case, which requires some special considerations. One such consideration is the expense associated with fitting these neural architectures — when used in conjunction with a sequentially growing training set, as one has in active learning, the model cannot be initialized from the previous round of optimization without damaging generalization performance. Instead, practitioners typically re-initialize model parameters each time new data are acquired and train the model from scratch~\citep{ash2019warm}. This structure has repositioned active learning to focus on the batch domain, where we are interested in simultaneously labeling a batch of $k$ samples to be integrated into the training set. The model is typically retrained only after the entire batch has been labeled. 

In the convex case, where a model can easily be updated to accommodate for a single sample, active learning algorithms have tended to focus on uncertainty or sensitivity. That is, a label for a given sample should be requested if the model is highly uncertain about its corresponding label, or if incorporating this sample into the training set will greatly reduce the set of plausible model weights. In contrast, a high-performing, batch-mode active learning algorithm must also consider diversity. If two samples are relatively similar to each other, it is inefficient to include them both in the batch, regardless of the model's uncertainty about their labels; having only one such sample labeled and integrated into the current hypothesis may be enough to resolve the model's uncertainty on the other.

Popular approaches for batch active learning rely on samplers that require all unlabeled data to be simultaneously available. This reliance poses several major concerns for the deployment of these algorithms. For one, the run time of these methods is conditioned on the number of unlabeled samples in a way that makes them unusable for extremely large datasets. To exacerbate the issue, it is unclear how to deploy these algorithms on modern databases, where samples might be stored in a fractured manner and cannot easily be made available in their entirety. 

It is especially unclear how to perform active learning in a streaming setting, where data are not all simultaneously available, and we do not know how many samples will be encountered. Here we might instead prefer to specify an acceptable labeling rate rather than a fixed acceptable batch size. In this streaming setup, it is further desirable to commit to a decision about whether to include an unlabeled sample in the batch as soon as it is encountered, rather than only after the stream has terminated. As a concrete example, consider an HCI application where a user interacts with the world while wearing an assistive or diagnostic device~\citep{bohus2022continual,singh2016krishnacam}. The software on the device might involve a classifier that detects objects being interacted with or the activity being performed by the user. Requesting a label to update the model to better classify these phenomena can only be done in the moment; it would be cumbersome to ask the user to provide a label corresponding to an event that occurred far in the past. How can we efficiently identify samples from a data stream for neural networks while respecting a maximum query rate?

We propose a simple active learning algorithm, Volume Sampling for Streaming Active Learning (VeSSAL)\footnote{Code for the implementation of VeSSAL can be found at \url{https://github.com/asaran/VeSSAL.git}}, that addresses the concerns mentioned above. VeSSAL is made to accommodate the streaming setting, and as such it only needs to see each unlabeled point once in order to arrive at a decision about whether it should be labeled. This makes VeSSAL attractive even for fixed datasets that might be extremely large or fractured, as these are often interacted with using streaming, distributed database frameworks.  VeSSAL is a natural choice for ``committal'' situations, when labeling decisions need to be made on the fly. On non-sequential datasets, where more conventional active learning algorithms could be exercised, VeSSAL can be significantly faster, especially for large batch sizes. 

Despite its simplicity and flexibility, VeSSAL is surprisingly high performing. We show that VeSSAL produces models with predictive capabilities on par with state-of-the-art approaches, even though they are not restricted to this streaming, committal setting. We further demonstrate this to be the case in adversarial situations, where VeSSAL is presented with data that have been sorted to induce domain drift. VeSSAL is hyperparameter free, making it a powerful candidate for a wide range of active learning scenarios.

The paper proceeds as follows. We overview related work in Section~\ref{sec:related}. In Section~\ref{sec:algo}, we present the mathematical formulation 
for the streaming active learning setting,  
along with details of our proposed algorithm. 
We give empirical support for our proposed approach in Section~\ref{sec:experiments} via experiments on three benchmark datasets and one real-world dataset, in conjunction with two neural network architectures and three different batch sizes. We conclude with a discussion in Section~\ref{sec:discussion}.
\vspace{-0.2cm}
\section{Related Work}
\label{sec:related}
\vspace{-0.1cm}
This section situates VeSSAL with respect to prior work on streaming active learning (Sec.~\ref{sec:related_streaming}) as well as batch active learning strategies for training neural networks (Sec.~\ref{sec:related_deep}).

\begin{figure}
    \includegraphics[width=.48\textwidth]{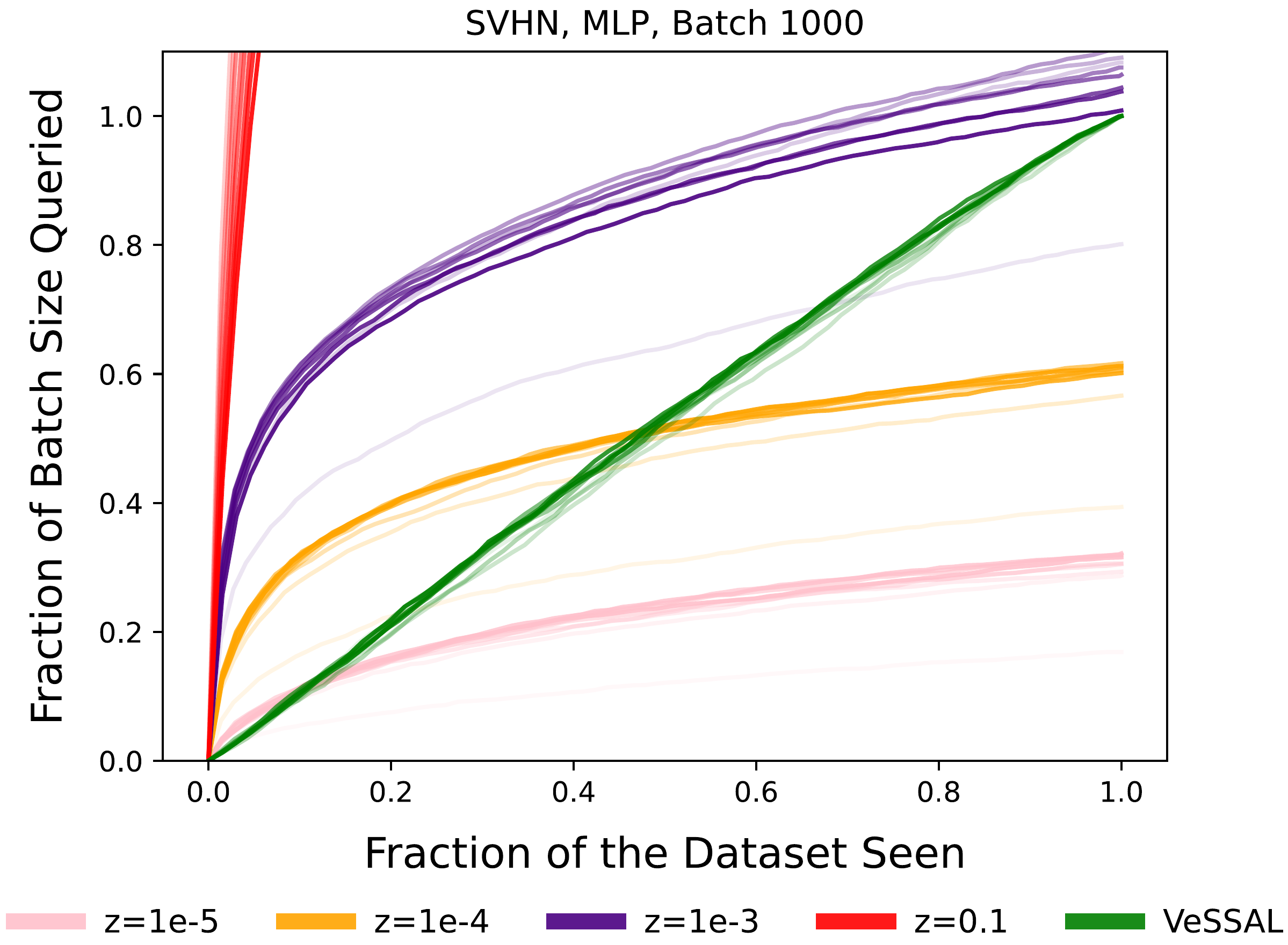}
    \caption{A comparison in terms of sampling rate for our proposed tuning approach and choosing fixed $z_t$ values, used to scale the probability mass on a candidate point as $p_t = z_t \cdot g(x_t)^\top \Sigma^{-1} g(x_t)$. We plot the fraction of the batch size that has been selected as a function of the amount of data in the stream that has been encountered. Fixed scaling values can drastically undersample or oversample, and distribute the labeling budget inequitably across the stream.   
    The active learning round is denoted by line opacity, with darker colors corresponding to higher round numbers — here we show the first ten rounds for each strategy.
    \label{fig:sampling_rate}}
    
\end{figure}

\subsection{Streaming Active Learning}
\label{sec:related_streaming}

Active learning has enjoyed many successes for problems in the convex learning setting~\citep{hanneke2014theory, BDL09, beygelzimer2010agnostic, roth2006margin,beygelzimer2010agnostic,huang2015efficient, hsu2010algorithms, hanneke2015minimax}. 
\citet{BDL09} propose a statistically consistent method using importance weighting for actively learning binary classifiers under general loss functions. 
~\mbox{\citet{krishnamurthy2017active}} present a version-space based active learning method with performance guarantees for cost-sensitive multiclass classification. 
Similar to these methods, there is a long line of theoretical work on active learning for linear models~\citep{hanneke2014theory}. While these approaches are indeed designed for the streaming setting, they rely on updating the linear hypothesis after each sample is labeled, precluding them from being used in conjunction with deep neural networks, where updating the model is known to be extremely expensive. Furthermore, \citet{sun2022information} consider the problem of filling a replay buffer for use in continual learning. Unlike in active learning, the model has access to a ground-truth label for each sample it encounters. \looseness=-1

Successful streaming-based techniques with theoretical guarantees have been developed for the related setting of adaptive sampling and low-rank matrix approximation \citep{frieze2004fast,deshpande2006adaptive, deshpande2006matrix, ghashami2014relative, bhaskara2019residual}. These methods find utility in several problem domains such as online PCA~\citep{boutsidis2014online, bhaskara2019residual}, online column subset selection~\citep{bhaskara2019residual}, and online k-means clustering~\citep{braverman2011streaming}. We take inspiration from this line of prior theoretical work to design a streaming algorithm for neural batch active learning.\looseness=-1

\subsection{Batch Active Learning for Deep Neural Networks}
\label{sec:related_deep}

In recent years, several advancements have been made in the area of pool-based active learning for deep neural networks~\citep{ren2021survey}. Prior approaches have either employed diversity-based sampling ~\citep{sener2018active, geifman2017deep, gissin2019discriminative}, uncertainty-based sampling~\citep{gal2017deep, ducoffe2018adversarial, beluch2018power} or both~\citep{ash2019deep, ash2021gone}.
~\citet{ash2019deep} propose a pool-based deep active learning method which leverages gradient embeddings to capture both diversity and uncertainty by pairing a gradient-based representation with an approximate k-DPP sampling technique. 
 \citet{gudovskiy2020deep} give an active learning approach to tackle the distribution shift between the train and test data via a feature density matching method using Fisher kernels.
Still, these approaches are not designed to handle the streaming setting. Instead, these samplers require access to the entire candidate pool in order to identify each valuable point to include in the batch of unlabeled samples. 

More recently, \citet{ban2022improved} proposed a stream-based deep active learning approach, albeit not a batch active learning algorithm. Instead, they query only a single point at a time, and add it to their replay buffer after its corresponding label has been obtained. Because this work is outside of the batch setting, there are no diversity considerations to the label acquisition rule. Instead, decisions are only based on predictive uncertainty, as measured by the difference in probability mass between the most likely and second most likely predicted label. \citet{lavania2021practical} propose a streaming submodular maximization-based active learning approach, but the queies are sampled in a non-committal fashion (unlike VeSSAL). 

Several works have designed active learning approaches specifically for image classification~\citep{kovashka2016crowdsourcing} and object detection~\citep{choi2021active,brust2018active,senzaki2021active}. \citet{sun2019active} leverage deep reinforcement learning to train a data selection policy for training neural networks to perform image classification. 
\citet{roy2018deep} use an uncertainty based sampling approach for active learning via the paradigm of ``query by committee'', leveraging the disagreement between convolutional layers for Single Shot Multibox Detector architectures \citep{liu2016ssd}. 
Still, these approaches assume access to the entire dataset for decision making, and cannot be used in the streaming setting.

\citet{brust2018active} use uncertainty-based margin sampling for streaming active learning with object detectors. They present various methods to aggregate uncertainty estimates for all objects in an image to determine their selection strategy. While their approach is designed for continual object learning settings, it is limited to the problem of object detection and cannot be applied to classification tasks.\looseness=-1

\begin{algorithm}[t]
\begin{algorithmic}[1]
\REQUIRE Neural network $f(x;\theta)$, unlabeled stream of samples $U$, ideal sampling rate $q$
\STATE Initialize $t = 1$\\
\STATE Initialize $\hat{\Sigma}^{-1}_0 = \lambda^{-1} I_d$  \COMMENT{regularized by $\lambda$ for stability}\\
\STATE Initialize $A_0 = 0_{d,d}$ \COMMENT{covariance over all data}\\
\STATE Initialize $B = \emptyset$ \COMMENT{set of chosen samples}\\
\FOR{$x_t \in U$:}
\STATE $A_t \gets \frac{t-1}{t}A_{t-1} + \frac{1}{t}g(x_t) g(x_t)^\top$\\
\STATE $p_t = q \cdot g(x_t)^\top \hat{\Sigma}^{-1}_{t} g(x_t) \tr ({\hat{\Sigma}^{-1}_{t} A_{t}})^{-1}$\\
\STATE \textbf{with} probability $\min(p_t, 1)$:\\
\STATE \ \ \ \ Query label $y_t$ for sample $x_t$
\STATE \ \ \ \ $B\gets B \cup (x_t, y_t)$\\

\STATE \ \ \ \ \ $\hat{\Sigma}_{t+1}^{-1} \gets \hat{\Sigma}_{t}^{-1} - \frac{\hat{\Sigma}_{t}^{-1}g(x_t)g(x_t)^\top \hat{\Sigma}_{t}^{-1}}{1 + g(x_t)^\top \hat{\Sigma}^{-1}_{t}g(x_t)}$ \COMMENT{rank-1 Woodbury update}\\

\STATE \textbf{else}: 
\STATE \ \ \ \ $\hat{\Sigma}_{t+1}^{-1} \gets \hat{\Sigma}_{t}^{-1}$
\STATE $t \gets t + 1$\\
\STATE \textbf{return} labeled batch $B$ for retraining $f$
\ENDFOR
\end{algorithmic}
\caption{Volume sampling for streaming active learning (VeSSAL)}
\label{alg:main}
\end{algorithm}

\section{Volume Sampling for Streaming Active Learning (VeSSAL)}
\label{sec:algo} 

Neural active learning algorithms that incorporate diversity can largely be thought of as making two design decisions. The first decision is how unlabeled candidate samples should be represented. Common choices include using the penultimate layer representation of the current state of the network~\citep{sener2018active} or using a hypothetical gradient that might be induced by a given sample~\citep{ash2019deep}. In either case, once data are in this space, the second decision is regarding how unlabeled points should be selected in order to encourage batch diversity. 

In VeSSAL, these decisions are made to produce a high-performing active learner that is amenable to the streaming, committal setting. Specifically, we assume that each candidate $x_t$ is seen only once, and that we must make a decision about whether or not to include it in the batch $B$ as soon as it is encountered. Once the labeling budget $k$ is allocated, we retrain the model and repeat the process.

VeSSAL performs approximate volume sampling over unlabeled candidate points in a gradient space computed with respect to the last layer of the neural network.
For a neural network $f$ with parameters $\theta$, last-layer parameters $\theta_L \in \theta$  and cross-entropy loss function $\ell$, the gradient representation for a sample $x_t$ is
\begin{equation}
g(x_t) = \frac{\partial}{\partial \theta_L} \ell(f(x_t;\theta), \hat y_t),
\label{eq:one}
\end{equation}
where $\hat{y}$ denotes the most likely label according to the current state of the model, i.e. $\hat{y_t} = \argmax f(x_t;\theta)$.

A typical way of doing volume sampling is to select a batch of points with probability proportional to the determinant of their gram or covariance matrix~\citep{kulesza2012determinantal}. In the latter case, this would mean the probability mass on a batch of samples $B$ is proportional to
\begin{equation}
\det \Big(\sum_{x \in B} g(x) g(x)^\top \Big),
\vspace{-3.5mm}
\end{equation}

where $|B|=k$, the pre-specified labeling budget.

There are several reasons to favor the covariance matrix version over the gram matrix. From a theoretical point of view, when used in an outer product, $g(x)$ could be thought of as a rank-1 approximation of the Fisher information matrix, $I(x;\theta) := \mathbb{E}_{y \sim p_{\theta}(\cdot \mid x,\theta)} \nabla^2 \ell(x,y;\theta)$~\citep{ash2021gone}. As such, this construction is reminiscent of a classic goal in active learning, which is to maximize the determinant for the Fisher~\citep{mackay1992information}. This objective is attractive because, in the realizable setting, it selects samples that maximize the information gained by model parameters after labeling. 

From a more practical point of view, the covariance matrix will stay fixed in dimensionality even as the batch size $k$ is changed. In the gram matrix alternative, which is suggested in \citet{ash2019deep}, the size of the matrix grows with $k$, potentially becoming intractable for larger batch sizes.

A process that samples from a distribution characterized by a determinant like this is referred to as a determinantal point process (DPP). Sampling from a DPP is usually done via Markov Chain Monte Carlo, and making this procedure efficient is an active area of research with wide-ranging statistical applications~\citep{bardenet2017few}. Still, the mixing times associated with these algorithms generally makes them too inefficient to be used in conjunction with modern active learning algorithms. For example, BADGE~\citep{ash2019deep} suggests using kmeans++ as a surrogate for DPP sampling, and Coreset~\citep{sener2018active} uses a furthest-first traversal approach (though not in gradient space). \looseness=-1

These sampling approaches are workable surrogates for true volume sampling, but they require all data to be simultaneously accessible. Our algorithm demonstrates that this is not necessary and that near state-of-the-art performance can be obtained by a sampler that (1) sees samples in a streaming fashion, such that each point is only observed once and that (2) commits to a labeling decision as soon as a sample is encountered.

\begin{figure*}[!t]
    \centering
    \includegraphics[width=1\textwidth]{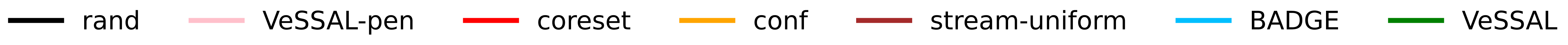}
    \includegraphics[width=0.32\textwidth]{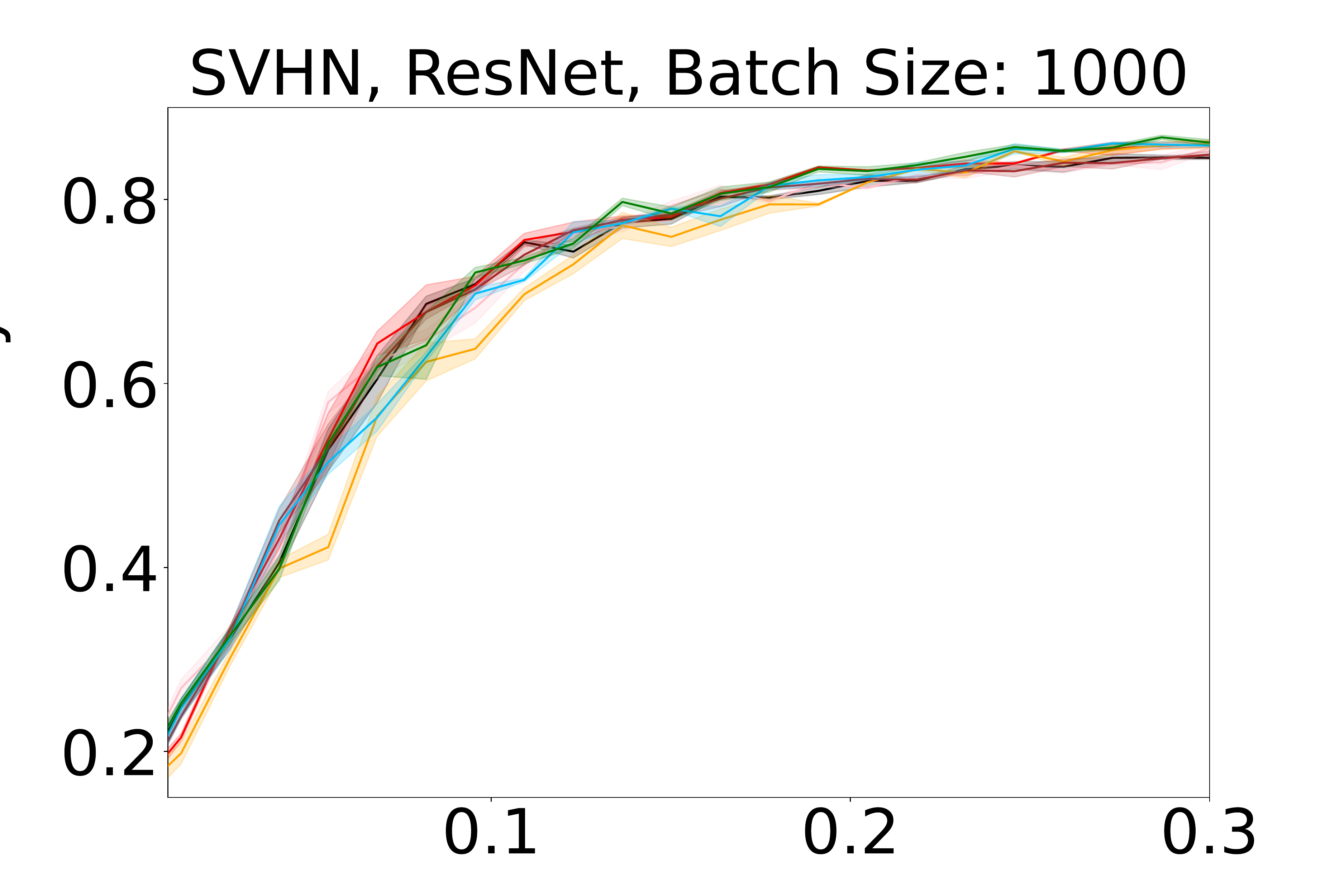}
    \includegraphics[width=0.32\textwidth]{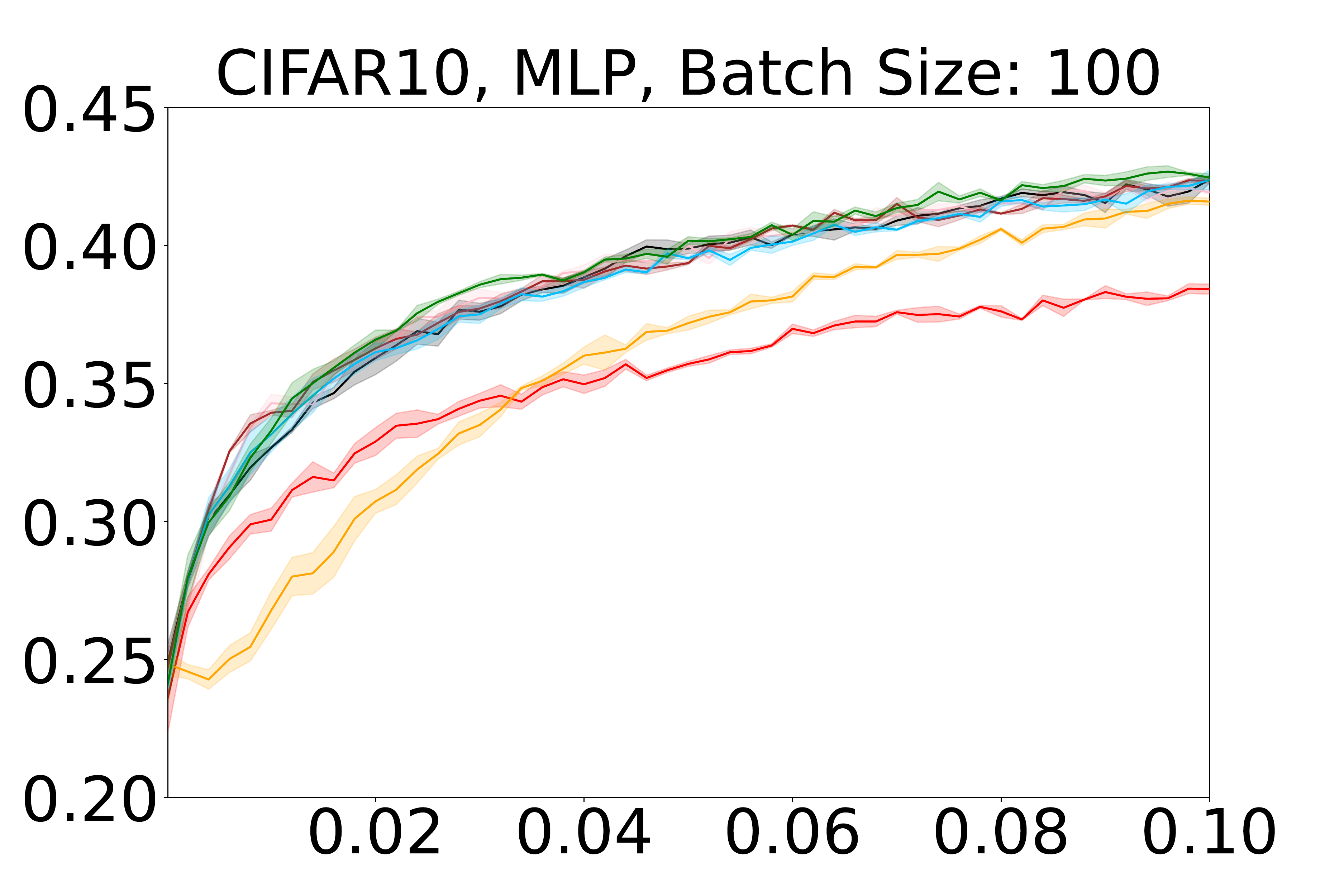}
    \includegraphics[width=0.32\textwidth]{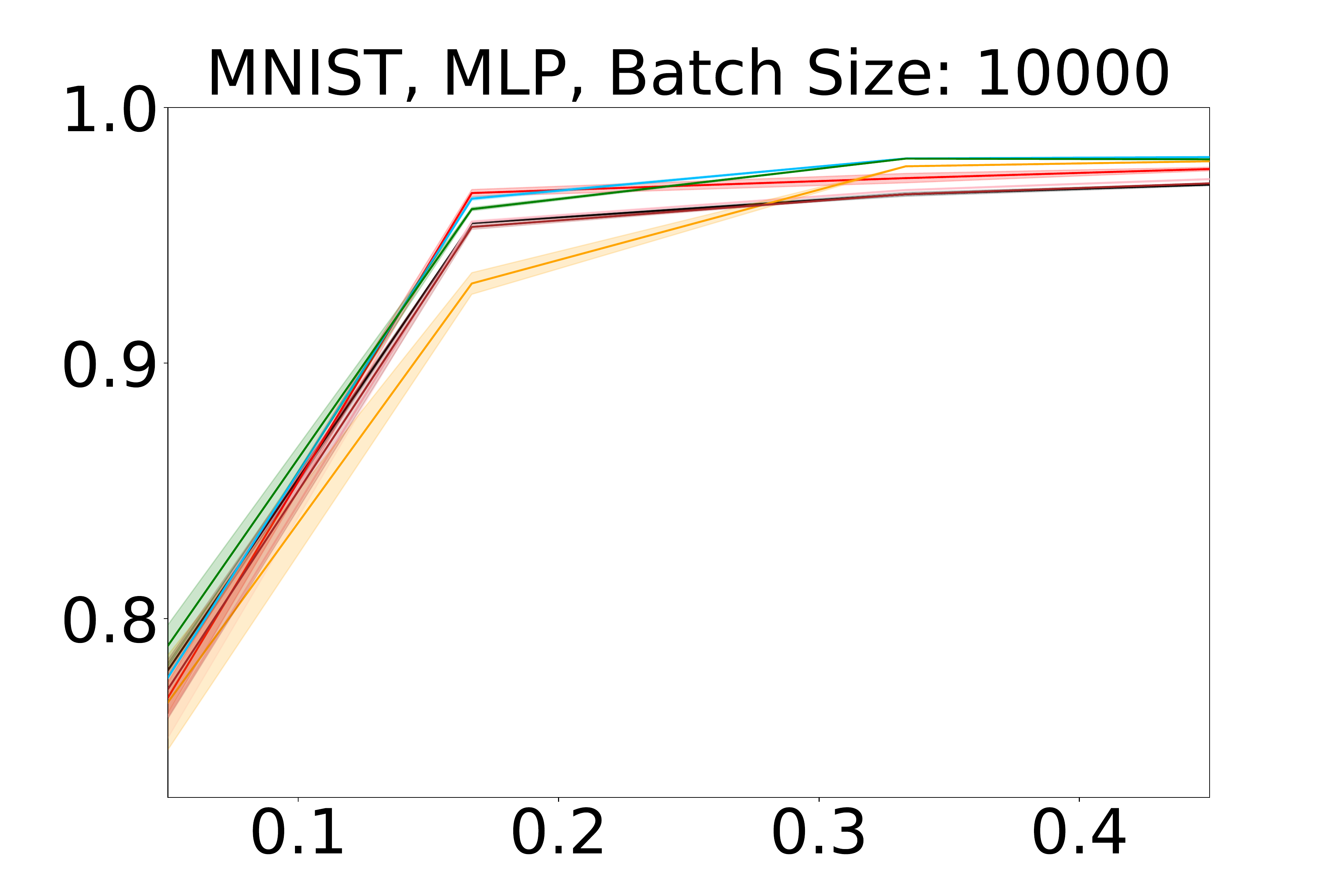}
    \caption{Learning curves for different neural active learning methods tested with i.i.d data streams. Two network architectures, three batch sizes, and three datasets are shown here. These plots have been zoomed to highlight discriminative regions, but complete results are shown in the Appendix and are aggregated in Figure~\ref{fig:heatmap_unordered} (a).}
    \label{fig:learning_curves_main}
\end{figure*}

VeSSAL choses a sample $x_t$ for labeling with probability 
 $p_t$ proportional to the determinantal contribution of its gradient when considering other items that have already been chosen,
\vspace{-1mm}
\begin{align*}
p_t \propto \det \Big(\hat{\Sigma}_t + g(x_t) g(x_t)^\top \Big) 
\end{align*}

Using the matrix determinant lemma \cite{greub2012linear, strang2006linear}, the expression for $p_t$ reduces to
\begin{align*}
p_t &\propto \det(\hat{\Sigma}_t)(1 + g(x_t)^\top \hat{\Sigma}_t^{-1} g(x_t))  \\ 
&\propto g(x_t)^\top \hat{\Sigma}_t^{-1} g(x_t).
\end{align*}

Here $\hat{\Sigma}_t$ is the covariance over samples that have already been selected for inclusion in the batch,  $\sum_{x \in B} g(x) g(x)^\top$.

To compute a $p_t$ in practice, $g(x_t)^\top \hat{\Sigma}_t^{-1} g(x_t)$ must be scaled by some value $z_t$, which reflects the labeling budget available to the algorithm. Because the amount of data in the stream might not be known, we consider tuning $z_t$ to reflect a desired labeling frequency $q$. In HCI applications~\citep{bohus2022continual, singh2016krishnacam, wang2021wanderlust}, for example, we might not know how long a user will interact with a device (the total number of candidate samples), but we might instead have some sense of an acceptable frequency with which labels can be queried. If we instead do know the total number of samples, then we could consider $q$ to be the ratio between the labeling budget and the size of the candidate set. Specifically, we desire to find some scalar $z_t$ such that 
\begin{equation}
\mathbb{E}_{x} [p_t] = \mathbb{E}_{x} \Big[ z_t \cdot g(x)^\top \hat{\Sigma}_t^{-1} g(x) \Big] = q.
\label{eq:three}
\end{equation}

How should this $z_t$ be chosen? Because the statistics of gradient representations $g(x)$ vary with the state of $f$, one fixed value is unlikely to work well across model architectures, datasets, batch sizes, and rounds of data selection. Instead, we aim to find an adaptive strategy, such that we both obtain the desired sampling frequency $q$ and that we do so in a way that does not allocate probability mass disproportionately across temporal regions of the stream.

One option for adaptively adjusting $z_t$ as selection progresses might be a multiplicative weights approach, where we select a scaling parameter from a distribution over a number of $z_t$ values, and constantly update this distribution to reflect whatever choice is giving us the best rate. Another option is to use gradient descent, making $z_t$ larger or smaller at each step in the service of minimizing error between the current and target sampling rate $q$. Unfortunately, these approaches are unlikely to work well, because the underlying distribution given by the determinantal contribution changes with each selected point. Specifically, every time a sample is added to the batch, and $\hat{\Sigma}_t$ is updated, the distribution of responses $g(x_t)^\top \hat{\Sigma}_t^{-1} g(x_t)$ can change drastically. This domain shift precludes adaptive solutions from efficiently finding a suitable $z_t$ like those mentioned above, because they assume the underlying distribution is stationary. 

To circumvent this, we simply rewrite the expectation to disentangle $\hat{\Sigma}^{-1}_t$ from statistics relating to $g(x)$:
\begin{align*}
\mathbb{E}_{x} \Big[ z_t \cdot g(x)^\top \hat{\Sigma}_t^{-1} g(x) \Big] &= z_t \cdot  \mathbb{E}_{x} \Big[ \tr \Big( g(x)^\top \hat{\Sigma}_t^{-1} g(x) \Big) \Big] \\
&= z_t \cdot  \mathbb{E}_{x} \Big[ \tr \Big(\hat{\Sigma}_t^{-1} g(x) g(x)^\top \Big) \Big] \\
&= z_t \cdot  \tr \Big(\hat{\Sigma}_t^{-1} \mathbb{E}_{x} \Big[ g(x) g(x)^\top \Big] \Big).
\end{align*}
Here, our ability to find a suitable $z_t$ relies only on our ability to estimate the covariance of $g(x)$, and is not affected by the frequently changing $\hat{\Sigma}_t^{-1}$. If we approximate $\mathbb{E}_{x} \Big[ g(x) g(x)^\top \Big]$ as $\frac{1}{t}\sum_{i=1}^t g(x_i) g(x_i)^\top$, this immediately suggests a way to compute $p_t$ that is amenable to the streaming setting:
\begin{equation}
p_t = \frac{q \cdot g(x_t)^\top \hat{\Sigma}_t^{-1} g(x_t)}{\tr \Big( \frac{1}{t} \hat{\Sigma}_t^{-1} \sum_{i=1}^t g(x_i) g(x_i)^\top  \Big)}.
\end{equation}

Empirically, we find that this auto-tuning of the probability mass on each sample to be far more effective than using a fixed value for $z_t$. In Figure~\ref{fig:sampling_rate}, we demonstrate that this approach not only consistently matches the desired labeling frequency $q$, but it also distributes our labeling budget equitably across the data stream. This is evident from the identity line between proportion of data seen and budget consumed. In contrast, fixed values of $z_t$ can drastically oversample or undersample, by a degree that varies with each round of active learning. Further, because of the nature of the determinant, these fixed-$z_t$ versions often sample far more aggressively in the beginning of the stream than the end.

The complete VeSSAL approach is presented as Algorithm~\ref{alg:main}. In it, the estimated covariance over all samples is denoted as $A$, which is initialized to all zeros. We increment $\hat{\Sigma}$ efficiently using a Woodbury update on each chosen sample~\citep{woodbury1950inverting}.

One interesting note is that our estimate for $z_t$ is only as good as our estimate of $\mathbb{E}_{x} \big[ g(x) g(x)^\top \big]$, which we obtain as a simple average of outer products of the $g(x_t)$ vectors observed in the stream. A consequence of this is that the estimate will be biased if data are ordered in the stream in a non-I.I.D. fashion. In the following section, we empirically demonstrate that this appears to not be an issue --- VeSSAL performs on-par with state-of-the-art, non-streaming algorithms regardless of how data are ordered.
\section{Experiments}
\label{sec:experiments}

\begin{figure*}
    \centering   
    \hspace{-0.55cm}
    \subfigure[I.I.D. Data Stream]{
    \includegraphics[scale=0.72]{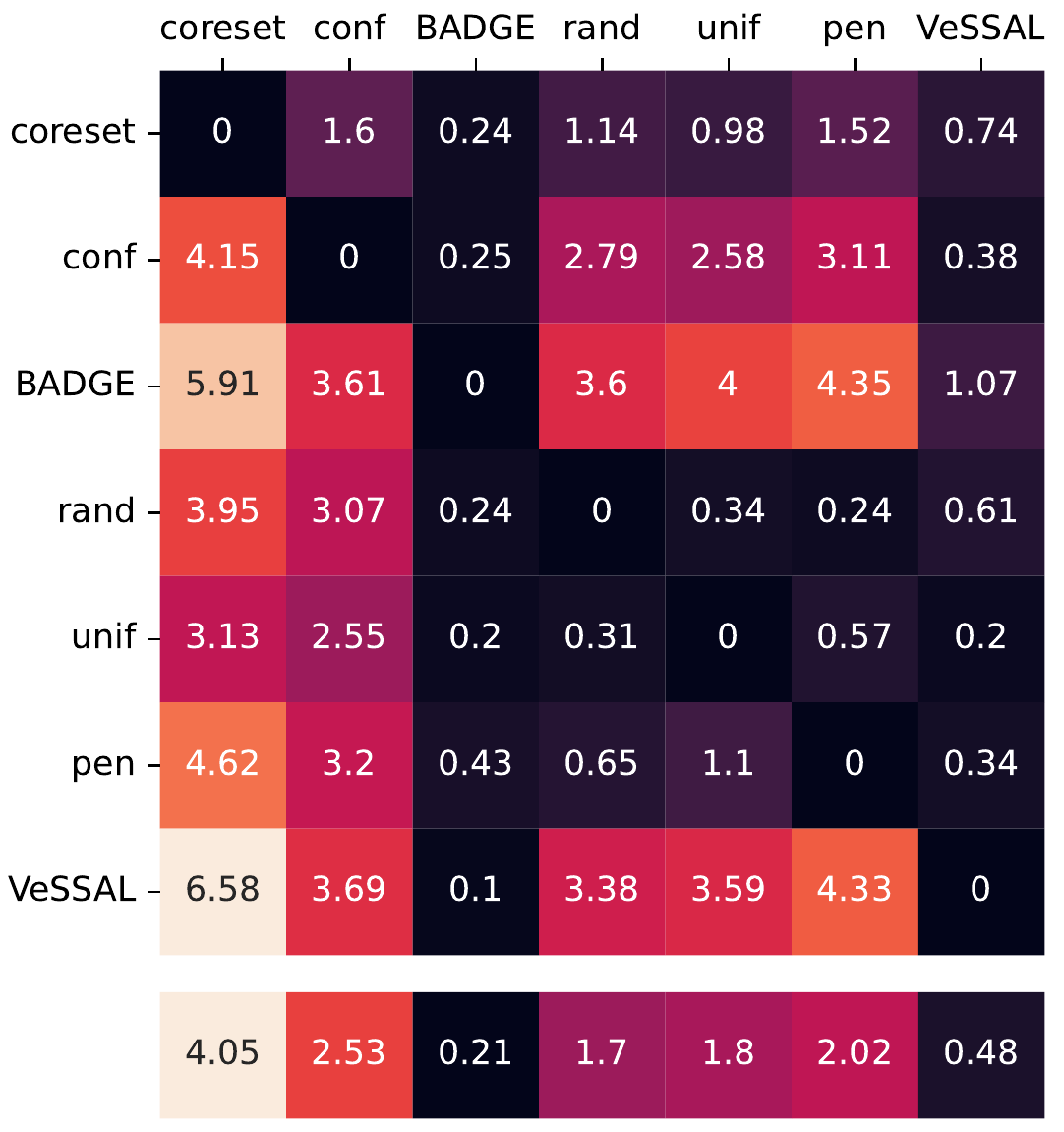}
    }
    \hspace{1cm}
    \subfigure[Non-I.I.D. Data Stream]{
     \includegraphics[scale=0.72]{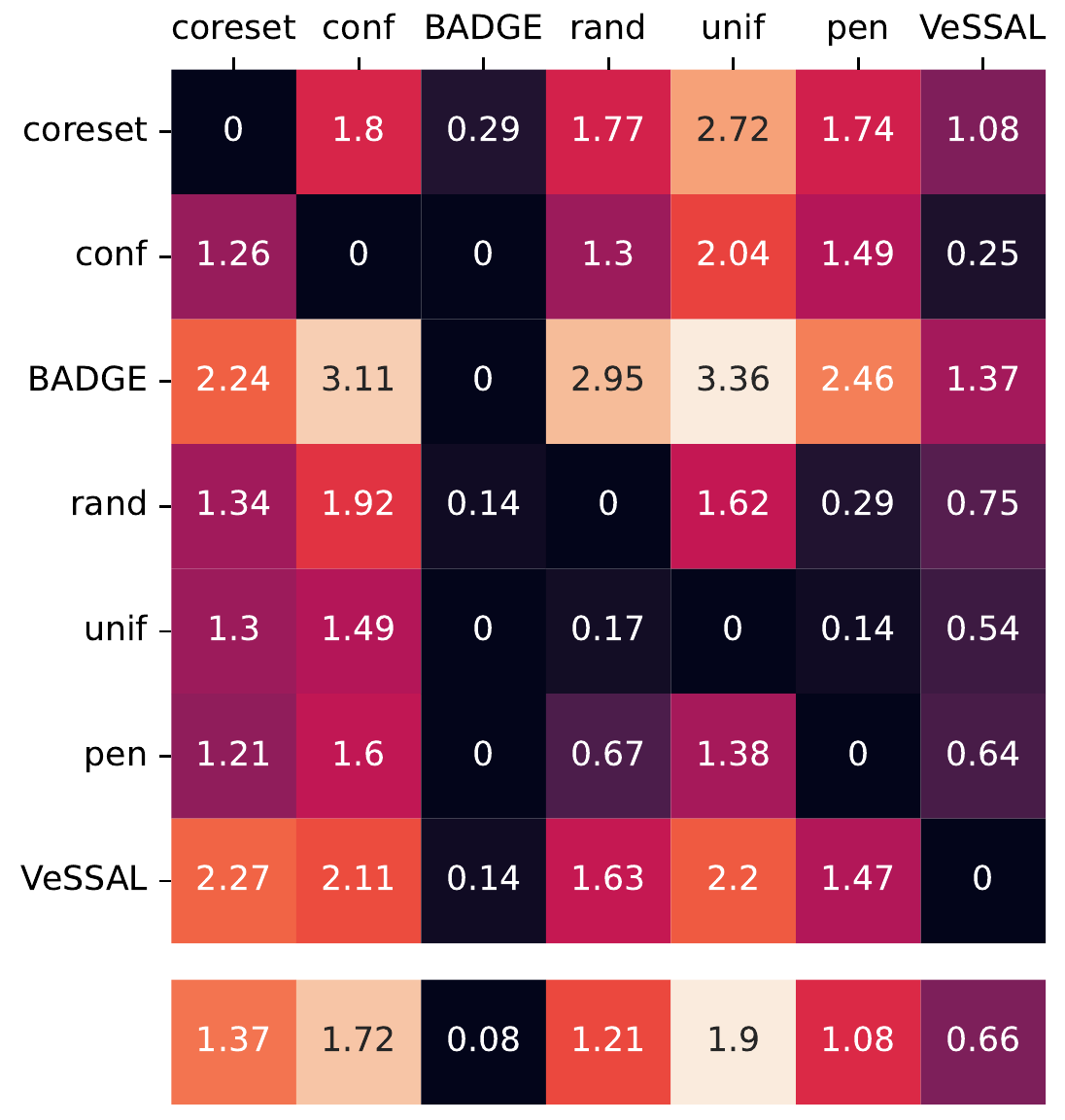}
     }
    \caption{Pairwise penalty matrix for all experiments with (a) I.I.D. data streams and (b) Non-I.I.D data streams. Each cell corresponds roughly to the amount of times the row algorithm outperforms the column algorithm by a statistically significant amount. Averages are shown at the bottom, where lower values imply better-performing algorithms. VeSSAL is the highest-performing streaming approach, and is only bested by BADGE, a non-streaming baseline.\looseness=-1 
    }
    \label{fig:heatmap_unordered}
\end{figure*}

We evaluate the performance of VeSSAL against several baselines on three academic benchmark datasets and one real-world dataset. In addition to measuring performance for a variety of architectures and batch sizes, we evaluate our approach in terms of robustness to feature drift in the data stream, and in terms of its fidelity to the predefined query rate.

\textbf{Baselines.} We compare our method against the following set of baselines. Most of these are non-streaming, meaning they have access to all unlabeled data when selecting samples to query. We also introduce two streaming baselines. 
\begin{itemize}
    \vspace{-0.3cm}
    \item \textbf{BADGE:} A recent, hyperparameter-free approach that incorporates both uncertainty and diversity in sampling using k-means++ in the hallucinated gradient space (Eq.~\ref{eq:one})~\citep{ash2019deep} (non-streaming).
    
    \item \textbf{rand:} A naive random sampling baseline (non-streaming). 
    
    \item \textbf{conf:} An uncertainty-based method that selects samples with smallest probability $p_{\hat{y}}$ of the top predicted class: $p_{\hat{y}} = \max f(x, \theta)$~\citep{wang2014new} (non-streaming).   
    
    \item \textbf{coreset:} A diversity-based method that uses a greedy approximation to the k-center problem on representations from the model's penultimate layer~\citep{sener2018active} (non-streaming). 
    
    \item \textbf{VeSSAL-pen:} This baseline is similar to VeSSAL but uses the penultimate layer embeddings instead of hallucinated gradients of the last layer, making it a purely diversity-based hyperparameter-free approach (streaming). 

    \item \textbf{stream-uniform:} A naive baseline for the streaming setting where data points are sampled at a fixed frequency as they arrive (streaming).
\end{itemize}

At each round of active learning, streaming algorithms are only permitted to see each unlabeled example once, at whatever time it is presented. Further, they must commit to a labeling decision as soon as a sample is encountered, and are unable to refine their decisions as more data arrive. This puts the streaming approaches at a marked disadvantage in comparison to their non-streaming peers.

\paragraph{Datasets.} We evaluate all algorithms on three image benchmarks, namely \texttt{SVHN}~\citep{svhn}, \texttt{MNIST}~\mbox{\citep{mnist}}, and \texttt{CIFAR10}~\citep{cifar}, and one real-world dataset from~\citet{bohus2022continual}. We refer to this dataset as \texttt{CLOW} and use it with permission from the authors. \texttt{CLOW} is collected through an augmented reality (AR) human-computer interaction device, where users provide object labels through a headset as they interact with objects in their home. This dataset includes $43$ object classes and $\sim11$K training samples. More details about the dataset and its preprocessing are described in Appendix~\ref{sec:clow_details}.

\begin{figure*}[!t]
    \centering
    \includegraphics[width=0.9\textwidth]{imgs/LR_legend_1.pdf}
    \includegraphics[width=0.32\textwidth]{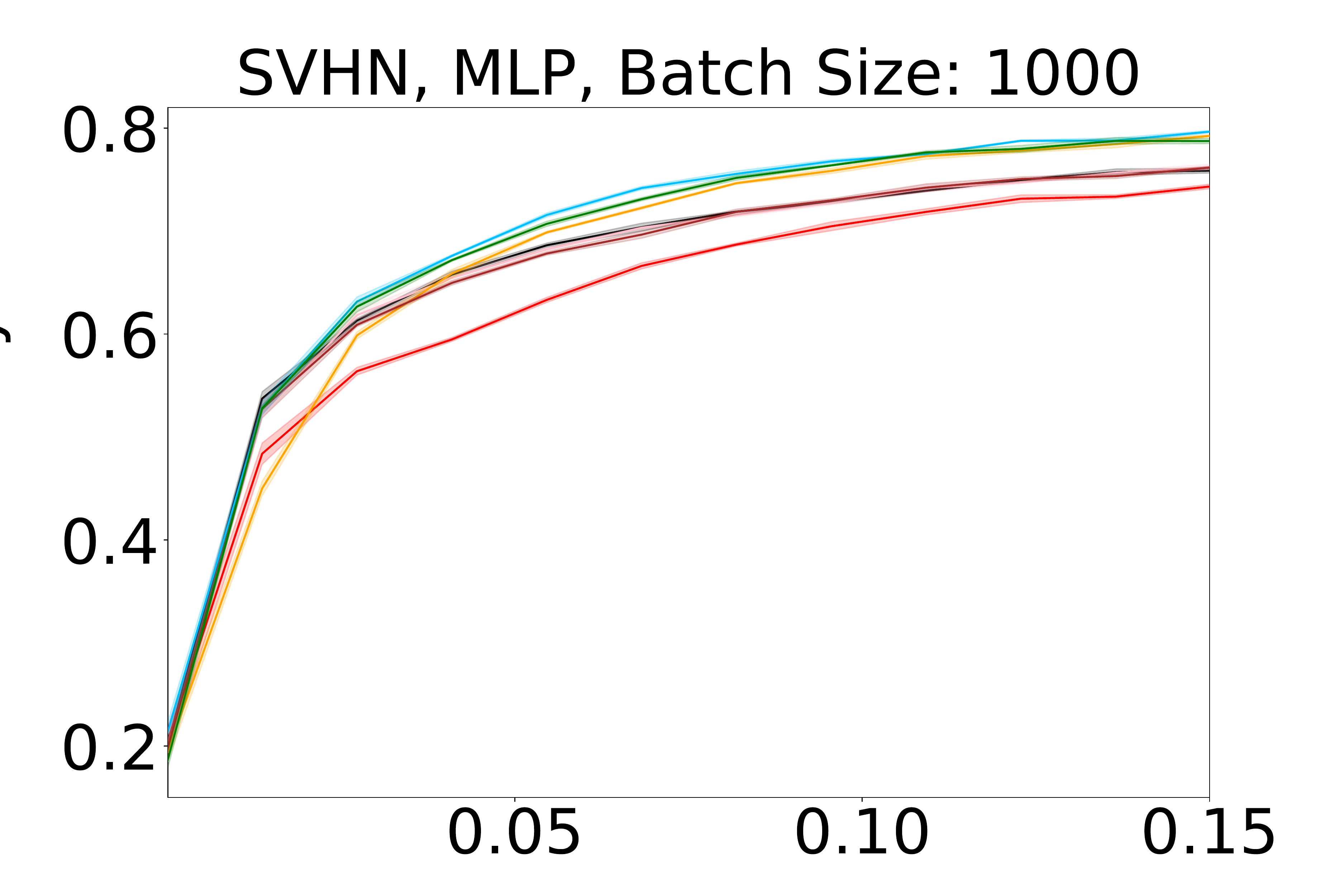}
    \includegraphics[width=0.32\textwidth]{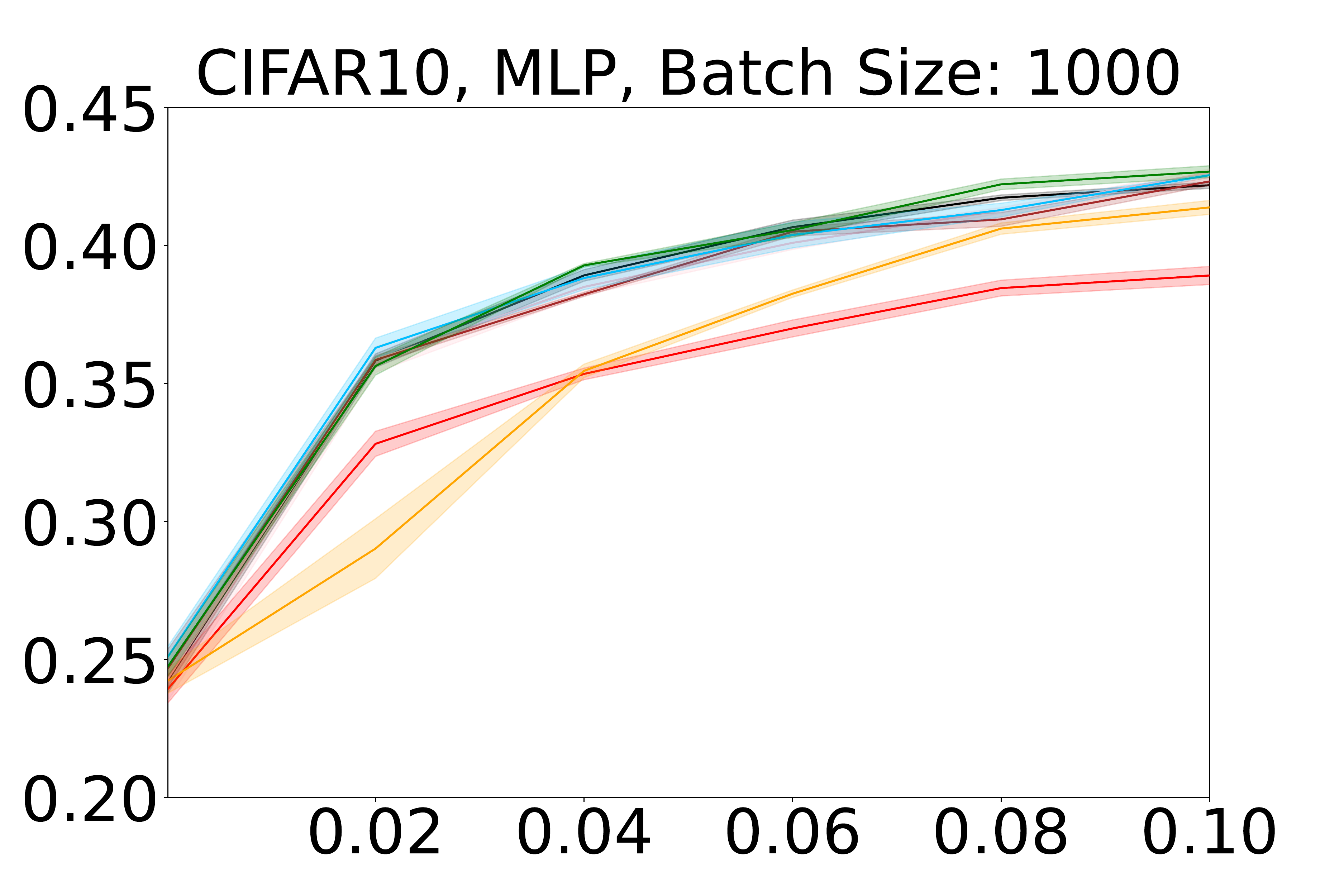}
    \includegraphics[width=0.32\textwidth]{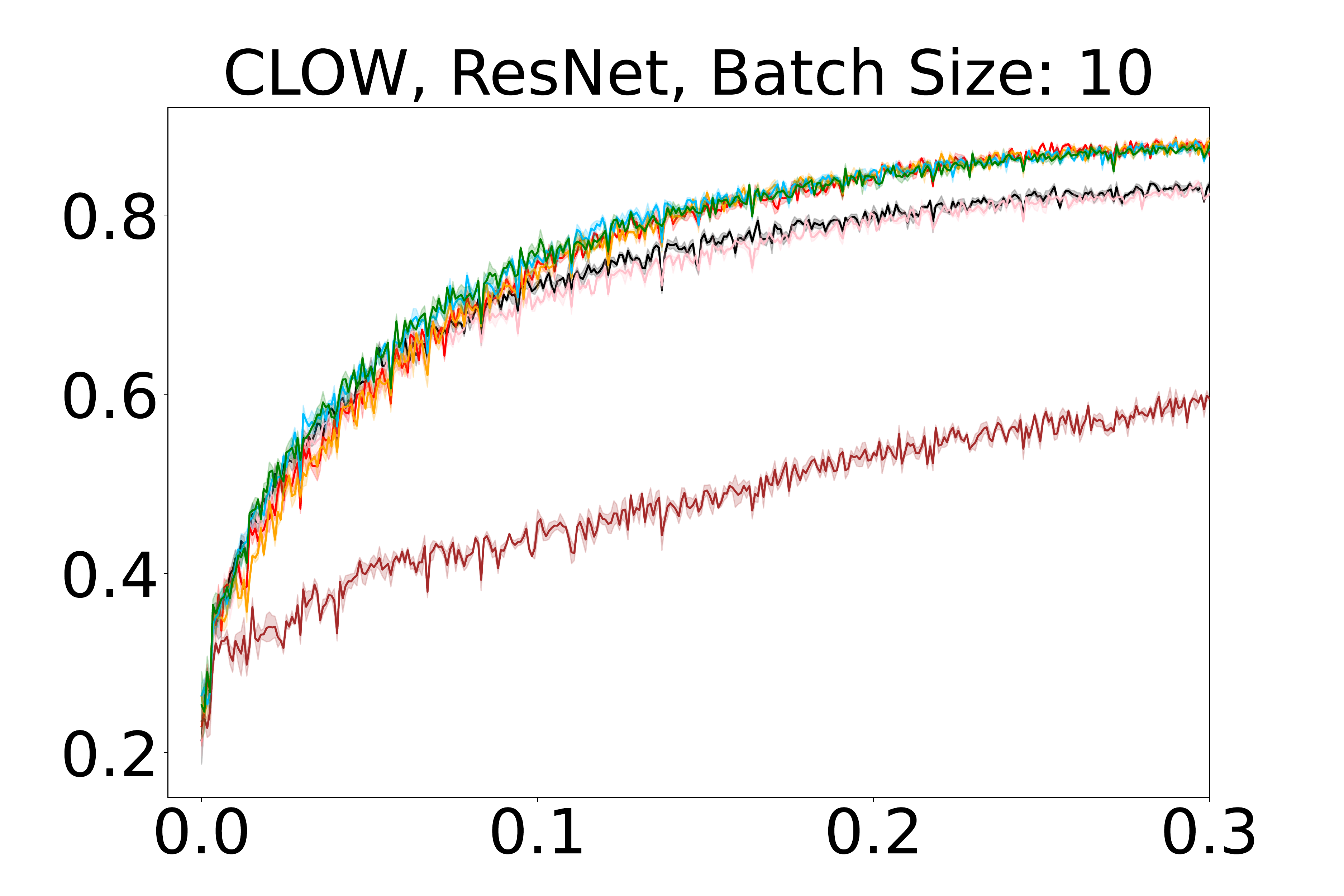}
    \caption{Learning curves for different neural active learning methods tested with non-I.I.D. data streams. Two network architectures, two batch sizes, and three datasets are shown here. These plots have been zoomed to highlight discriminative regions, but complete results are shown in the Appendix and are aggregated in Figure~\ref{fig:heatmap_unordered}(b).}
    \label{fig:learning_curves_drift}
\end{figure*}

\paragraph{Setup.} We perform multiple rounds of active learning in all experiments, with a fixed budget $k$ in each round. We consider a single round of active learning to correspond to a single batch of acquired points. We conduct as many rounds as are required to fully label the dataset under consideration. On these datasets, where the number of candidates is known, we choose to let the target rate $q_t$ evolve throughout the streaming process as $q_t = \frac{k - |B_t|}{n - t}$, where $n$ is the total number of samples in the unlabeled pool and $B_t$ is the set of samples that have been added to the batch so far. This allows the algorithm to adjust its sampling behavior in case of a flawed approximation of $\mathbb{E}_x [g(x) g(x)^\top]$. Still, even with this precaution, the sampling rate seems to be somewhat constant at $\frac{k}{n}$ (Figure~\ref{fig:sampling_rate}). We emphasize that in the case of real-world streaming settings, where the total number of unlabeled samples is unknown, $q_t$ can be set to any desired frequency.\looseness=-1

It is worth mentioning that this setup technically makes our approach committal only for a fixed round of active learning. That is, a sample that is not selected at one round will be made available again on the next. Our approach is made to work with truly committal environments, but we adopt this setup so that we can readily compare with non-committal, non-streaming, batch-mode active learning benchmarks.

After each round of data acquisition, we train the models from scratch and measure accuracy on a held-out test set.
We primarily experiment with two architectures, a two-layer MLP and an 18-layer ResNet~\citep{resnet}. All datasets are considered with both architectures except for \texttt{MNIST}, for which we only use the MLP. Models are trained with the Adam optimizer~\citep{kingma2014adam} with a fixed learning rate $0.001$ until they reach $>99\%$ training accuracy. We experiment with different budgets per round $k \in \{100, 1\mathrm{K}, 10\mathrm{K}\}$ for the benchmark datasets and $k \in \{10, 100, 1\mathrm{K}\}$ for the \texttt{CLOW} dataset (since it has a total of $\sim11\mathrm{K}$ training samples). Each experiment was repeated three times, and we show both mean and standard error in our learning curve plots. In all experiments we start with $100$ labeled samples and acquire the rest of the labeled samples via active learning. All methods are implemented in PyTorch \citep{pytorch}. We set $\lambda$ in Algorithm~\ref{alg:main} to $.01$ in all VeSSAL experiments, which ensures a numerically stable inversion.

In the subsections that follow, we show experimental results under two different assumptions about the data stream: I.I.D data streams and adversarially ordered data streams.

\begin{figure*}[t]
   
    \centering
    \subfigure[VeSSAL]{\includegraphics[width=0.9\textwidth]{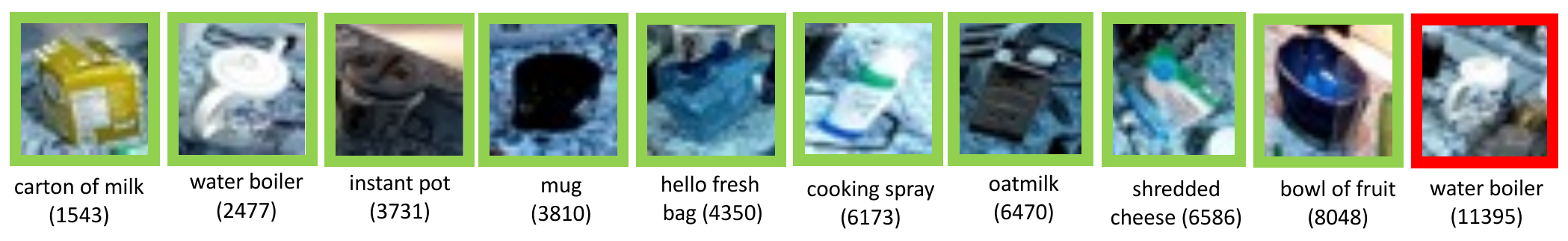}}
    \subfigure[Uniform Rate Sampling]{\includegraphics[width=0.9\textwidth]{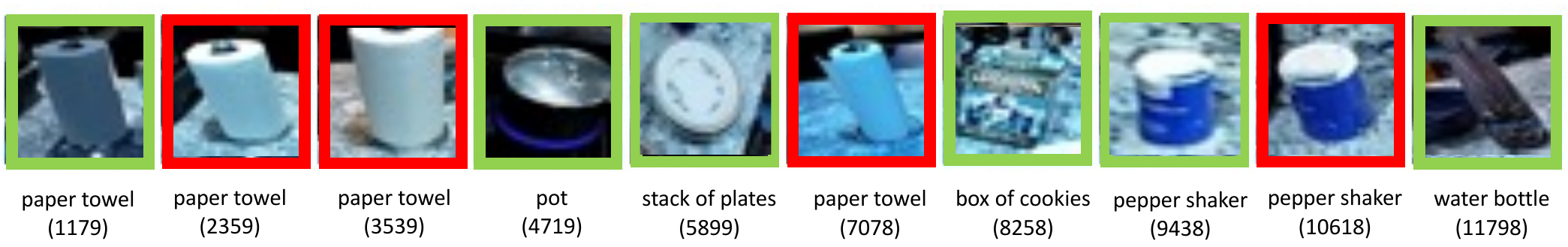}}
    \caption{The first round of queries on the data stream from the \texttt{CLOW} dataset for different streaming active learning methods with an MLP and budget $k = 10$ (top: VeSSAL, bottom: uniform rate sampling). Red boxes denote repeated classes and green boxes denote unique classes. VeSSAL only repeats a single class, corresponding to an object view that is quite different than the other selection of the same class. Each queried image is described by its label name and time stamp (in parentheses) depicting the index at which it arrives in the stream. Here, there are about $11.8$k candidates, and the indices suggest that sampling  mass is well-distributed across the stream.
    }
    \label{fig:hololens_queries}
\end{figure*}

\subsection{I.I.D. Data Stream}

In this section we discuss experimental results when the data stream is randomized, meaning there is no induced correlation between consecutive data points or in the order of samples in the stream. We show that VeSSAL is superior or equal to non-streaming algorithms in this setting.

Representative learning curves averaged over three replicates for various choices of batch size, dataset, and model architecture in this setting are shown in Figure~\ref{fig:learning_curves_main}. In each, VeSSAL performs about as well as the highest-performing, non-streaming baseline, despite being restricted to the demands of the streaming setting. Detailed results for each combination of dataset, batch size, and network architecture are shown in Appendix~\ref{sec:lr_iid}.\looseness=-1

We aggregate results following the protocol of \citet{ash2019deep}. For each active learning experiment, we only consider a subset of labeling budgets where learning is still progressing. This is due to the fact that with labeling budgets approaching the data-set size, all algorithms achieve similar accuracy. At exponentially spaced intervals, we calculate if a given row algorithm outperforms a given column algorithm by a statistically significant margin according to a two-sided t-test. When this happens we increment the corresponding cell of Figure~\ref{fig:heatmap_unordered}(a) by 1 divided by the total number of evaluations in the experiment. More experimental details can be found in Appendix~\ref{sec:heatmap_details}.

Here, higher-performing algorithms are associated with a lower column-wise average, displayed at the bottom of the figure. We see that overall, VeSSAL is the highest performing streaming method. When considering all baselines, VeSSAL is only outperformed by BADGE, which is not encumbered by streaming requirements. In a small number of experiments, VeSSAL surprisingly even manages to outperform BADGE.

\subsection{Non-I.I.D. Data Stream}

To investigate the robustness of our algorithm to non-I.I.D. circumstances, we compare all methods under naturally occurring or artificially induced domain drift, where the observed data distribution is non-stationary. Note that non-streaming baselines are not at all affected by this change, and it is only a burden for streaming approaches that use sequential estimates of data statistics for decision making. Despite the disadvantage, we show that VeSSAL is still performing roughly as well as state-of-the-art, non-streaming baselines.\looseness=-1

\textbf{Artificial Drift} We adversarially sort the unlabeled data to introduce domain drift. To do so, sort two academic datasets (\texttt{CIFAR-10}, \texttt{SVHN}) by their first principal component.

\textbf{Natural Drift} The \texttt{CLOW} data stream is sorted by the timestamp at which objects were encountered and labeled by a user, and hence naturally contains feature drift 
(Figure~\ref{fig:hololens_queries}). For this dataset, MLP and ResNet-18 architectures have pretrained components. In the MLP case, we use data representations taken from the visual encoder of the multimodal CLIP model~\citep{radford2021learning}, and although the MLP is trained from scratch at each round, the CLIP feature extractor is fixed. In the ResNet case, we use a model that has been pretrained on ImageNet~\citep{russakovsky2015imagenet}, and refine the entire model with actively selected data. Each time new data are acquired, the ResNet is reset to to the ImageNet pretrained weights before being updated.\looseness=-1

Figure~\ref{fig:learning_curves_drift} highlights the effectiveness of VeSSAL under the challenging setting of feature drift in the data stream. VeSSAL performs both on par with other non-streaming skyline approaches and better than other streaming baseline methods. Detailed learning curves for all datasets, architectures, and hyperaparameters are shown in Appendix~\ref{sec:lr_non_iid}. In Figure~\ref{fig:heatmap_unordered}(b), a pairwise comparison matrix analogous to Figure~\ref{fig:heatmap_unordered}(a) shows that, with the exception of BADGE, VeSSAL outperforms all streaming and non-streaming baselines in presence of distribution shift. Again, VeSSAL sometimes outperforms BADGE even though BADGE sees all unlabeled data before sampling. Figure~\ref{fig:hololens_queries} contains qualitative evidence that VeSSAL samples diverse images even when data points are correlated and the streaming uniform sampling baseline repeatedly queries duplicate objects.

\begin{figure}[!thb]
    \centering

    \includegraphics[width=0.24\textwidth]{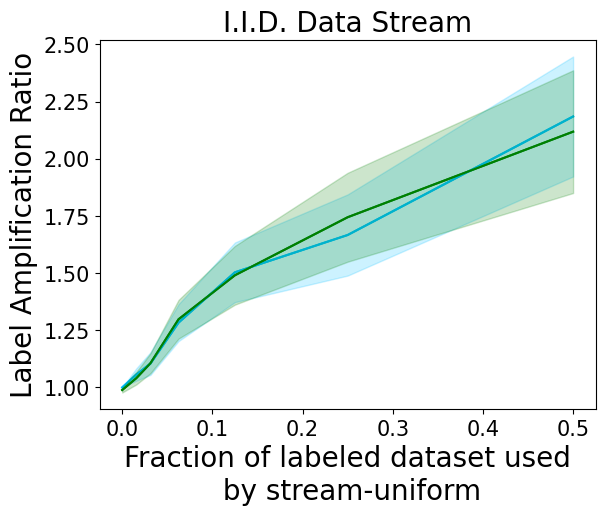}
    \includegraphics[width=0.22\textwidth]{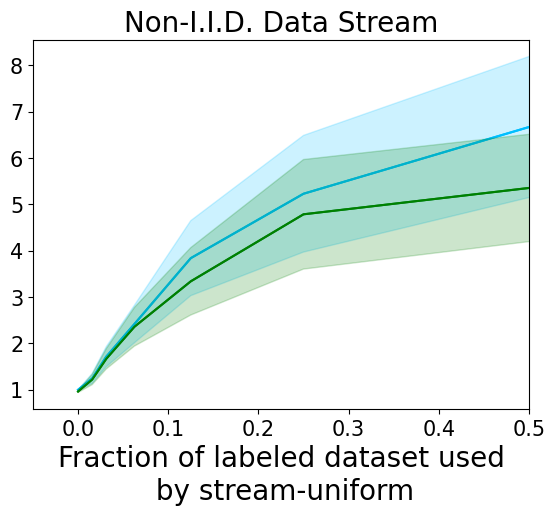}\\
    \includegraphics[width=0.20\textwidth]{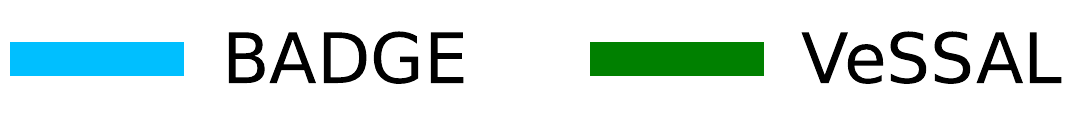}
    \caption{Label amplification for BADGE and VeSSAL with respect to stream-uniform, averaged over all experiments. VeSSAL achieves roughly the same label efficiency as BADGE in the I.I.D. case, even though VeSSAL is limited by streaming requirements. In the non-I.I.D. setting, which is adversarial for streaming algorithms like VeSSAL but not for pool-based algorithms like BADGE, VeSSAL only does slightly worse than BADGE.\looseness=-1
    }
    \label{fig:label_amp}
\end{figure}

\subsection{Label Amplification} 
The promise of active learning is that it can deliver significantly more predictive power for a fixed labeling budget than naive sampling. This is demonstrated by Figure~\ref{fig:label_amp}, where algorithm performance is cast in terms of ``label amplification'' instead of accuracy. As learning progresses, we plot the ratio between the number of samples used by streaming uniform sampling and the number of samples required by active sampling to achieve the same performance. For a good active learning algorithm, labeling amplification will be much larger than one, reflecting the increase in labeling efficiency over passive sampling. Our plots average over all experiments conducted, and despite VeSSAL being constrained to the streaming, committal setting, they show that it is roughly as efficient as BADGE in the I.I.D. streaming case and only slightly worse in the non-I.I.D. case. 

\subsection{Compute Requirements} 

VeSSAL is able to decide whether to include a given unlabeled sample in the batch as soon as it is encountered. In doing so, and unlike previous neural batch active learning algorithms, VeSSAL does not need to compare every unlabeled candidate point to every sample that has already been selected. Particularly for large batch sizes, this makes VeSSAL substantially more time efficient than baseline neural active learning algorithms. Figure~\ref{fig:compute_plot} (logged y-axis version shown in Appendix Figure~\ref{fig:compute_plot_log}) demonstrates this by comparing the run time of several active learning algorithms as a function of their query batch size using the CIFAR-10 dataset. VeSSAL enjoys run times times that stay nearly fixed for increasing batch sizes, while other approaches have compute requirements that grow superlinearly. For large batch sizes, VeSSAL can be more efficient than its counterparts by several orders of magnitude.

\begin{figure}
    \includegraphics[width=0.5\textwidth]{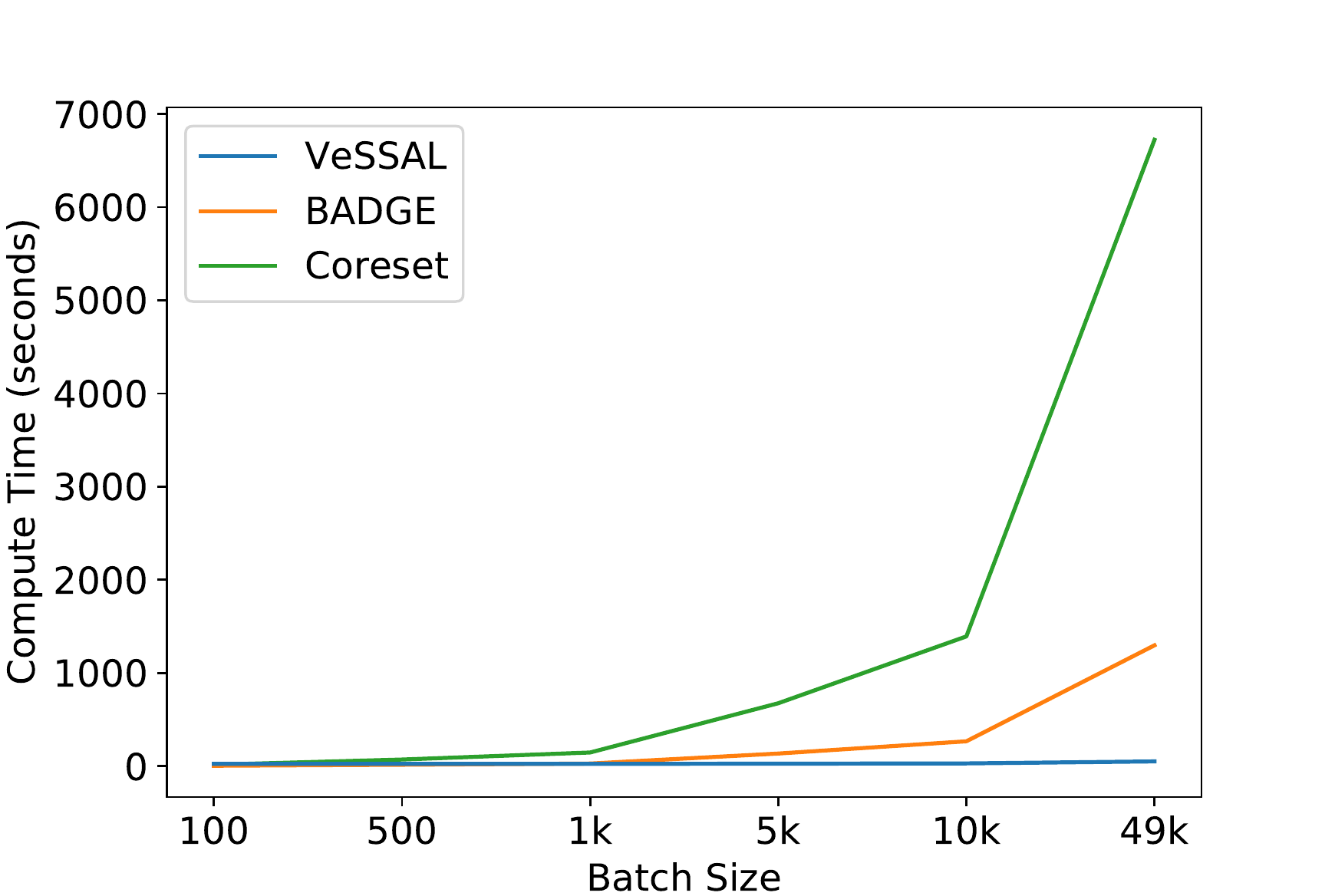}
    \caption{The compute time required to select a batch for three different algorithms on CIFAR-10 as a function of the query batch size. While non-streaming algorithms require compute that grows superlinearly as a function of labeling budget, VeSSAL stays relatively constant. Results are averaged over five replicates and each algorithm was given identical computational resources. }
    \label{fig:compute_plot}
\end{figure}
\section{Discussion}
\label{sec:discussion}

We presented VeSSAL, a new approach for batch active learning with deep neural networks in a streaming setting. Unlike prior pool-based active learning approaches for deep neural networks, our method can commit to queries as soon as samples are made available to the model from a data stream. Even in fixed-data settings, VeSSAL performs roughly as well as state-of-the-art methods — despite the fact that they are not hindered by streaming constraints. We envision several potential benefits of the proposed approach, expanding the applicability of neural networks for real-world, interaction-centric applications. Our algorithm can be run in settings that are inherently streaming, committal, or on datasets too large to be entirely stored in one place.\looseness=-1

Our work also opens up exciting directions for future research. Currently, VeSSAL assumes that the number of label classes is known a priori. In real-world applications, where an interaction environment continues to evolve, the number of label classes may grow over time. 
Moreover, some queries can be costlier than others. For example, asking a user for an object label in their peripheral vision could distract them from performing the current task. Addressing these challenges are exciting topics for further work.

\section*{Acknowledgements}

The authors would like to thank Sean Andrist, Dan Bohus, and the Platform for Situated Intelligence (PSI) team at Microsoft Research Redmond for providing inspiration on the problem statement, feedback on our approach, and access to the \texttt{CLOW} dataset.

\bibliography{active}

\begin{thebibliography}{58}
\providecommand{\natexlab}[1]{#1}
\providecommand{\url}[1]{\texttt{#1}}
\expandafter\ifx\csname urlstyle\endcsname\relax
  \providecommand{\doi}[1]{doi: #1}\else
  \providecommand{\doi}{doi: \begingroup \urlstyle{rm}\Url}\fi

\bibitem[Ash et~al.(2021)Ash, Goel, Krishnamurthy, and Kakade]{ash2021gone}
Ash, J., Goel, S., Krishnamurthy, A., and Kakade, S.
\newblock Gone fishing: Neural active learning with fisher embeddings.
\newblock \emph{Advances in Neural Information Processing Systems},
  34:\penalty0 8927--8939, 2021.

\bibitem[Ash \& Adams(2020)Ash and Adams]{ash2019warm}
Ash, J.~T. and Adams, R.~P.
\newblock On warm-starting neural network training.
\newblock \emph{Advances in Neural Information Processing Systems}, 2020.

\bibitem[Ash et~al.(2020)Ash, Zhang, Krishnamurthy, Langford, and
  Agarwal]{ash2019deep}
Ash, J.~T., Zhang, C., Krishnamurthy, A., Langford, J., and Agarwal, A.
\newblock Deep batch active learning by diverse, uncertain gradient lower
  bounds.
\newblock \emph{International Conference on Learning Representations}, 2020.

\bibitem[Ban et~al.(2022)Ban, Zhang, Tong, Banerjee, and He]{ban2022improved}
Ban, Y., Zhang, Y., Tong, H., Banerjee, A., and He, J.
\newblock Improved algorithms for neural active learning.
\newblock \emph{Advances in eural Information Processing Systems}, 2022.

\bibitem[Bardenet et~al.(2017)Bardenet, Lavancier, Mary, and
  Vasseur]{bardenet2017few}
Bardenet, R., Lavancier, F., Mary, X., and Vasseur, A.
\newblock On a few statistical applications of determinantal point processes.
\newblock \emph{ESAIM: Proceedings and Surveys}, 60:\penalty0 180--202, 2017.

\bibitem[Beluch et~al.(2018)Beluch, Genewein, N{\"u}rnberger, and
  K{\"o}hler]{beluch2018power}
Beluch, W.~H., Genewein, T., N{\"u}rnberger, A., and K{\"o}hler, J.~M.
\newblock The power of ensembles for active learning in image classification.
\newblock In \emph{Proceedings of the IEEE conference on computer vision and
  pattern recognition}, pp.\  9368--9377, 2018.

\bibitem[Beygelzimer et~al.(2009)Beygelzimer, Dasgupta, and Langford]{BDL09}
Beygelzimer, A., Dasgupta, S., and Langford, J.
\newblock Importance weighted active learning.
\newblock In \emph{Twenty-Sixth International Conference on Machine Learning},
  2009.

\bibitem[Beygelzimer et~al.(2010)Beygelzimer, Hsu, Langford, and
  Zhang]{beygelzimer2010agnostic}
Beygelzimer, A., Hsu, D.~J., Langford, J., and Zhang, T.
\newblock Agnostic active learning without constraints.
\newblock In \emph{Neural Information Processing Systems}, 2010.

\bibitem[Bhaskara et~al.(2019)Bhaskara, Lattanzi, Vassilvitskii, and
  Zadimoghaddam]{bhaskara2019residual}
Bhaskara, A., Lattanzi, S., Vassilvitskii, S., and Zadimoghaddam, M.
\newblock Residual based sampling for online low rank approximation.
\newblock In \emph{2019 IEEE 60th Annual Symposium on Foundations of Computer
  Science (FOCS)}, pp.\  1596--1614. IEEE, 2019.

\bibitem[Bingham \& Mannila(2001)Bingham and Mannila]{bingham2001random}
Bingham, E. and Mannila, H.
\newblock Random projection in dimensionality reduction: applications to image
  and text data.
\newblock In \emph{Proceedings of the seventh ACM SIGKDD international
  conference on Knowledge discovery and data mining}, pp.\  245--250, 2001.

\bibitem[Bohus et~al.(2022)Bohus, Andrist, Feniello, Saw, and
  Horvitz]{bohus2022continual}
Bohus, D., Andrist, S., Feniello, A., Saw, N., and Horvitz, E.
\newblock Continual learning about objects in the wild: An interactive
  approach.
\newblock In \emph{Proceedings of the 2022 International Conference on
  Multimodal Interaction}, pp.\  476--486, 2022.

\bibitem[Boutsidis et~al.(2014)Boutsidis, Garber, Karnin, and
  Liberty]{boutsidis2014online}
Boutsidis, C., Garber, D., Karnin, Z., and Liberty, E.
\newblock Online principal components analysis.
\newblock In \emph{Proceedings of the twenty-sixth annual ACM-SIAM symposium on
  Discrete algorithms}, pp.\  887--901. SIAM, 2014.

\bibitem[Braverman et~al.(2011)Braverman, Meyerson, Ostrovsky, Roytman,
  Shindler, and Tagiku]{braverman2011streaming}
Braverman, V., Meyerson, A., Ostrovsky, R., Roytman, A., Shindler, M., and
  Tagiku, B.
\newblock Streaming k-means on well-clusterable data.
\newblock In \emph{Proceedings of the twenty-second annual ACM-SIAM symposium
  on Discrete Algorithms}, pp.\  26--40. SIAM, 2011.

\bibitem[Brust et~al.(2018)Brust, K{\"a}ding, and Denzler]{brust2018active}
Brust, C.-A., K{\"a}ding, C., and Denzler, J.
\newblock Active learning for deep object detection.
\newblock \emph{arXiv preprint arXiv:1809.09875}, 2018.

\bibitem[Choi et~al.(2021)Choi, Elezi, Lee, Farabet, and
  Alvarez]{choi2021active}
Choi, J., Elezi, I., Lee, H.-J., Farabet, C., and Alvarez, J.~M.
\newblock Active learning for deep object detection via probabilistic modeling.
\newblock In \emph{Proceedings of the IEEE/CVF International Conference on
  Computer Vision}, pp.\  10264--10273, 2021.

\bibitem[Dasgupta(2011)]{D11}
Dasgupta, S.
\newblock Two faces of active learning.
\newblock \emph{Theoretical computer science}, 2011.

\bibitem[Deshpande \& Vempala(2006)Deshpande and
  Vempala]{deshpande2006adaptive}
Deshpande, A. and Vempala, S.
\newblock Adaptive sampling and fast low-rank matrix approximation.
\newblock In \emph{Approximation, Randomization, and Combinatorial
  Optimization. Algorithms and Techniques}, pp.\  292--303. Springer, 2006.

\bibitem[Deshpande et~al.(2006)Deshpande, Rademacher, Vempala, and
  Wang]{deshpande2006matrix}
Deshpande, A., Rademacher, L., Vempala, S.~S., and Wang, G.
\newblock Matrix approximation and projective clustering via volume sampling.
\newblock \emph{Theory of Computing}, 2\penalty0 (1):\penalty0 225--247, 2006.

\bibitem[Ducoffe \& Precioso(2018)Ducoffe and Precioso]{ducoffe2018adversarial}
Ducoffe, M. and Precioso, F.
\newblock Adversarial active learning for deep networks: a margin based
  approach.
\newblock \emph{arXiv:1802.09841}, 2018.

\bibitem[Frieze et~al.(2004)Frieze, Kannan, and Vempala]{frieze2004fast}
Frieze, A., Kannan, R., and Vempala, S.
\newblock Fast monte-carlo algorithms for finding low-rank approximations.
\newblock \emph{Journal of the ACM (JACM)}, 51\penalty0 (6):\penalty0
  1025--1041, 2004.

\bibitem[Gal et~al.(2017)Gal, Islam, and Ghahramani]{gal2017deep}
Gal, Y., Islam, R., and Ghahramani, Z.
\newblock Deep bayesian active learning with image data.
\newblock In \emph{International Conference on Machine Learning}, 2017.

\bibitem[Geifman \& El-Yaniv(2017)Geifman and El-Yaniv]{geifman2017deep}
Geifman, Y. and El-Yaniv, R.
\newblock Deep active learning over the long tail.
\newblock \emph{arXiv:1711.00941}, 2017.

\bibitem[Ghashami \& Phillips(2014)Ghashami and Phillips]{ghashami2014relative}
Ghashami, M. and Phillips, J.~M.
\newblock Relative errors for deterministic low-rank matrix approximations.
\newblock In \emph{Proceedings of the twenty-fifth annual ACM-SIAM symposium on
  Discrete algorithms}, pp.\  707--717. SIAM, 2014.

\bibitem[Gissin \& Shalev-Shwartz(2019)Gissin and
  Shalev-Shwartz]{gissin2019discriminative}
Gissin, D. and Shalev-Shwartz, S.
\newblock Discriminative active learning.
\newblock \emph{arXiv:1907.06347}, 2019.

\bibitem[Greub(2012)]{greub2012linear}
Greub, W.~H.
\newblock \emph{Linear algebra}, volume~23.
\newblock Springer Science \& Business Media, 2012.

\bibitem[Gudovskiy et~al.(2020)Gudovskiy, Hodgkinson, Yamaguchi, and
  Tsukizawa]{gudovskiy2020deep}
Gudovskiy, D., Hodgkinson, A., Yamaguchi, T., and Tsukizawa, S.
\newblock Deep active learning for biased datasets via fisher kernel
  self-supervision.
\newblock In \emph{Proceedings of the IEEE/CVF Conference on Computer Vision
  and Pattern Recognition}, pp.\  9041--9049, 2020.

\bibitem[Hanneke(2014{\natexlab{a}})]{H14}
Hanneke, S.
\newblock Theory of disagreement-based active learning.
\newblock \emph{Foundations and Trends in Machine Learning},
  2014{\natexlab{a}}.

\bibitem[Hanneke(2014{\natexlab{b}})]{hanneke2014theory}
Hanneke, S.
\newblock Theory of active learning.
\newblock \emph{Foundations and Trends in Machine Learning}, 7\penalty0 (2-3),
  2014{\natexlab{b}}.

\bibitem[Hanneke \& Yang(2015)Hanneke and Yang]{hanneke2015minimax}
Hanneke, S. and Yang, L.
\newblock Minimax analysis of active learning.
\newblock \emph{J. Mach. Learn. Res.}, 16\penalty0 (1):\penalty0 3487--3602,
  2015.

\bibitem[He et~al.(2016)He, Zhang, Ren, and Sun]{resnet}
He, K., Zhang, X., Ren, S., and Sun, J.
\newblock Deep residual learning for image recognition.
\newblock In \emph{Proceedings of the IEEE conference on computer vision and
  pattern recognition}, pp.\  770--778, 2016.

\bibitem[Hsu(2010)]{hsu2010algorithms}
Hsu, D.~J.
\newblock \emph{Algorithms for active learning}.
\newblock PhD thesis, UC San Diego, 2010.

\bibitem[Huang et~al.(2015)Huang, Agarwal, Hsu, Langford, and
  Schapire]{huang2015efficient}
Huang, T.-K., Agarwal, A., Hsu, D.~J., Langford, J., and Schapire, R.~E.
\newblock Efficient and parsimonious agnostic active learning.
\newblock \emph{Advances in Neural Information Processing Systems}, 28, 2015.

\bibitem[Kingma \& Ba(2014)Kingma and Ba]{kingma2014adam}
Kingma, D.~P. and Ba, J.
\newblock Adam: A method for stochastic optimization.
\newblock \emph{arXiv preprint arXiv:1412.6980}, 2014.

\bibitem[Kovashka et~al.(2016)Kovashka, Russakovsky, Fei-Fei, Grauman,
  et~al.]{kovashka2016crowdsourcing}
Kovashka, A., Russakovsky, O., Fei-Fei, L., Grauman, K., et~al.
\newblock Crowdsourcing in computer vision.
\newblock \emph{Foundations and Trends{\textregistered} in computer graphics
  and Vision}, 10\penalty0 (3):\penalty0 177--243, 2016.

\bibitem[Krishnamurthy et~al.(2017)Krishnamurthy, Agarwal, Huang,
  Daum{\'e}~III, and Langford]{krishnamurthy2017active}
Krishnamurthy, A., Agarwal, A., Huang, T.-K., Daum{\'e}~III, H., and Langford,
  J.
\newblock Active learning for cost-sensitive classification.
\newblock In \emph{International Conference on Machine Learning}, pp.\
  1915--1924. PMLR, 2017.

\bibitem[Krizhevsky(2009)]{cifar}
Krizhevsky, A.
\newblock Learning multiple layers of features from tiny images.
\newblock Technical report, Citeseer, 2009.

\bibitem[Kulesza et~al.(2012)Kulesza, Taskar, et~al.]{kulesza2012determinantal}
Kulesza, A., Taskar, B., et~al.
\newblock Determinantal point processes for machine learning.
\newblock \emph{Foundations and Trends{\textregistered} in Machine Learning},
  5\penalty0 (2--3):\penalty0 123--286, 2012.

\bibitem[Lavania et~al.(2021)Lavania, Wei, Iyer, and
  Bilmes]{lavania2021practical}
Lavania, C., Wei, K., Iyer, R., and Bilmes, J.
\newblock A practical online framework for extracting running video summaries
  under a fixed memory budget.
\newblock In \emph{Proceedings of the 2021 SIAM International Conference on
  Data Mining (SDM)}, pp.\  226--234. SIAM, 2021.

\bibitem[LeCun et~al.(1998)LeCun, Bottou, Bengio, Haffner, et~al.]{mnist}
LeCun, Y., Bottou, L., Bengio, Y., Haffner, P., et~al.
\newblock Gradient-based learning applied to document recognition.
\newblock \emph{IEEE}, 1998.

\bibitem[Liu et~al.(2016)Liu, Anguelov, Erhan, Szegedy, Reed, Fu, and
  Berg]{liu2016ssd}
Liu, W., Anguelov, D., Erhan, D., Szegedy, C., Reed, S., Fu, C.-Y., and Berg,
  A.~C.
\newblock {SSD}: Single shot multibox detector.
\newblock In \emph{European conference on computer vision}, pp.\  21--37.
  Springer, 2016.

\bibitem[MacKay(1992)]{mackay1992information}
MacKay, D.~J.
\newblock Information-based objective functions for active data selection.
\newblock \emph{Neural computation}, 4\penalty0 (4):\penalty0 590--604, 1992.

\bibitem[Netzer et~al.(2011)Netzer, Wang, Coates, Bissacco, Wu, and Ng]{svhn}
Netzer, Y., Wang, T., Coates, A., Bissacco, A., Wu, B., and Ng, A.~Y.
\newblock Reading digits in natural images with unsupervised feature learning.
\newblock 2011.

\bibitem[Paszke et~al.(2017)Paszke, Gross, Chintala, Chanan, Yang, DeVito, Lin,
  Desmaison, Antiga, and Lerer]{pytorch}
Paszke, A., Gross, S., Chintala, S., Chanan, G., Yang, E., DeVito, Z., Lin, Z.,
  Desmaison, A., Antiga, L., and Lerer, A.
\newblock Automatic differentiation in pytorch.
\newblock 2017.

\bibitem[Radford et~al.(2021)Radford, Kim, Hallacy, Ramesh, Goh, Agarwal,
  Sastry, Askell, Mishkin, Clark, et~al.]{radford2021learning}
Radford, A., Kim, J.~W., Hallacy, C., Ramesh, A., Goh, G., Agarwal, S., Sastry,
  G., Askell, A., Mishkin, P., Clark, J., et~al.
\newblock Learning transferable visual models from natural language
  supervision.
\newblock In \emph{International conference on machine learning}, pp.\
  8748--8763. PMLR, 2021.

\bibitem[Ren et~al.(2021)Ren, Xiao, Chang, Huang, Li, Gupta, Chen, and
  Wang]{ren2021survey}
Ren, P., Xiao, Y., Chang, X., Huang, P.-Y., Li, Z., Gupta, B.~B., Chen, X., and
  Wang, X.
\newblock A survey of deep active learning.
\newblock \emph{ACM computing surveys (CSUR)}, 54\penalty0 (9):\penalty0 1--40,
  2021.

\bibitem[Roth \& Small(2006)Roth and Small]{roth2006margin}
Roth, D. and Small, K.
\newblock Margin-based active learning for structured output spaces.
\newblock In \emph{European Conference on Machine Learning}, 2006.

\bibitem[Roy et~al.(2018)Roy, Unmesh, and Namboodiri]{roy2018deep}
Roy, S., Unmesh, A., and Namboodiri, V.~P.
\newblock Deep active learning for object detection.
\newblock In \emph{BMVC}, pp.\ ~91, 2018.

\bibitem[Russakovsky et~al.(2015)Russakovsky, Deng, Su, Krause, Satheesh, Ma,
  Huang, Karpathy, Khosla, Bernstein, et~al.]{russakovsky2015imagenet}
Russakovsky, O., Deng, J., Su, H., Krause, J., Satheesh, S., Ma, S., Huang, Z.,
  Karpathy, A., Khosla, A., Bernstein, M., et~al.
\newblock Imagenet large scale visual recognition challenge.
\newblock \emph{International journal of computer vision}, 115:\penalty0
  211--252, 2015.

\bibitem[Sener \& Savarese(2018)Sener and Savarese]{sener2018active}
Sener, O. and Savarese, S.
\newblock Active learning for convolutional neural networks: A core-set
  approach.
\newblock In \emph{International Conference on Learning Representations}, 2018.

\bibitem[Senzaki \& Hamelain(2021)Senzaki and Hamelain]{senzaki2021active}
Senzaki, Y. and Hamelain, C.
\newblock Active learning for deep neural networks on edge devices.
\newblock \emph{arXiv preprint arXiv:2106.10836}, 2021.

\bibitem[Settles(2010)]{S10}
Settles, B.
\newblock Active learning literature survey.
\newblock \emph{University of Wisconsin, Madison}, 2010.

\bibitem[Singh et~al.(2016)Singh, Fatahalian, and Efros]{singh2016krishnacam}
Singh, K.~K., Fatahalian, K., and Efros, A.~A.
\newblock Krishnacam: Using a longitudinal, single-person, egocentric dataset
  for scene understanding tasks.
\newblock In \emph{2016 IEEE Winter Conference on Applications of Computer
  Vision (WACV)}, pp.\  1--9. IEEE, 2016.

\bibitem[Strang(2006)]{strang2006linear}
Strang, G.
\newblock \emph{Linear algebra and its applications.}
\newblock Belmont, CA: Thomson, Brooks/Cole, 2006.

\bibitem[Sun \& Gong(2019)Sun and Gong]{sun2019active}
Sun, L. and Gong, Y.
\newblock Active learning for image classification: A deep reinforcement
  learning approach.
\newblock In \emph{2019 2nd China Symposium on Cognitive Computing and Hybrid
  Intelligence (CCHI)}, pp.\  71--76. IEEE, 2019.

\bibitem[Sun et~al.(2022)Sun, Calandriello, Hu, Li, and
  Titsias]{sun2022information}
Sun, S., Calandriello, D., Hu, H., Li, A., and Titsias, M.
\newblock Information-theoretic online memory selection for continual learning.
\newblock \emph{journal={International Conference on Learning
  Representations}}, 2022.

\bibitem[Wang \& Shang(2014)Wang and Shang]{wang2014new}
Wang, D. and Shang, Y.
\newblock A new active labeling method for deep learning.
\newblock In \emph{International Joint Conference on Neural Networks}, 2014.

\bibitem[Wang et~al.(2021)Wang, Wang, Shang-Guan, and
  Gupta]{wang2021wanderlust}
Wang, J., Wang, X., Shang-Guan, Y., and Gupta, A.
\newblock Wanderlust: Online continual object detection in the real world.
\newblock In \emph{Proceedings of the IEEE/CVF International Conference on
  Computer Vision}, pp.\  10829--10838, 2021.

\bibitem[Woodbury(1950)]{woodbury1950inverting}
Woodbury, M.~A.
\newblock \emph{Inverting modified matrices}.
\newblock Statistical Research Group, 1950.

\end{thebibliography}
\bibliographystyle{icml2023}

\newpage
\appendix

\onecolumn

\section{Additional Experimental Details}
\subsection{Heatmaps for pairwise comparisons between algorithms}
\label{sec:heatmap_details}

Figures~\ref{fig:heatmap_unordered}(a) and~\ref{fig:heatmap_unordered}(b) show a comprehensive pairwise comparison of all algorithms, summarizing experiment results over all datasets ($D$), architectures ($A$), batch sizes ($k$), and total labeling budgets ($L$). Intuitively, entry $M_{i,j}$ in the matrix is the number of settings where algorithm $i$ outperforms algorithm $j$ by a statistically significant amount.
For each active learning experiment, we only consider labeling budgets $r$ where a random strategy has not yet hit 99\% of its final performance. We checkpoint the learning curves at exponentially spaced intervals until reaching this point, $L_i = N_0 + 2^{i}k \leq r$, $L = [L_0, L_1, \cdots]$, for an number of seed samples $N_0$.

For each ($D$, $A$, $k$, $L$) combination, at each we have 3 scores for algorithm $i$, and 3 scores for algorithm $j$, leading to 3 score deltas ${d_{i,j}^{1}, d_{i,j}^{2}, d_{i,j}^{3}}$ (since each experiment was repeated 3 times). We apply the two-sided $t$-test on these score deltas to decide if an algorithm significantly wins a pairwise comparison. 
Specifically we compute the t-score $t = \sqrt{3}\mu/\sigma$ as follows:
\begin{align*}
\mu = \frac{1}{3}\sum_{s=1}^{3} d_{i,j}^{s} \hspace{10mm}   \sigma = \sqrt{\frac{1}{2}\sum_{s=1}^{3}(d_{i,j}^{s} - \mu)^2}
\end{align*}
If the $t > 2.92$ algorithm $i$ is considered to outperform algorithm $j$, and vice versa if $t < -2.92$. Suppose there are $n_{D, A, B}$ labeling budgets for each (D, A, B) combination. Then when an algorithm $i$ wins a pairwise comparison with algorithm $j$, a value of $\frac{1}{|L|}$, where $|L|$ is the total number of evaluations in the experiment, is added to the corresponding entry in the matrix $M_{i,j}$.

\subsection{Details of the \texttt{CLOW} dataset}
\label{sec:clow_details}

\citet{bohus2022continual} introduced a dataset collected via a mixed-reality interactive approach for continual learning about objects in the wild (CLOW). We refer to this dataset as \texttt{CLOW} in our work. The dataset was collected at two sites -- home environments of end-users wearing a mixed-reality headset as they go about performing everyday tasks. We use data from Site 2 as reported in ~\cite{bohus2022continual}. The mixed-reality multimodal interface enables users to label objects in their surroundings via their gaze, speech, and gestures. The system uses color and depth cameras to detect and track objects in the scene. Different views of an object are captured by the device as the user interacts with and labels it. The stream of images thus naturally exhibits correlation due to the notion of object permanence in the environment across time. The labels from the user are linked to multiple object views collected over time, and can be used to improve object recognition over time. While the object label collection occurs via a user-initiated approach, the authors of this work identify the need for an active learning solution to minimize the labeling burden on a user while an object recognition model continues to learn about objects in the environment.

\begin{figure}
    \centering   
\includegraphics[width=0.45\textwidth]{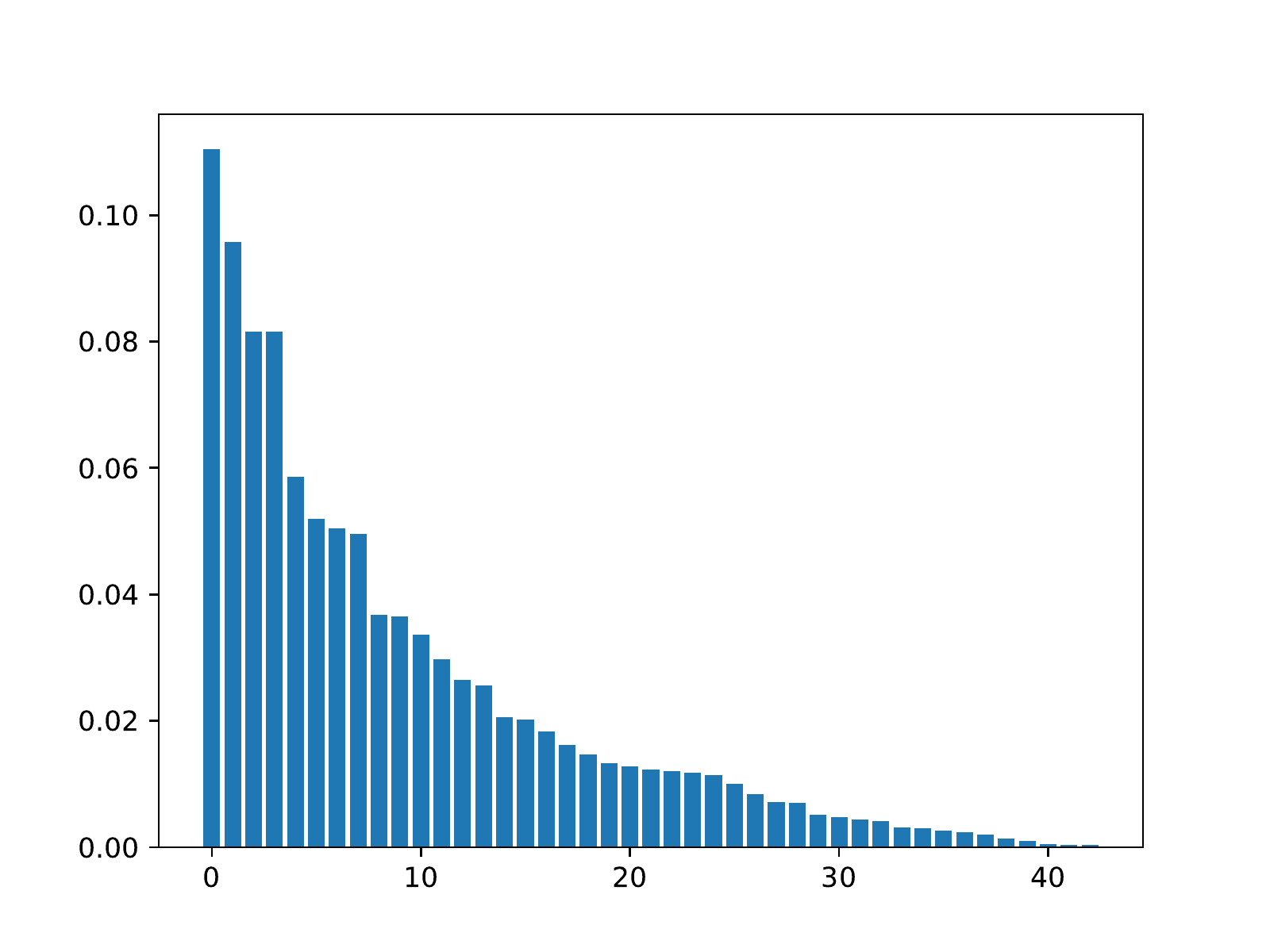}
    \caption{Distribution of 43 class labels from the Site2 \texttt{CLOW} dataset. The y-axis denoted the fraction of images that belong to a specific class label (class ID denoted by the indices of the x-axis).
    }
    \label{fig:hololens_class_dist}
\end{figure}

\textbf{Preprocessing the \texttt{CLOW} dataset:} \texttt{CLOW} is collected at two home environments (Site1 and Site 2)~\cite{bohus2022continual}, with streaming images of objects being recorded at 5Hz. We use data from Site 2 which consists of a total of 47 object classes and 55,657 object images. ~\citet{bohus2022continual} filter this dataset to remove images with blur and occlusion via hand-defined thresholds on linear and angular speed of the headset as well as the overlap of human hands with object views. The filtered version of their dataset at Site 2 consists of 15,095 images from 47 object classes. We further filter the data stream from Site 2 by removing images with an interaction of less than 5 frames (1 second), i.e. if an object appears for less than 5 consecutive frames in the image stream, then we discard images from such a short interaction. This provides us a total of 14,981 images from 43 object classes. We split this data stream into train and test sets based on the following criteria: for each user interaction with an object instance, the first 80\% are used as part of the train set and the last 20\% of the interaction is used as part of the test set. This provides us with 11,899 training images and 3082 test images. This strategy ensures that we have 20\% of the data for each object in the stream as part of the evaluation set. The distribution of the 11,899 training images into different object classes is shown in Fig~\ref{fig:hololens_class_dist}. We downsample each image to size $32\times32\times3$ before passing it as an input to the CLIP visual encoder model~\cite{radford2021learning} or the ImageNet pretrained ResNet-18 model~\cite{russakovsky2015imagenet}. To efficiently compute the covariance matrix as part of VeSSAL for this dataset with 43 object classes, we employ random projection~\cite{bingham2001random} to reduce the dimensionality of the gradient embeddings to size 2560.

\section{Learning curves for I.I.D. Data Stream Experiments}
\label{sec:lr_iid}
\begin{figure*}[!htb]
    \centering
    \includegraphics[width=0.9\textwidth]{imgs/LR_legend_1.pdf}
    \subfigure[MLP]{
    \includegraphics[width=0.3\textwidth]{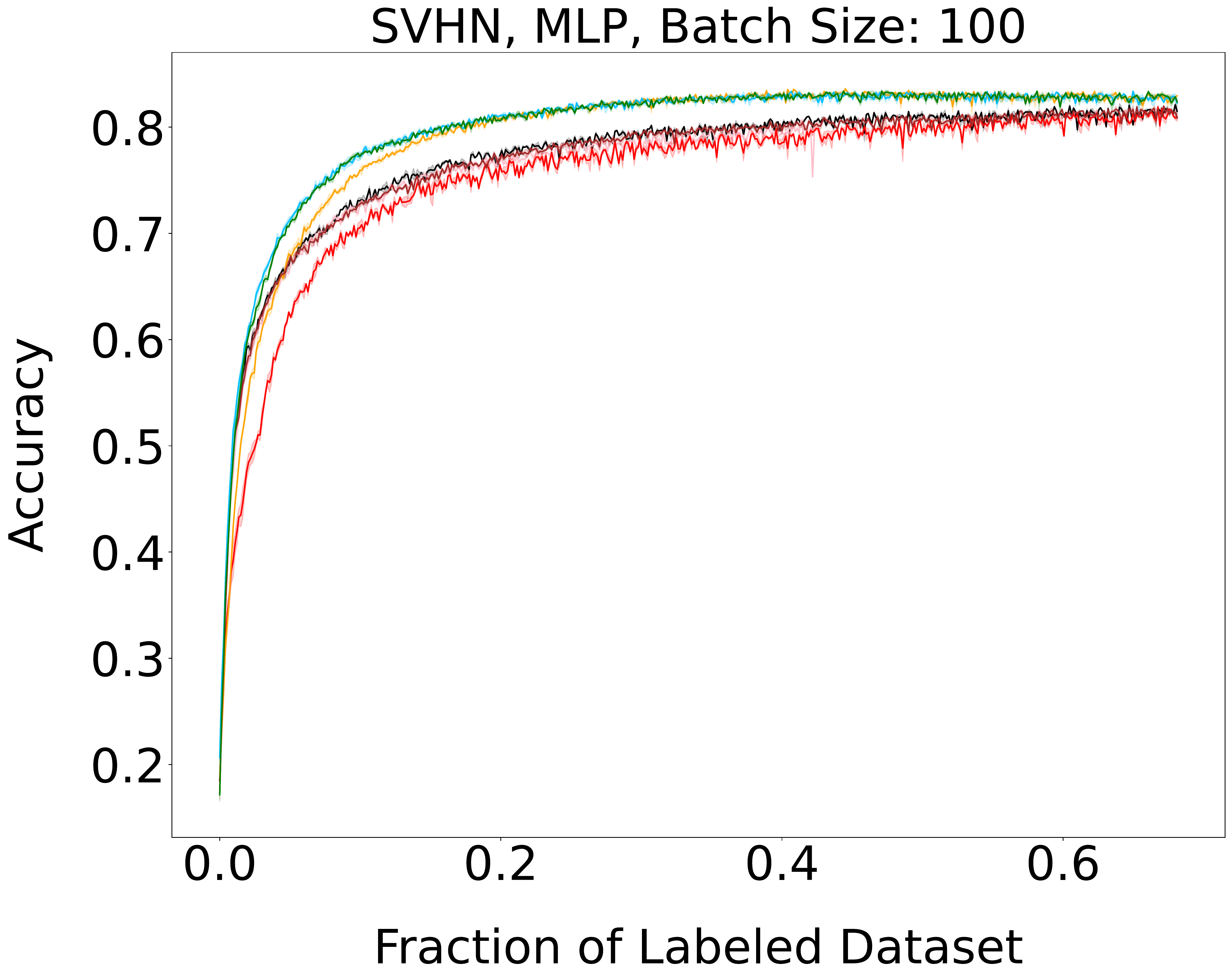}
    \includegraphics[width=0.3\textwidth]{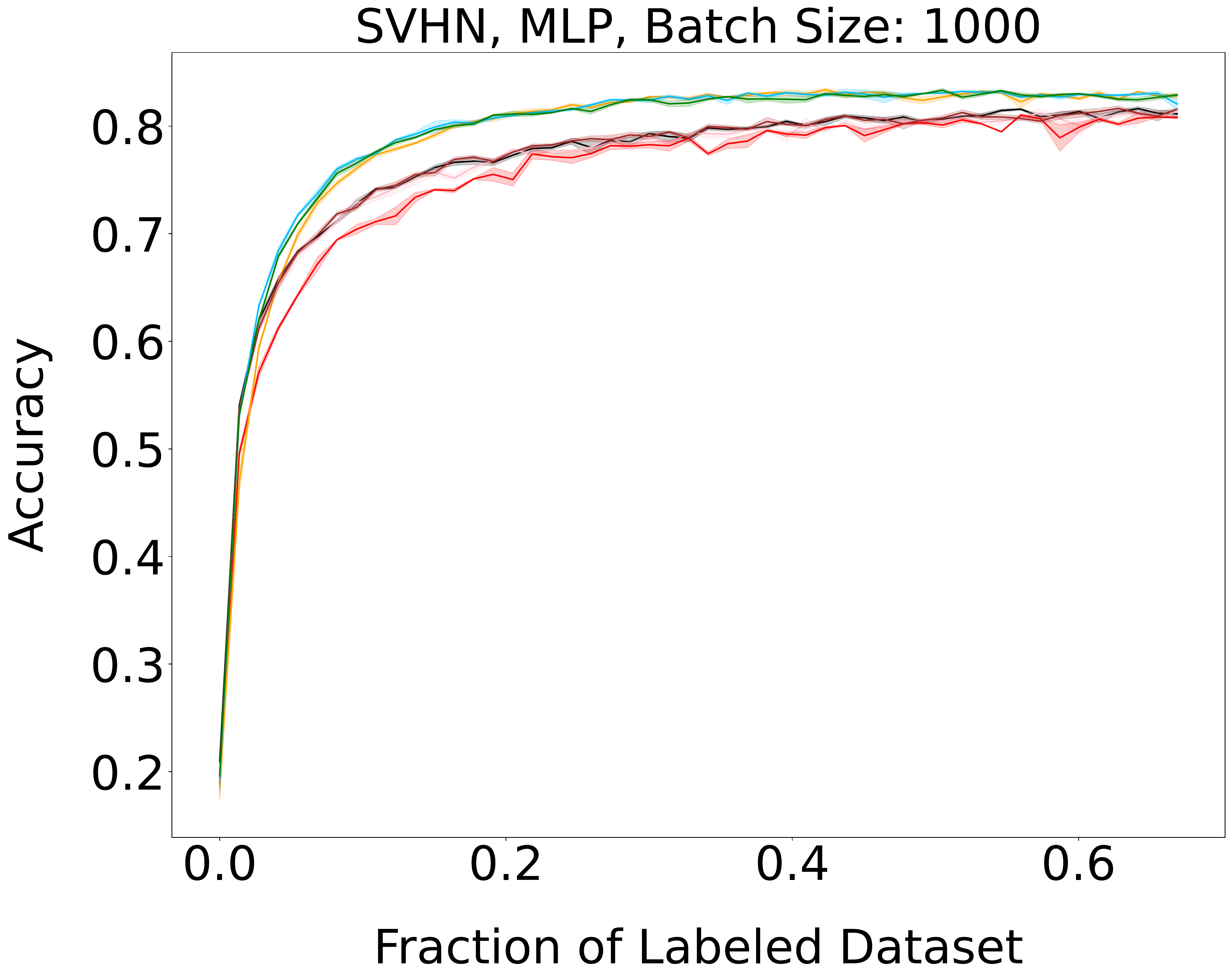}
    \includegraphics[width=0.3\textwidth]{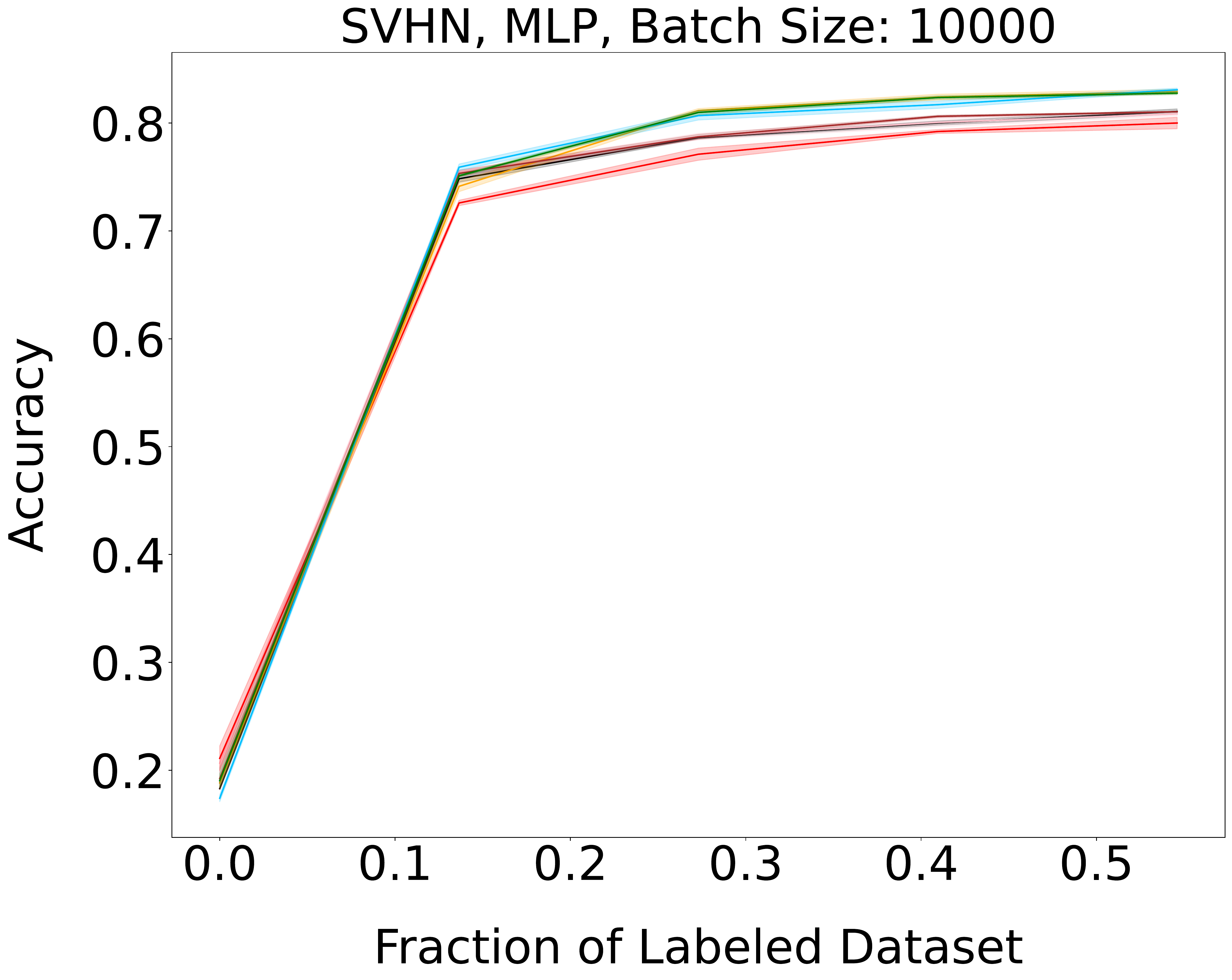}
    }
    \subfigure[ResNet]{
    \includegraphics[width=0.3\textwidth]{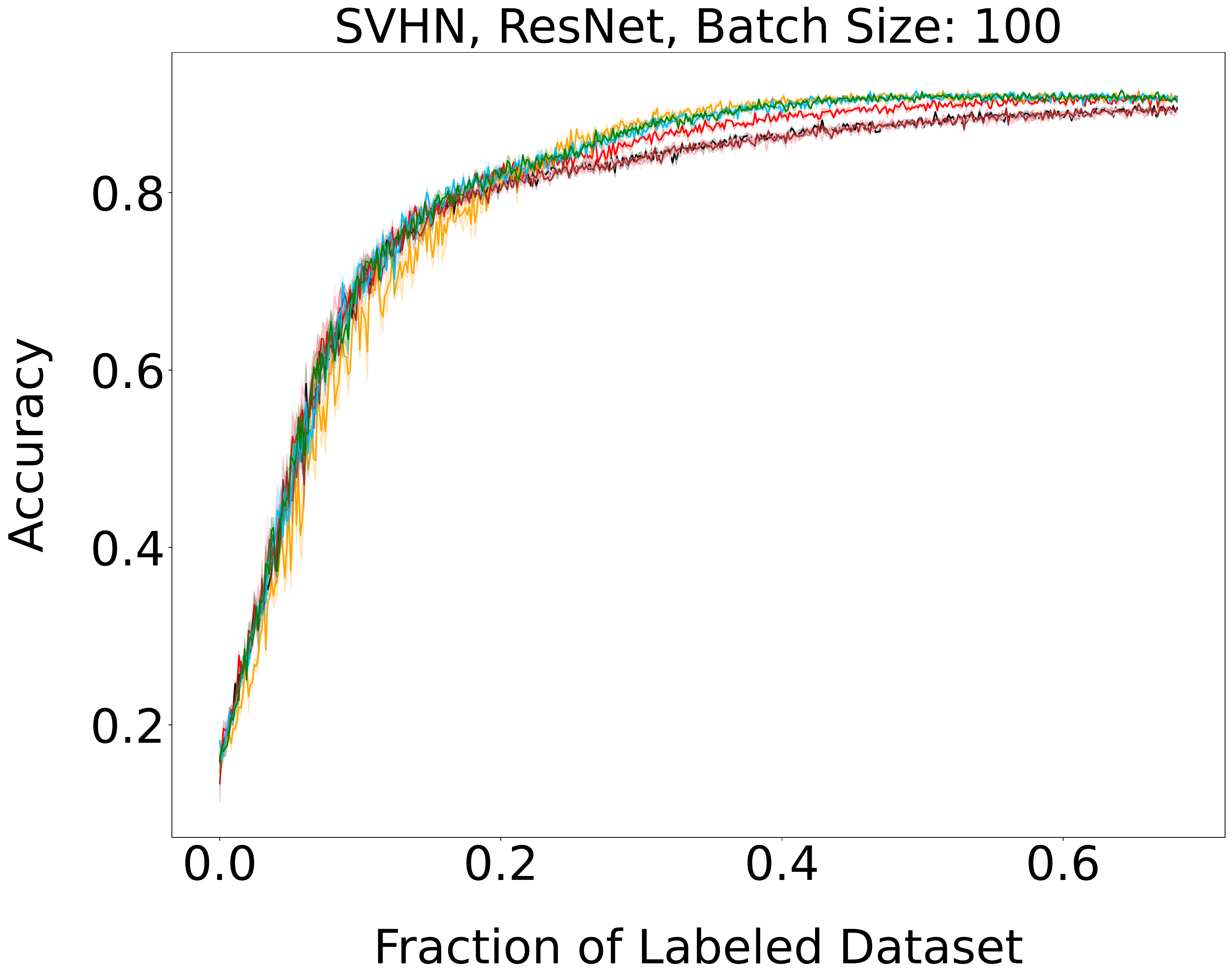}
    \includegraphics[width=0.3\textwidth]{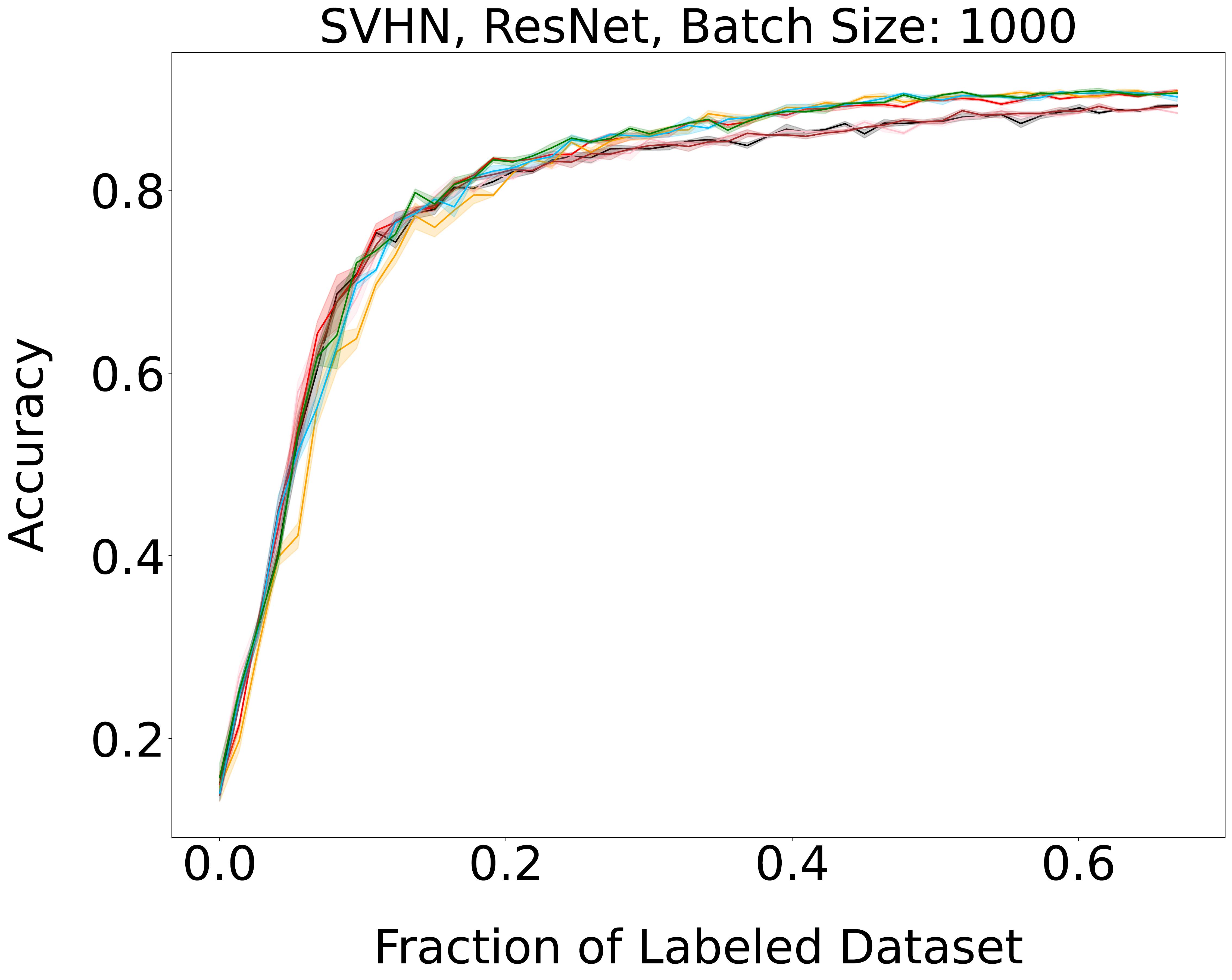}
    \includegraphics[width=0.3\textwidth]{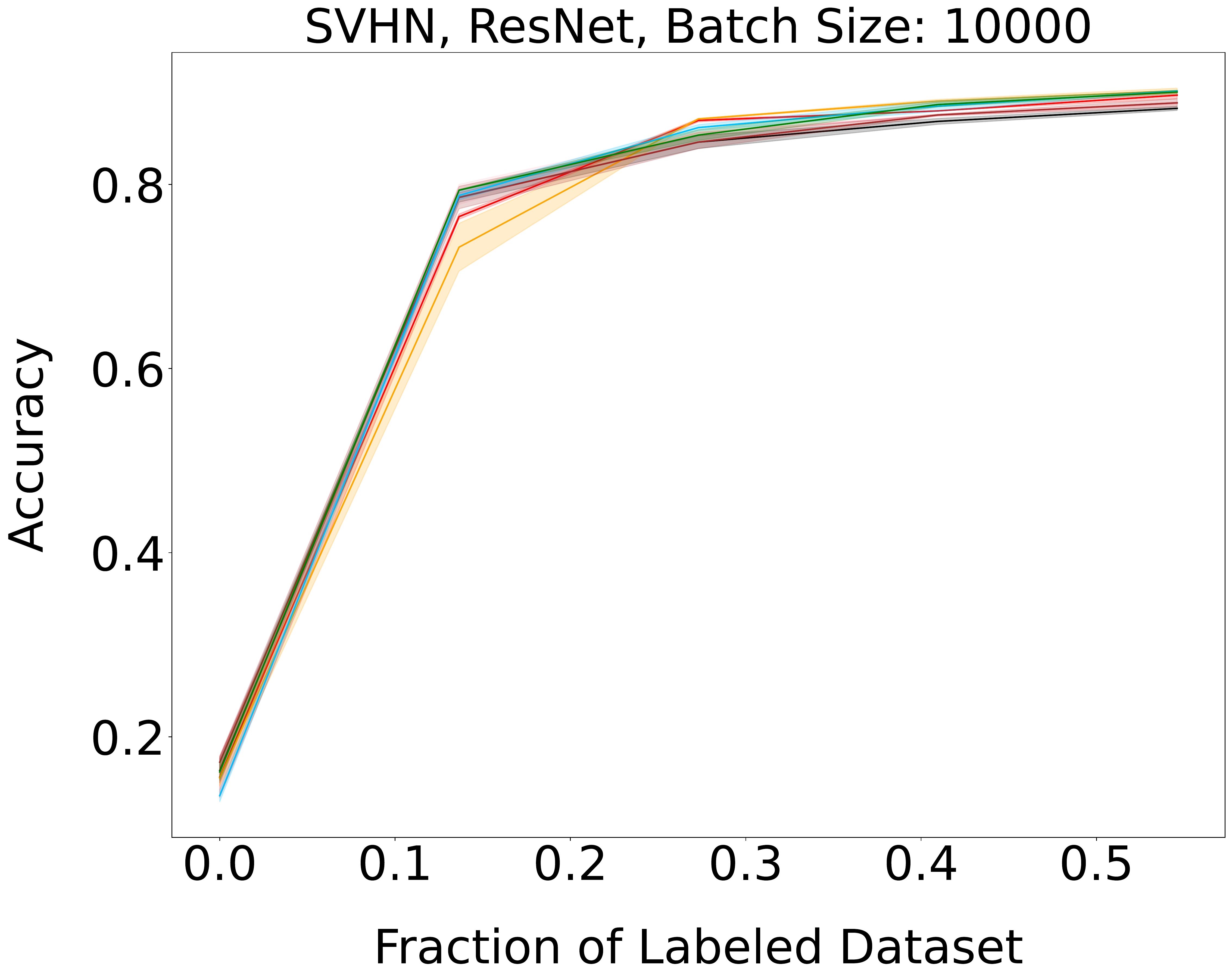}
    }
    \caption{Learning curves for the \texttt{SVHN} dataset with two network architectures and three batch sizes. Streaming active learning algorithms observe data in a randomized order.}
    \label{fig:svhn}
\end{figure*}

\begin{figure*}
    \centering
    \includegraphics[width=0.9\textwidth]{imgs/LR_legend_1.pdf}
    \subfigure[MLP]{
    \includegraphics[width=0.3\textwidth]{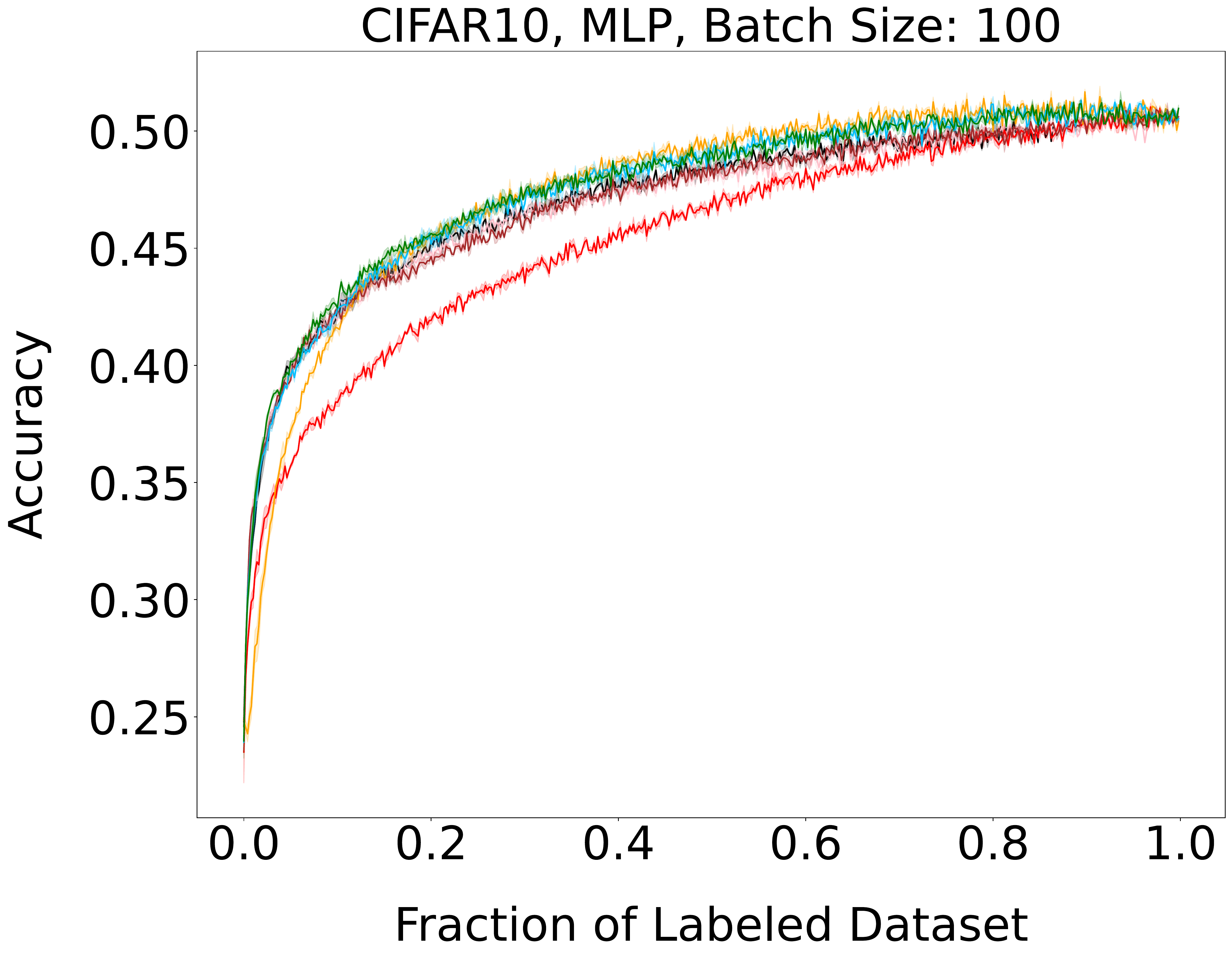}
    \includegraphics[width=0.3\textwidth]{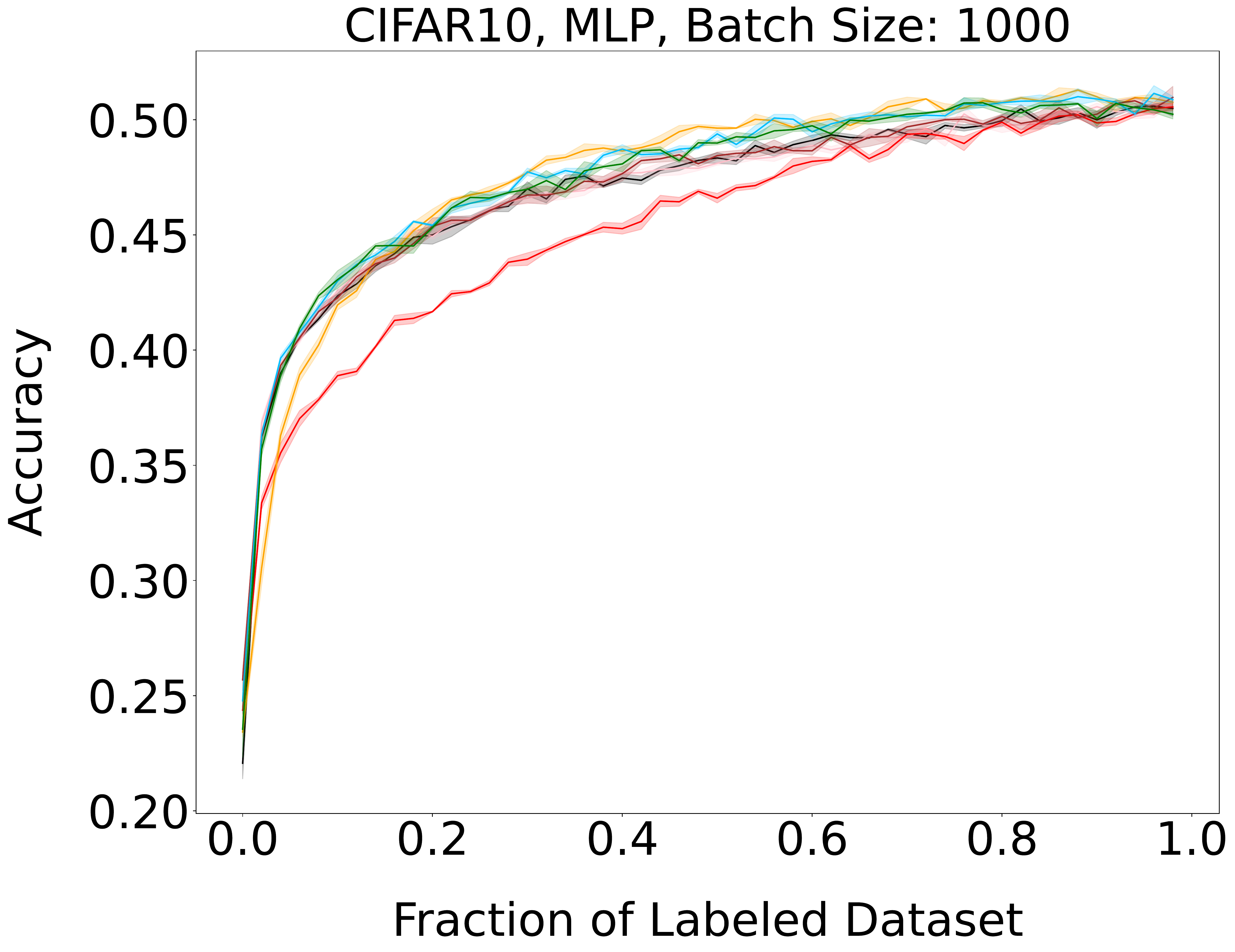}
    \includegraphics[width=0.3\textwidth]{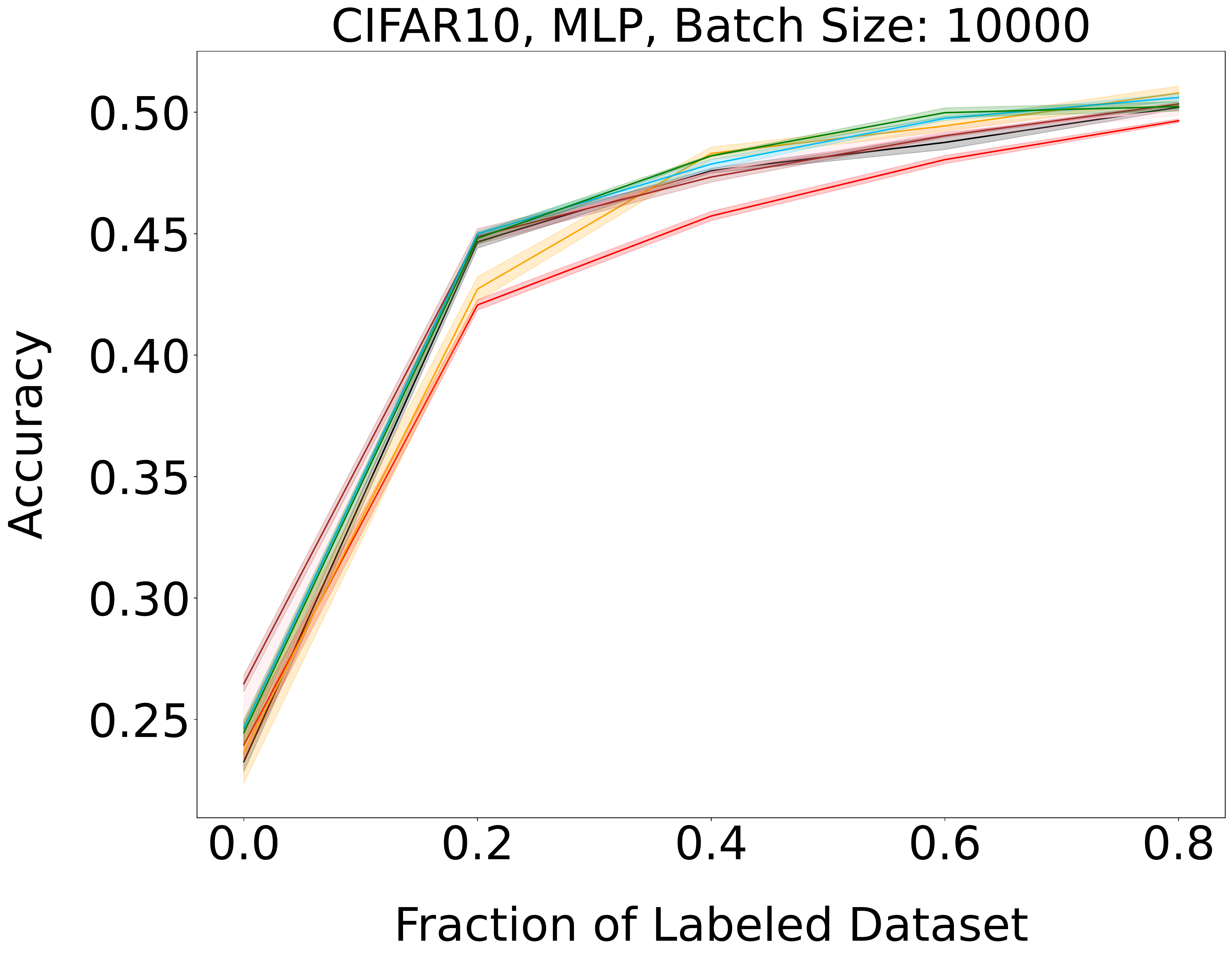}
    }
    \subfigure[ResNet]{
    \includegraphics[width=0.3\textwidth]{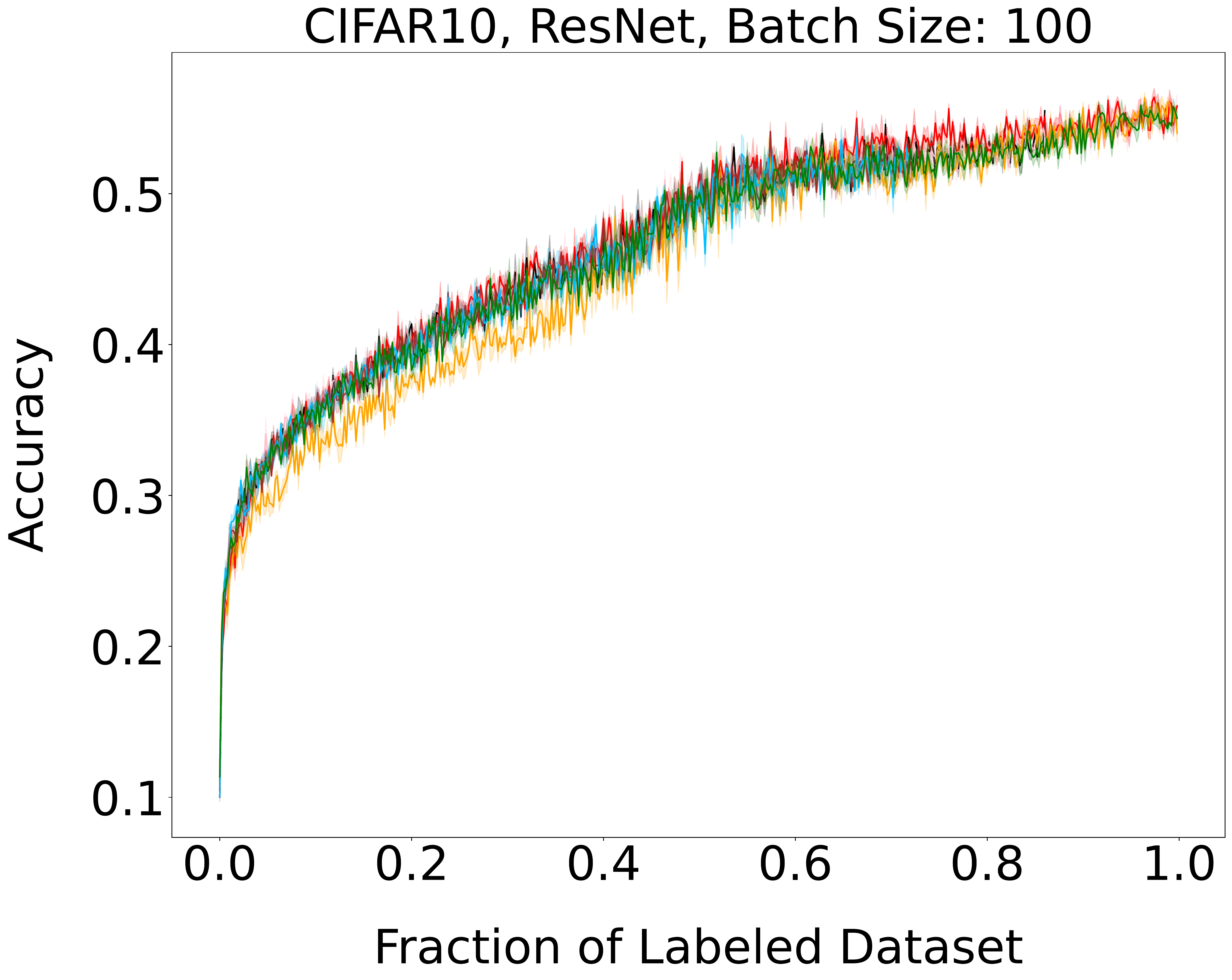}
    \includegraphics[width=0.3\textwidth]{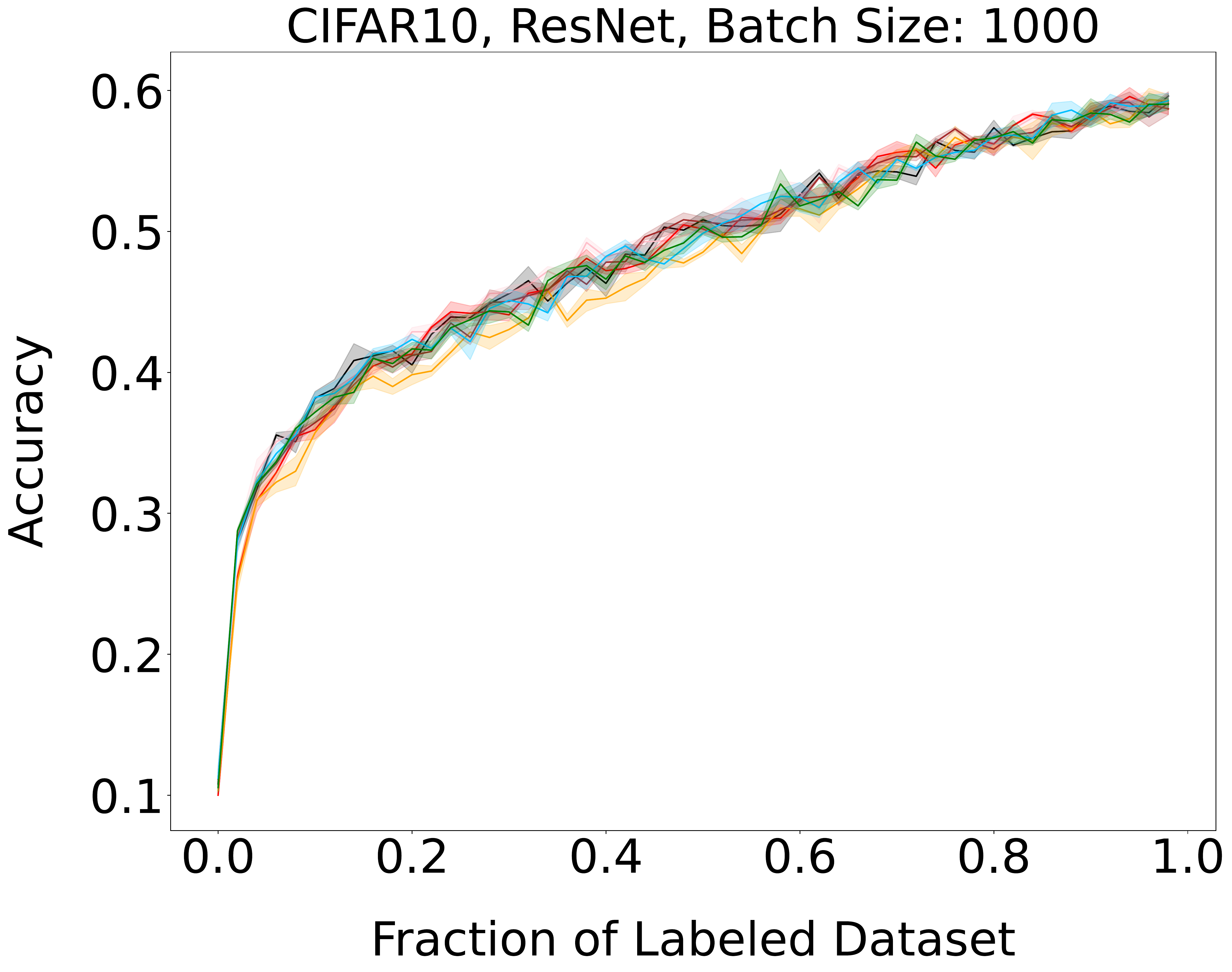}
    \includegraphics[width=0.3\textwidth]{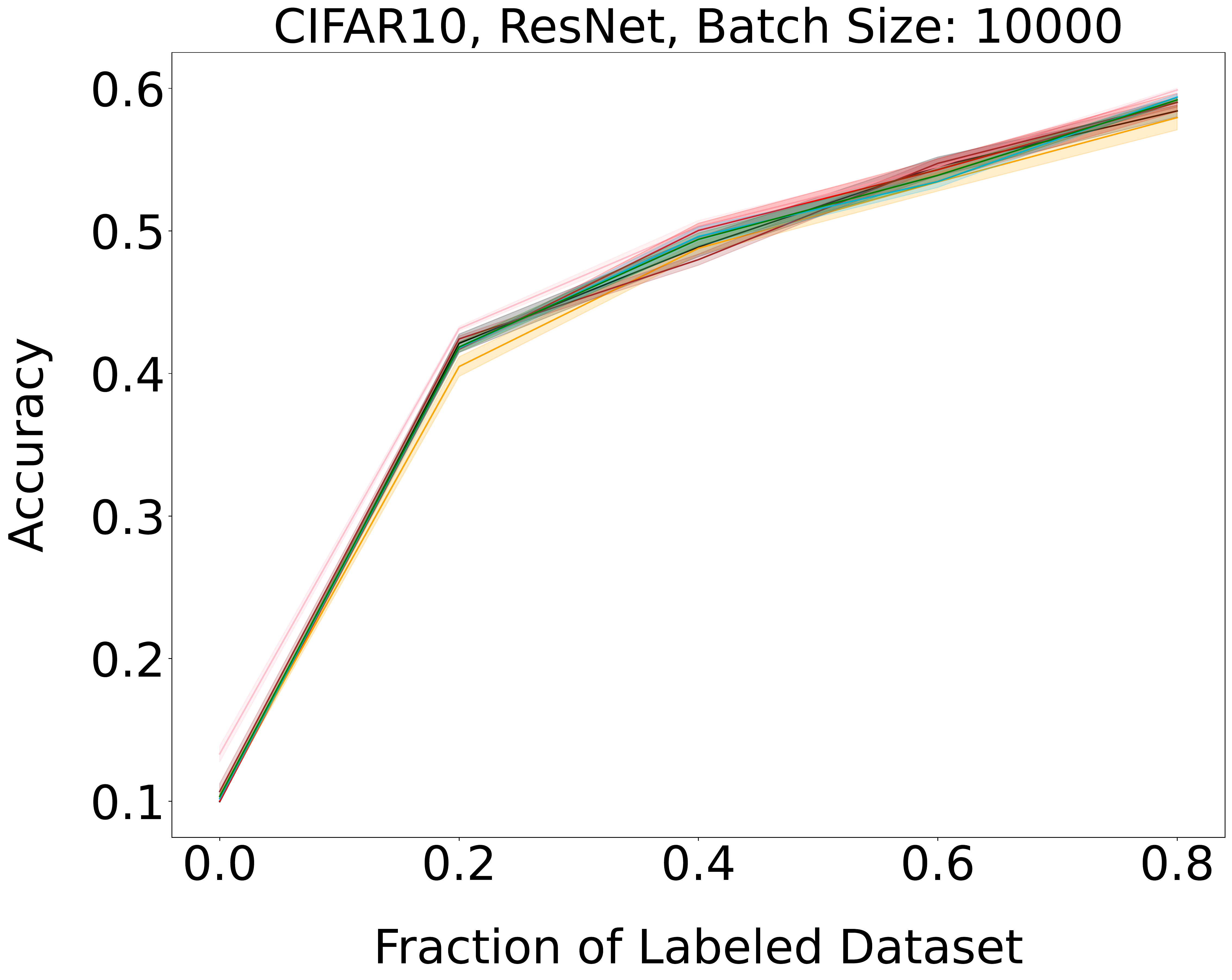}
    }
    \subfigure[ResNet w/ Data Augmentation]{
    \includegraphics[width=0.3\textwidth]{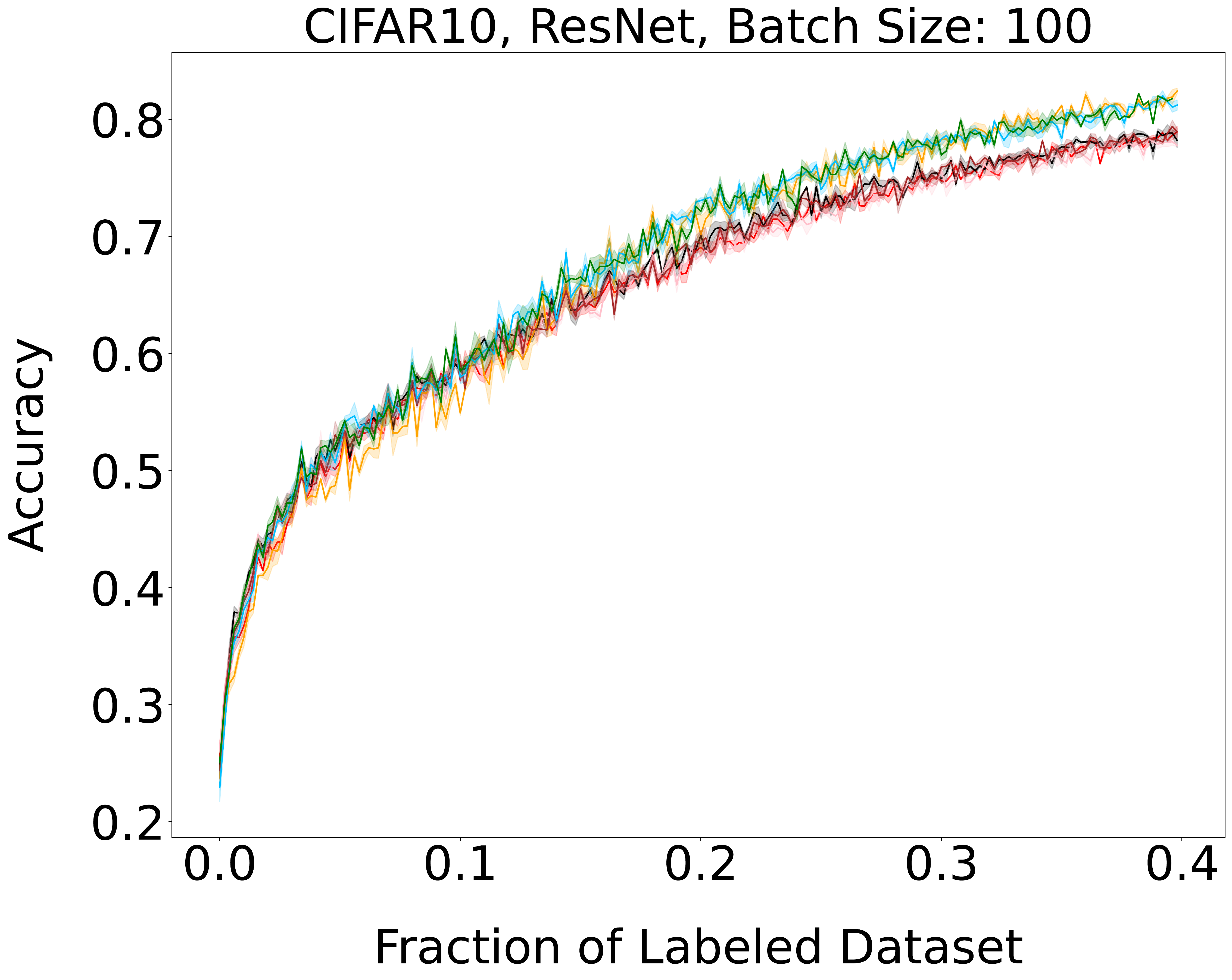}
    \includegraphics[width=0.3\textwidth]{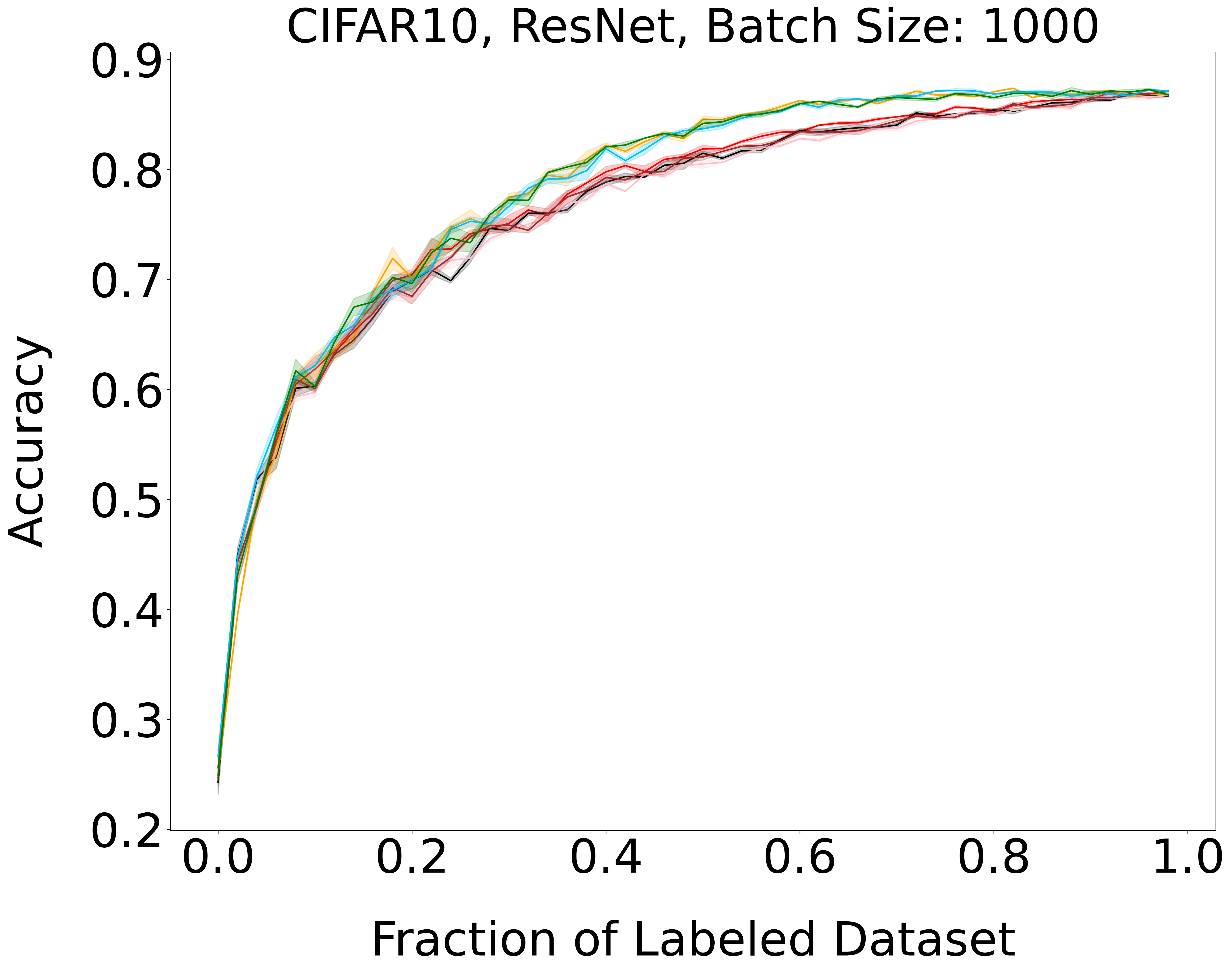}
    \includegraphics[width=0.3\textwidth]{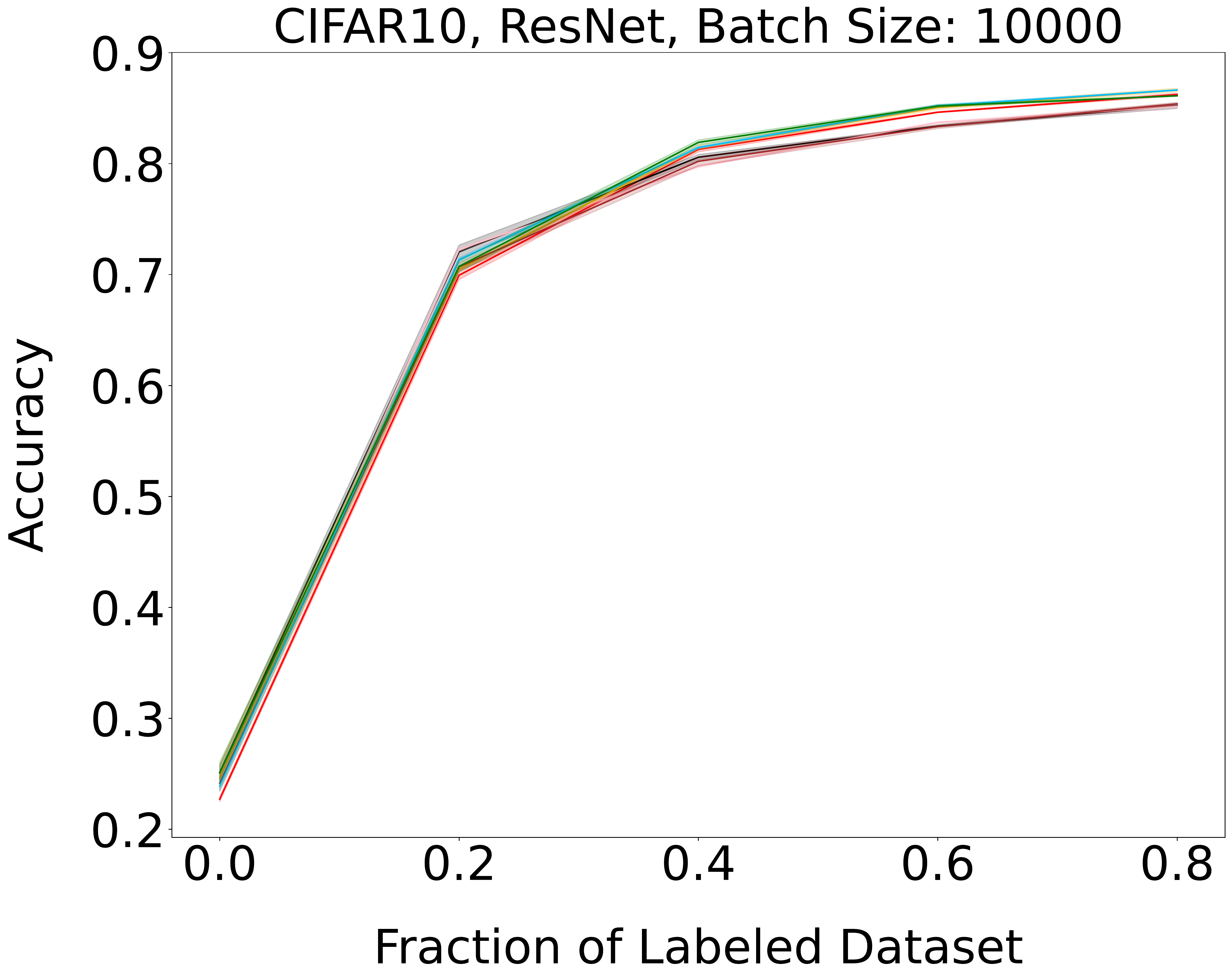}
    }
    \caption{Learning curves for the \texttt{CIFAR-10} dataset with two network architectures (MLP, ResNet) and three batch sizes (100, 1000, 10000). We show results for both ResNet training (b) without data augmentation (similar to \citet{ash2019deep}) and (c) with data augmentation. Streaming active learning algorithms observe data in a randomized order.}
    \label{fig:cifar}
\end{figure*}

\begin{figure*}
    \centering
    \includegraphics[width=0.9\textwidth]{imgs/LR_legend_1.pdf}
\includegraphics[width=0.3\textwidth]{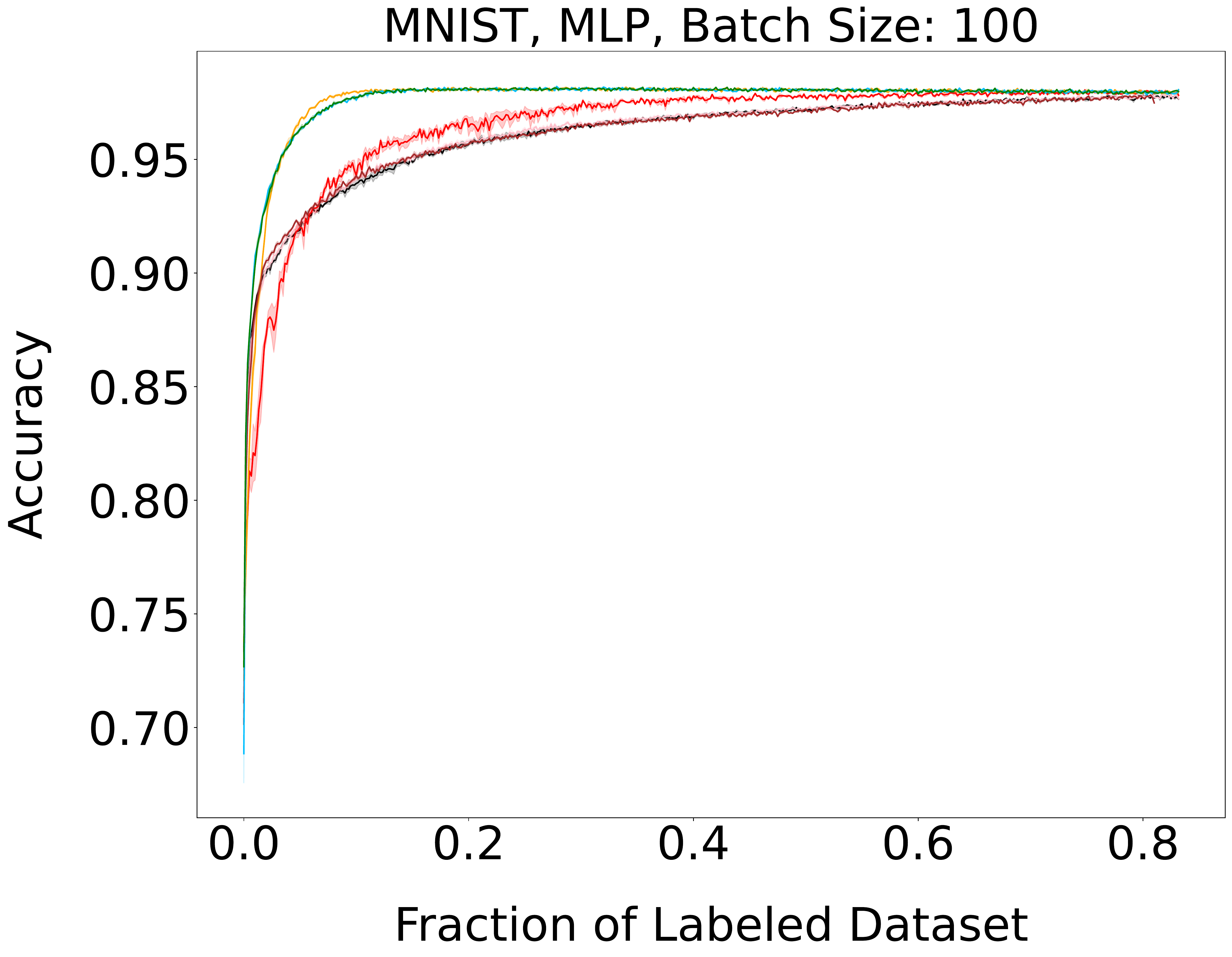}
\includegraphics[width=0.3\textwidth]{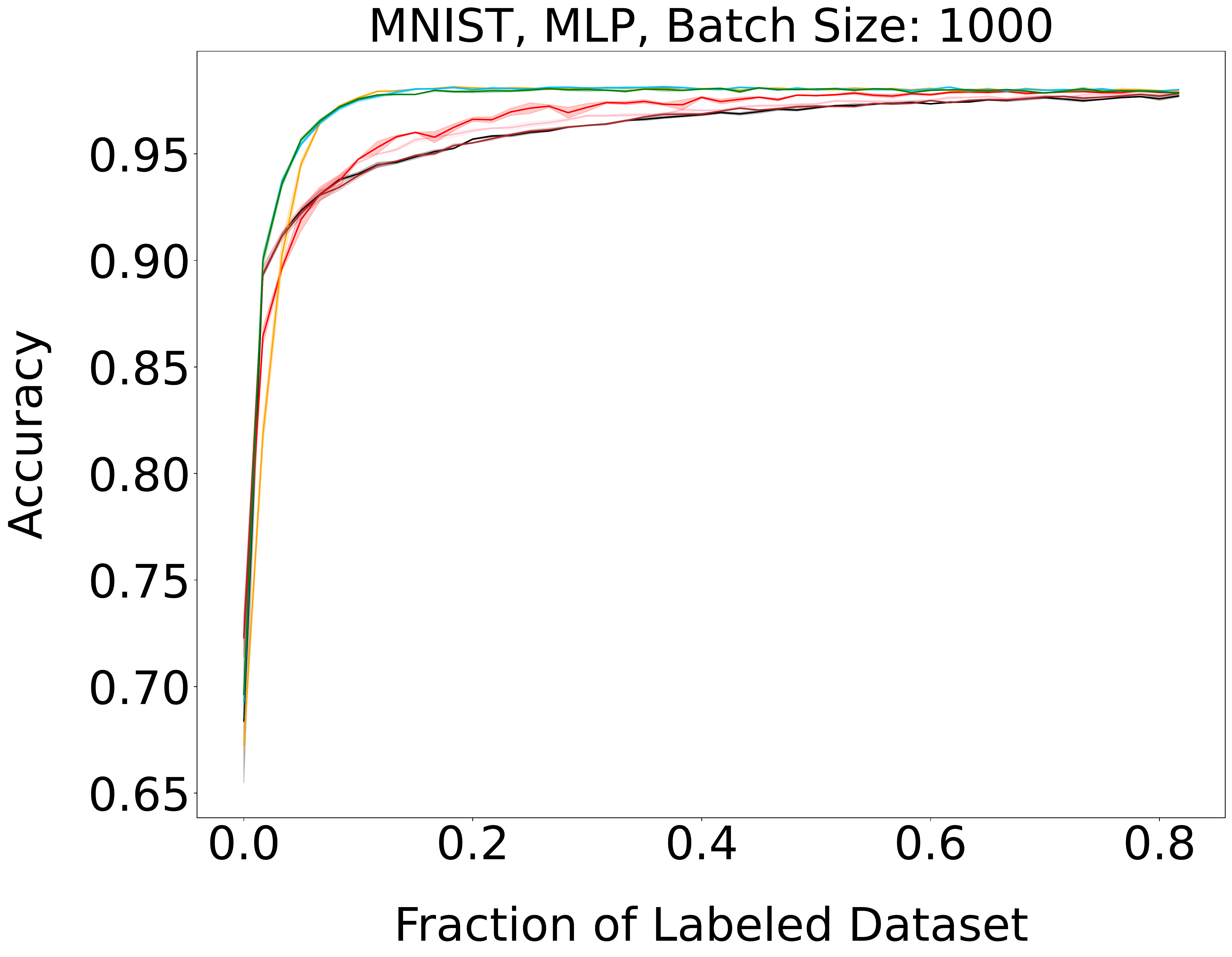}
\includegraphics[width=0.3\textwidth]{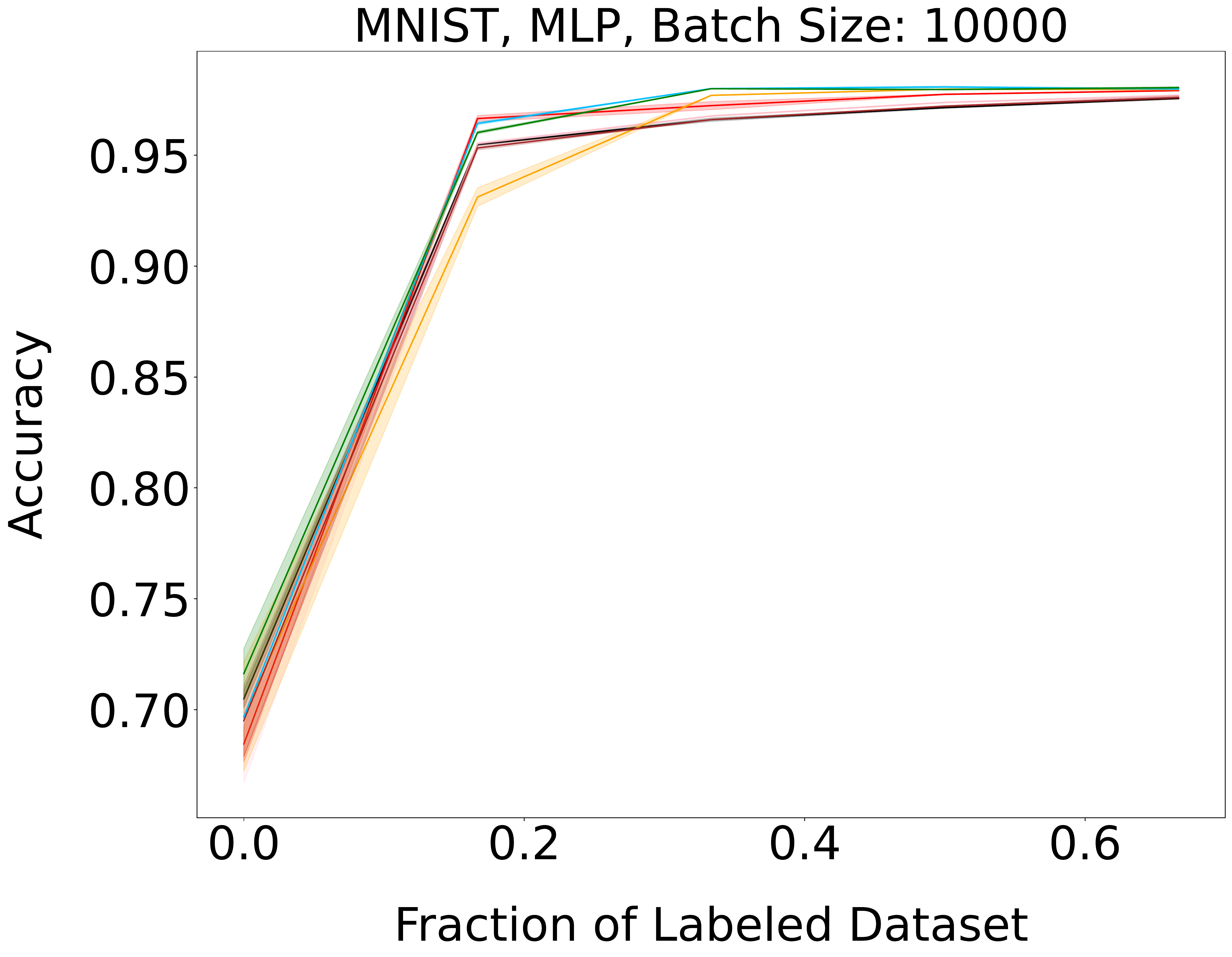}
    \caption{Learning curves for the \texttt{MNIST} dataset with the MLP network architecture and three batch sizes. Streaming active learning algorithms observe data in a randomized order.}
    \label{fig:mnist}
\end{figure*}

We show learning curves below for all experiments with randomized data streams from the following datasets: \texttt{SVHN} (Fig.~\ref{fig:svhn}), \texttt{CIFAR-10} (Fig.~\ref{fig:cifar}), \texttt{MNIST} (Fig.~\ref{fig:mnist})).

\clearpage
\section{Learning curves for non-I.I.D. Data Stream Experiments}
\label{sec:lr_non_iid}

\begin{figure*}[!htb]
    \centering
    \includegraphics[width=0.9\textwidth]{imgs/LR_legend_1.pdf}
    \includegraphics[width=0.3\textwidth]{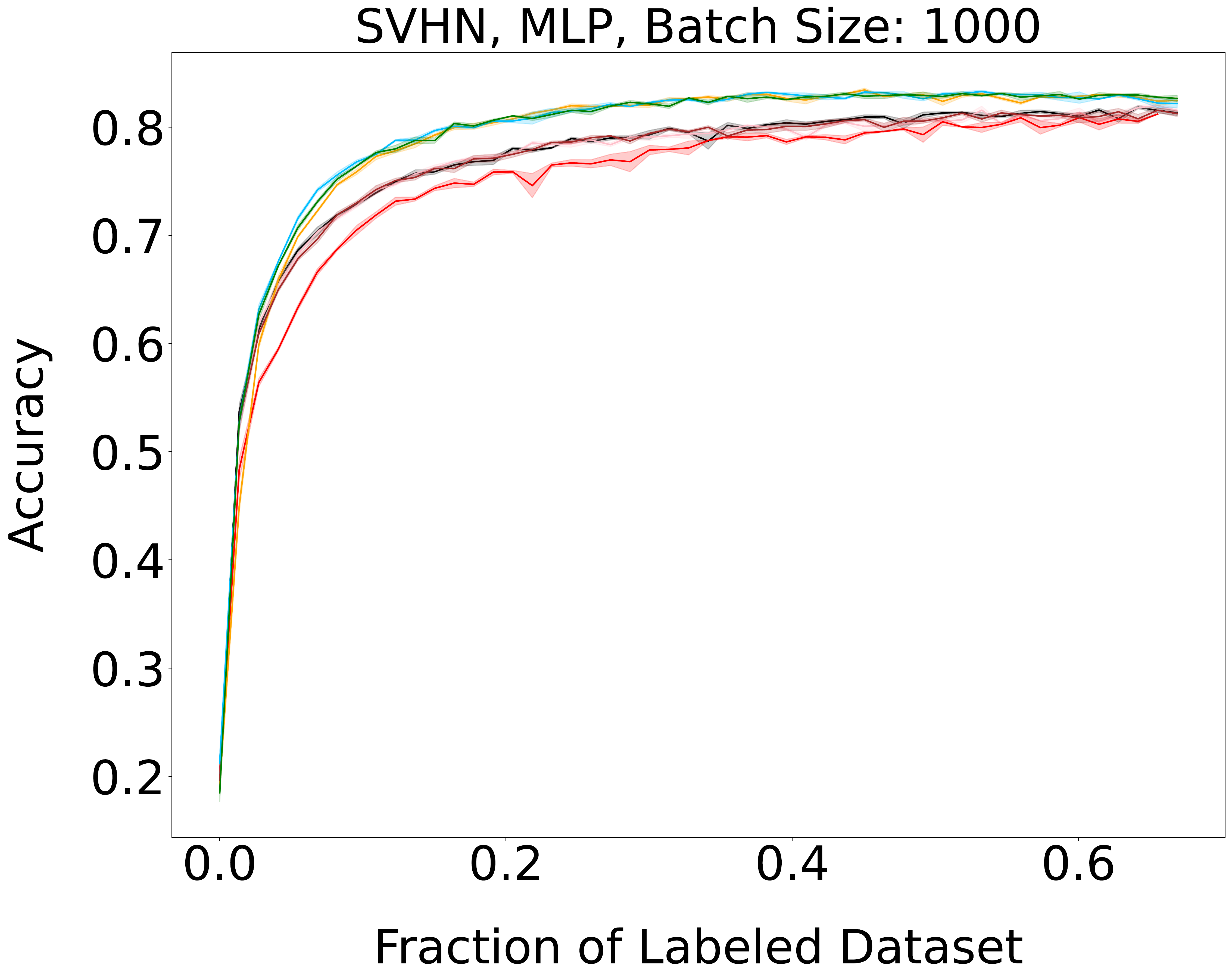}
    \includegraphics[width=0.3\textwidth]{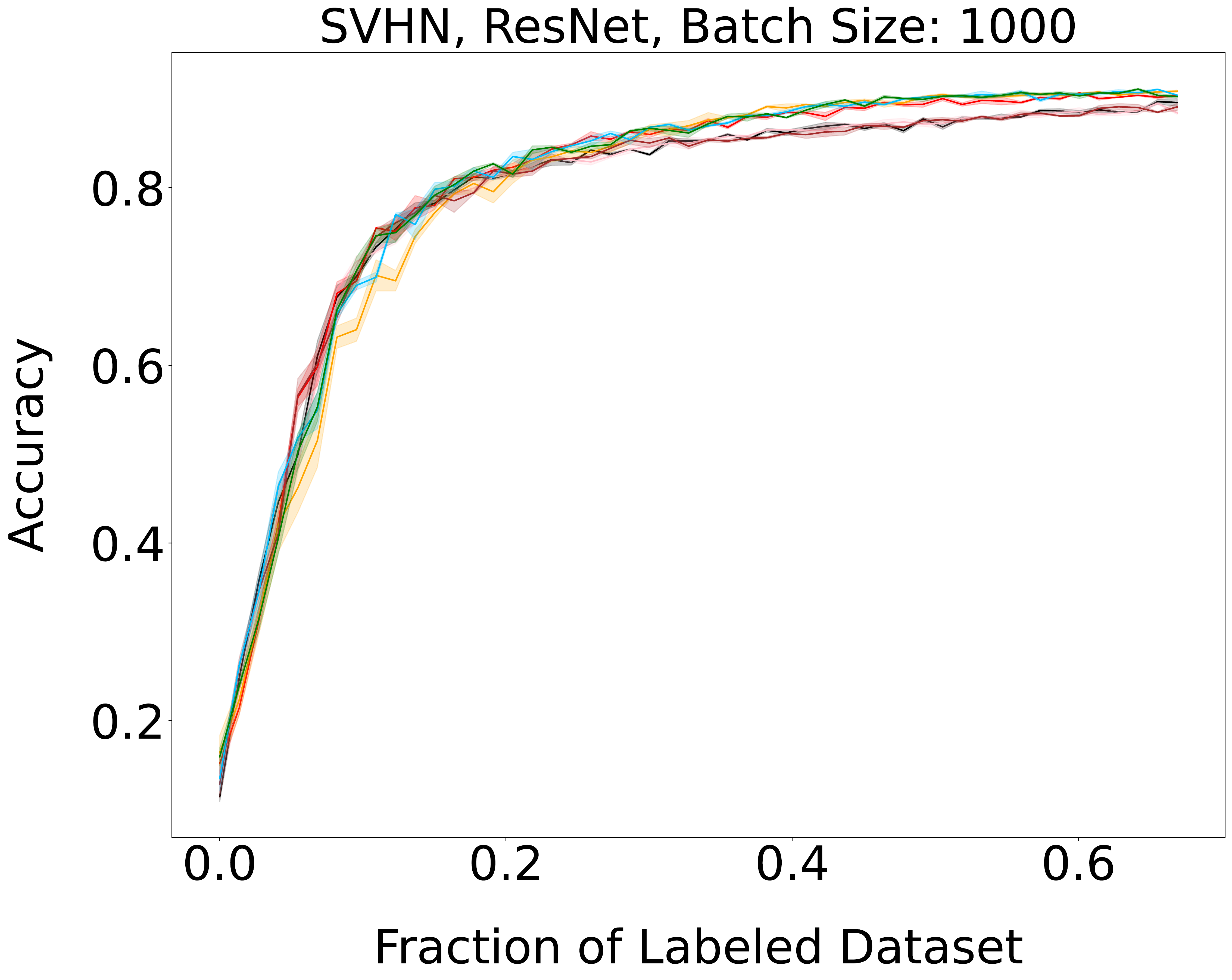}
    \caption{Learning curves for the \texttt{SVHN} dataset with two network architectures and batch size 1000. Streaming active learning algorithms observe data sorted by their 1st principal component.}
    \vspace{2cm}
    \label{fig:svhn_1000_sorted}
\end{figure*}

\begin{figure*}[!htb]
    \centering
    \includegraphics[width=0.9\textwidth]{imgs/LR_legend_1.pdf}
    \subfigure[MLP]{
    \includegraphics[width=0.3\textwidth]{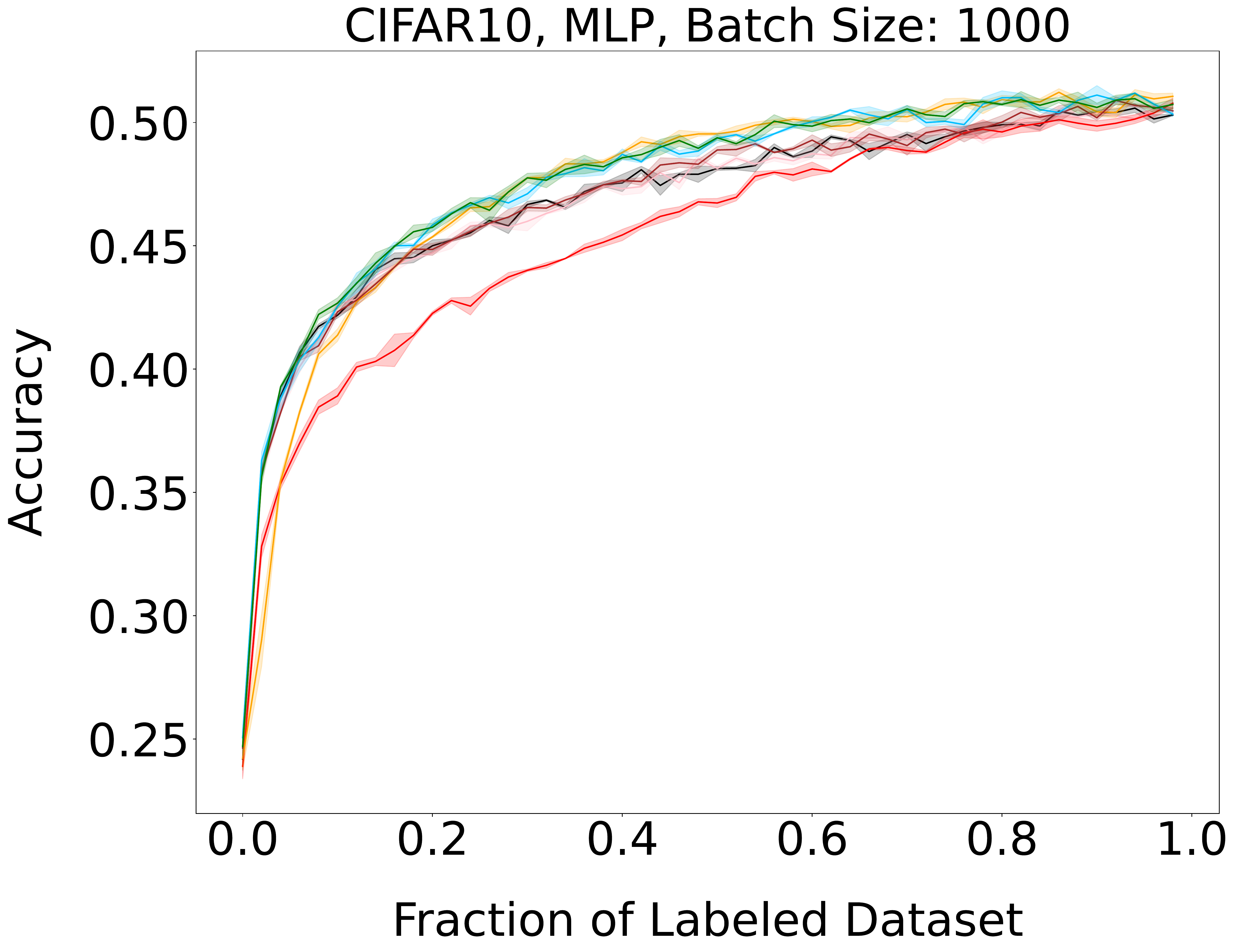}
    }
    \subfigure[ResNet]{
    \includegraphics[width=0.3\textwidth]{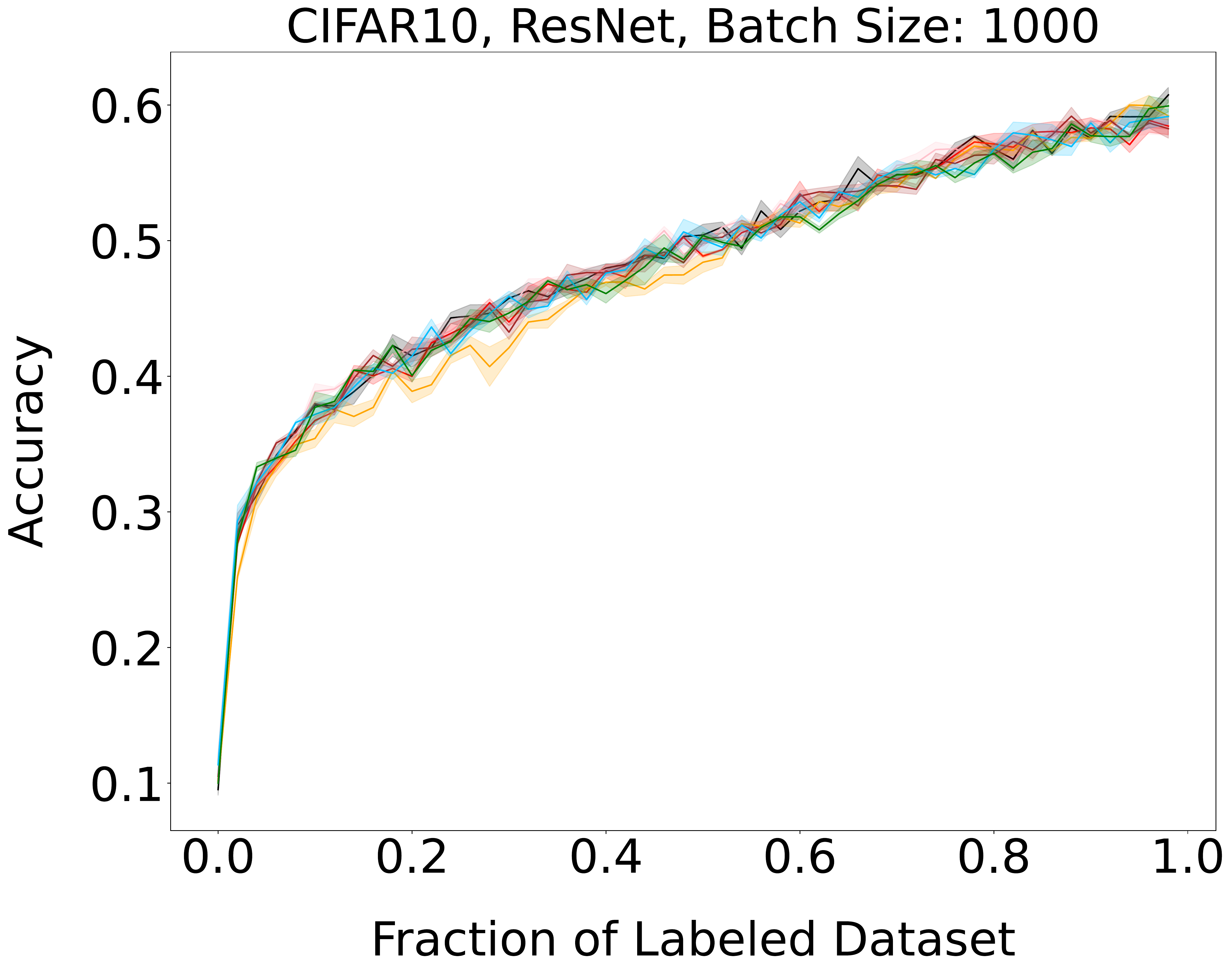}
    }
    \subfigure[ResNet w/ Data Augmentation]{
    \includegraphics[width=0.3\textwidth]{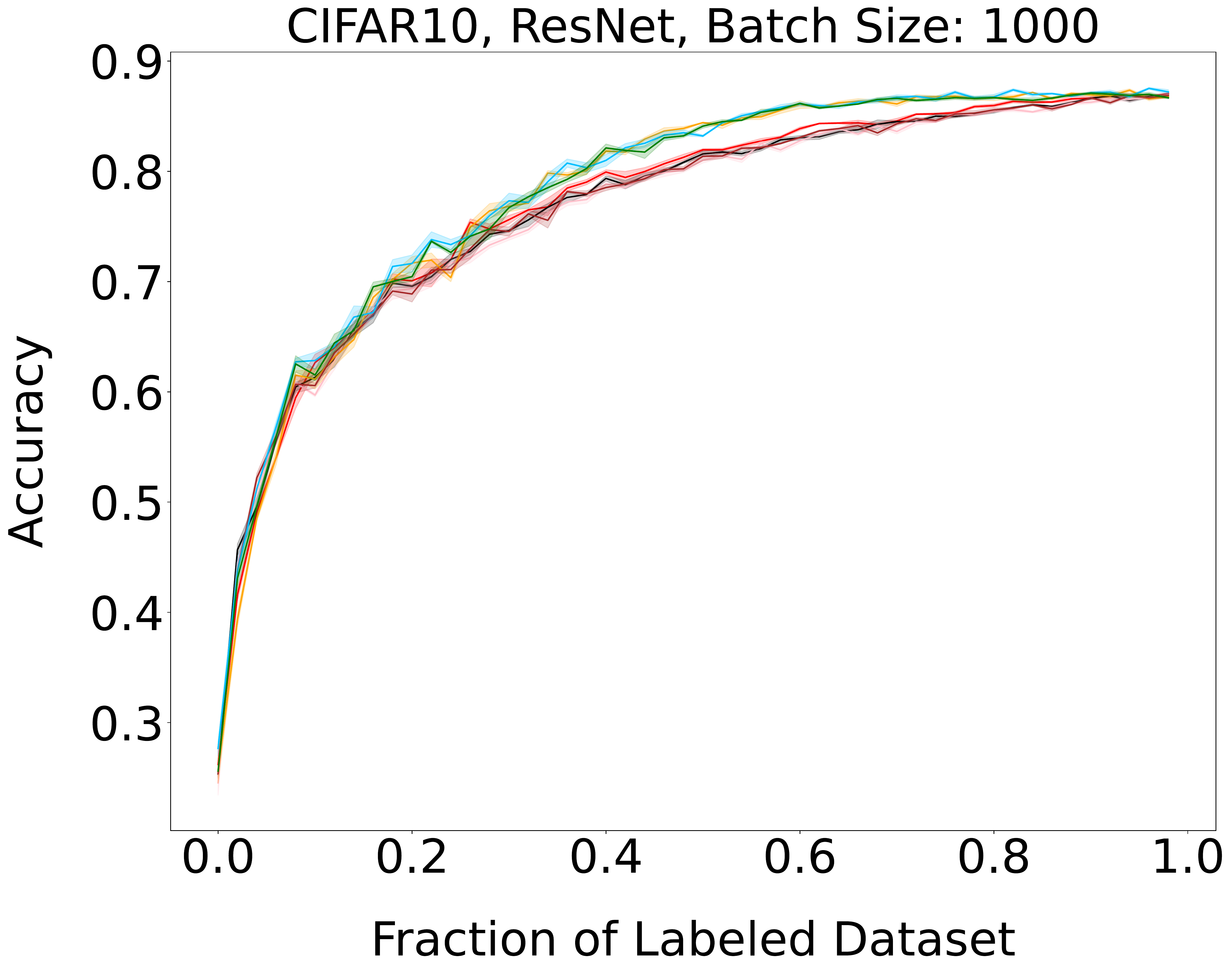}
    }
    \caption{Learning curves for the \texttt{CIFAR-10} dataset with two network architectures and batch size 1000. Streaming active learning algorithms observe data sorted by their 1st principal component.}
    \label{fig:cifar_1000_sorted}
\end{figure*}

\begin{figure*}[!htb]
    \centering
    \includegraphics[width=0.9\textwidth]{imgs/LR_legend_1.pdf}
      \subfigure[MLP]{
\includegraphics[width=0.3\textwidth]{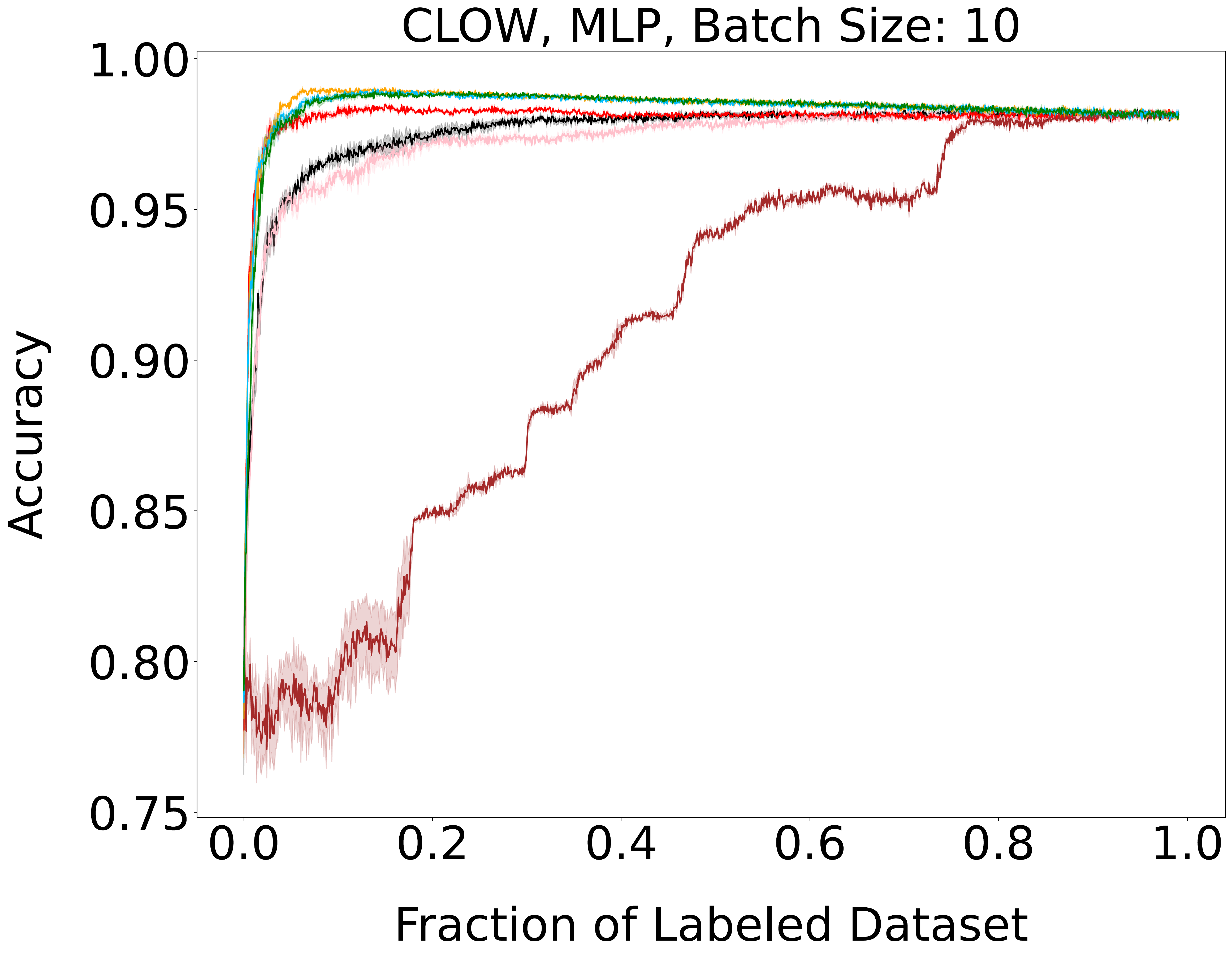}
\includegraphics[width=0.3\textwidth]{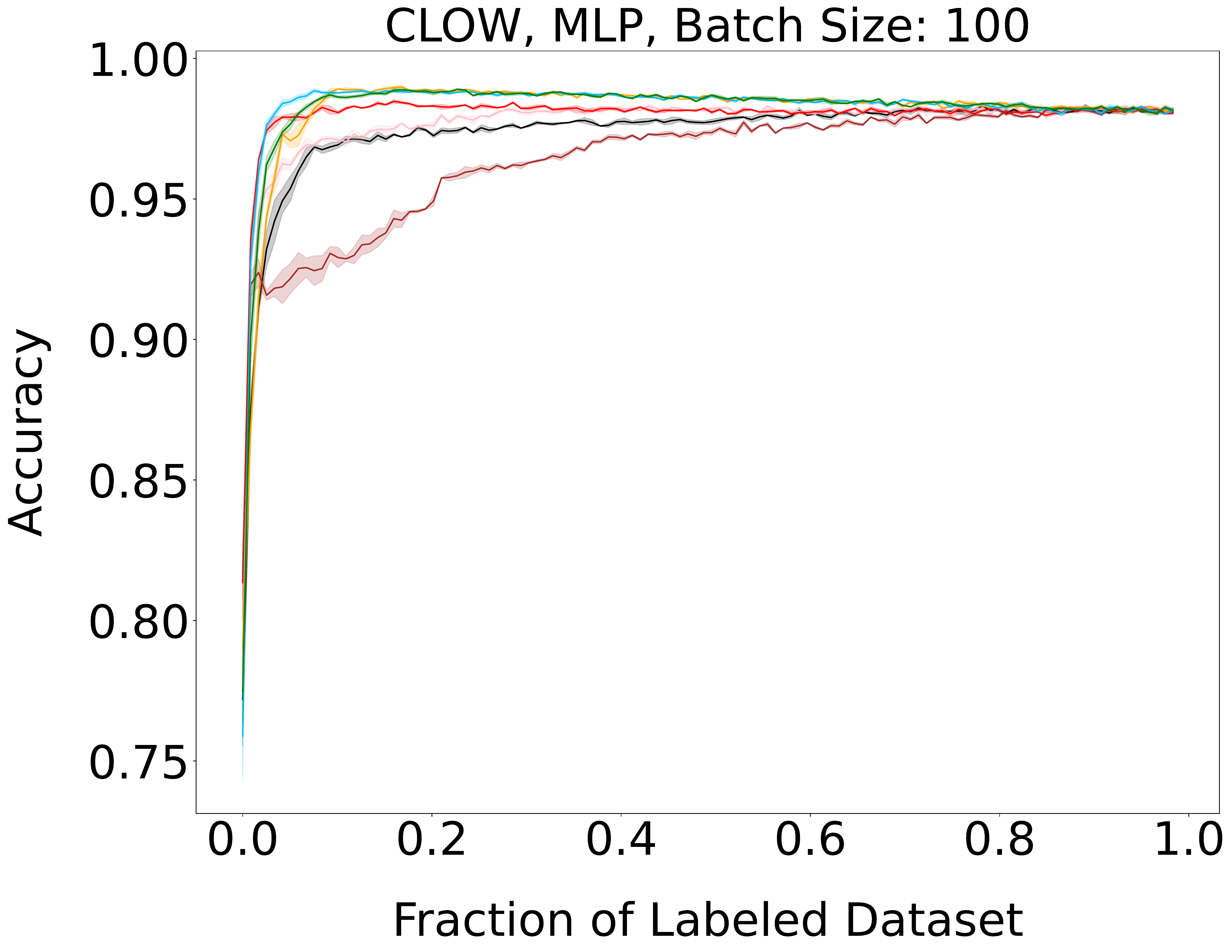}
    }
    \subfigure[ResNet]{
\includegraphics[width=0.3\textwidth]{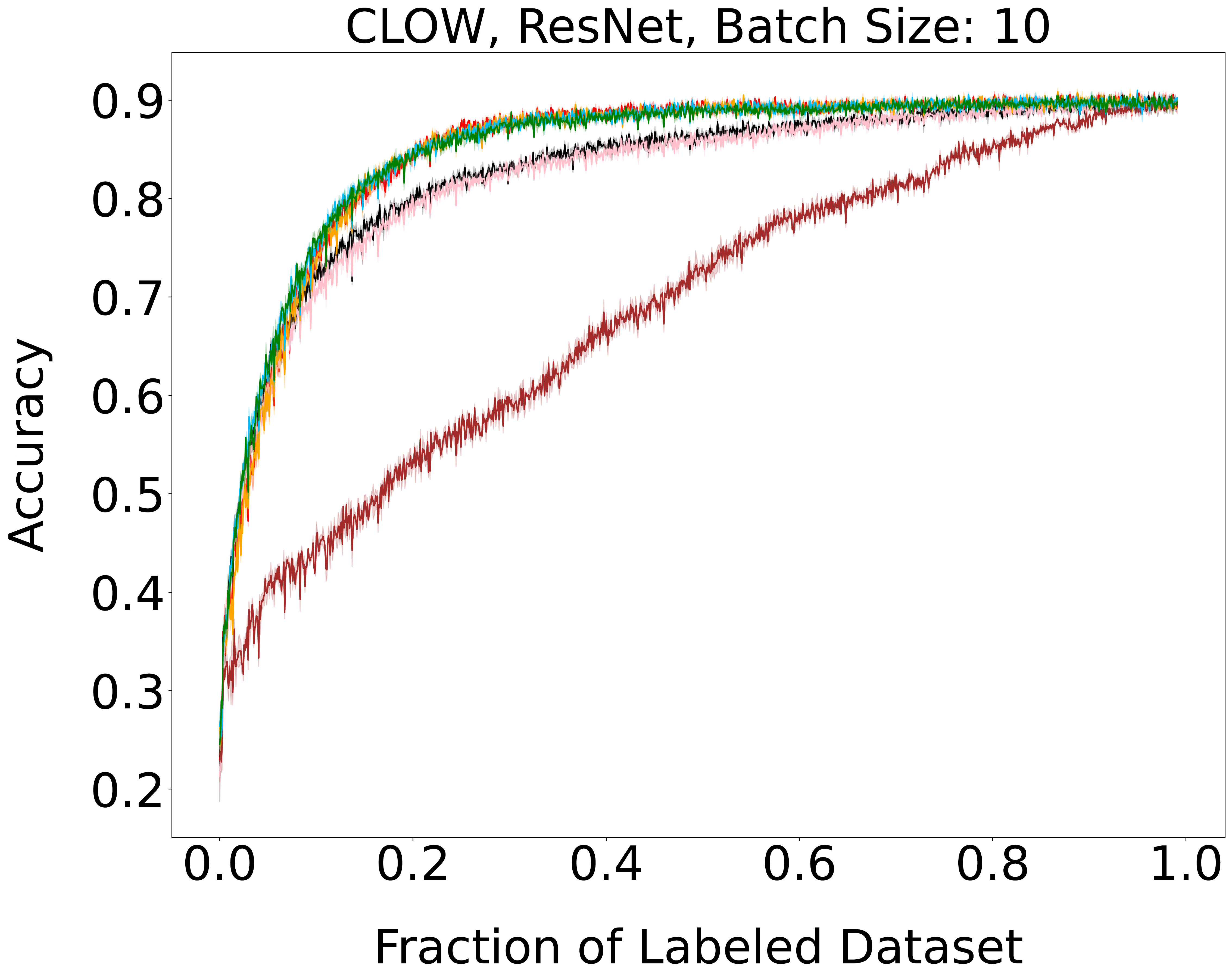}
\includegraphics[width=0.3\textwidth]{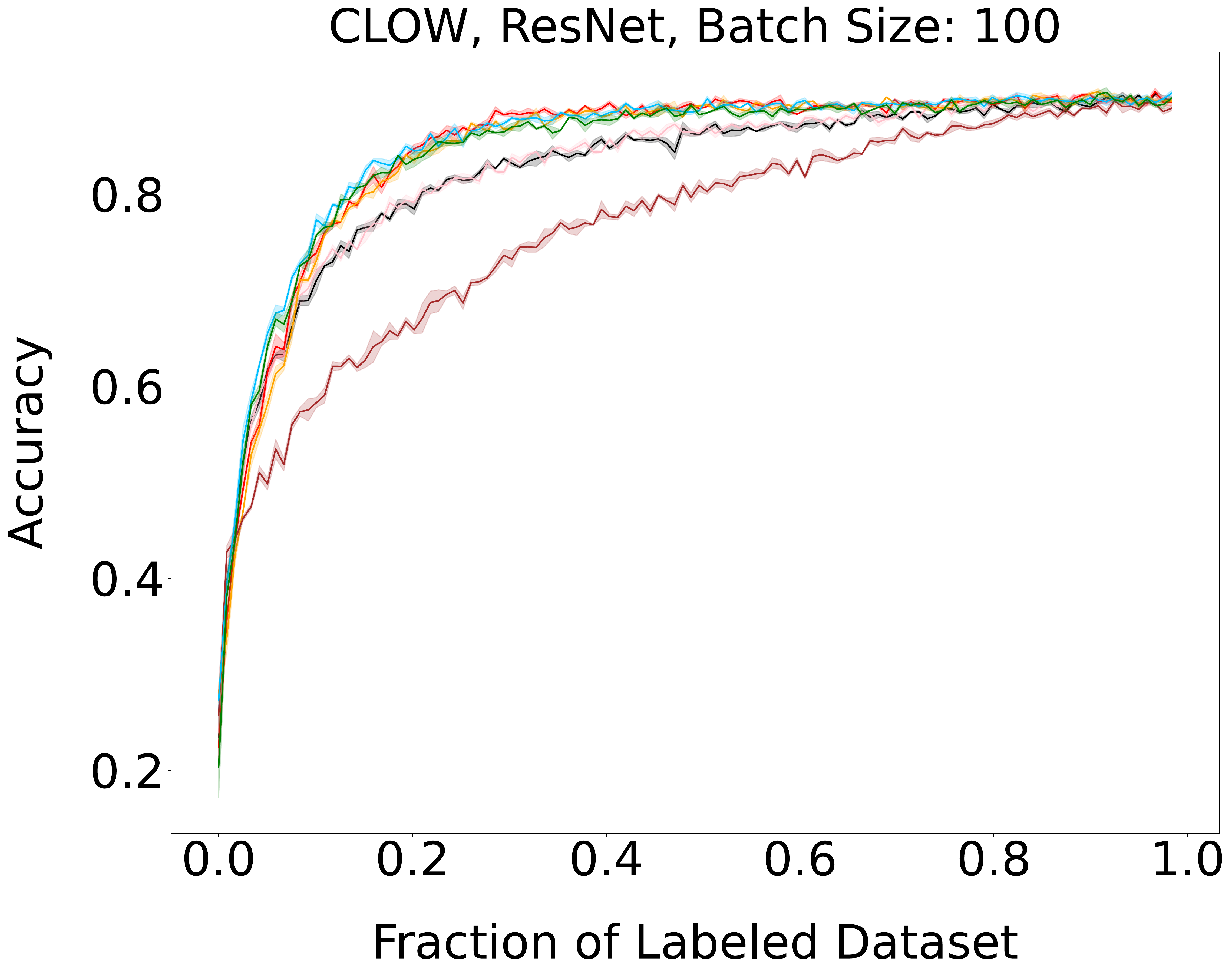}
    }

    \caption{Learning curves for the \texttt{CLOW} dataset with two network architectures and two batch sizes. Streaming active learning algorithms observe object images ordered by timestamps at which users interacted with and provided a label for the corresponding object.}
    \label{fig:hololens_sorted}
\end{figure*}

We show learning curves for all experiments with non-i.i.d. data streams from the following datasets: \texttt{SVHN} (Fig.~\ref{fig:svhn_1000_sorted}), \texttt{CIFAR-10} (Fig.~\ref{fig:cifar_1000_sorted}), \texttt{CLOW} (Fig.~\ref{fig:hololens_sorted})).

\begin{figure*}
    \centering
    \includegraphics[width=0.44\textwidth]{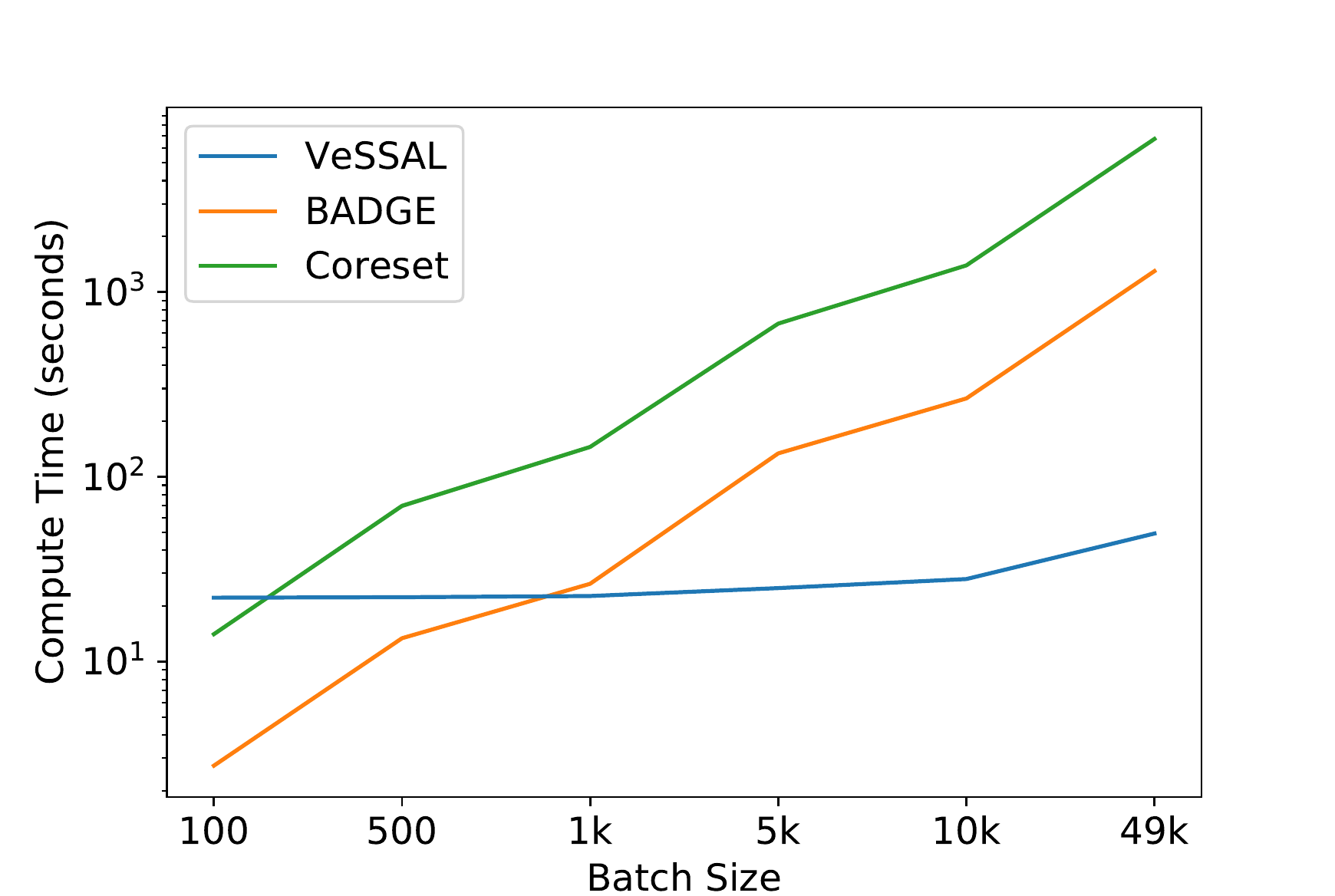}
    \caption{A y-axis logged version of Figure~\ref{fig:compute_plot}, showing compute time required to select a batch for three different algorithms on CIFAR-10 as a function of the query batch size. While non-streaming algorithms require compute that grows superlinearly as a function of labeling budget, VeSSAL stays relatively constant. Results are averaged over five replicates and each algorithm was given identical computational resources.}
    \label{fig:compute_plot_log}
\end{figure*}

\begin{figure*}
    \centering
    \includegraphics[width=0.94\textwidth]{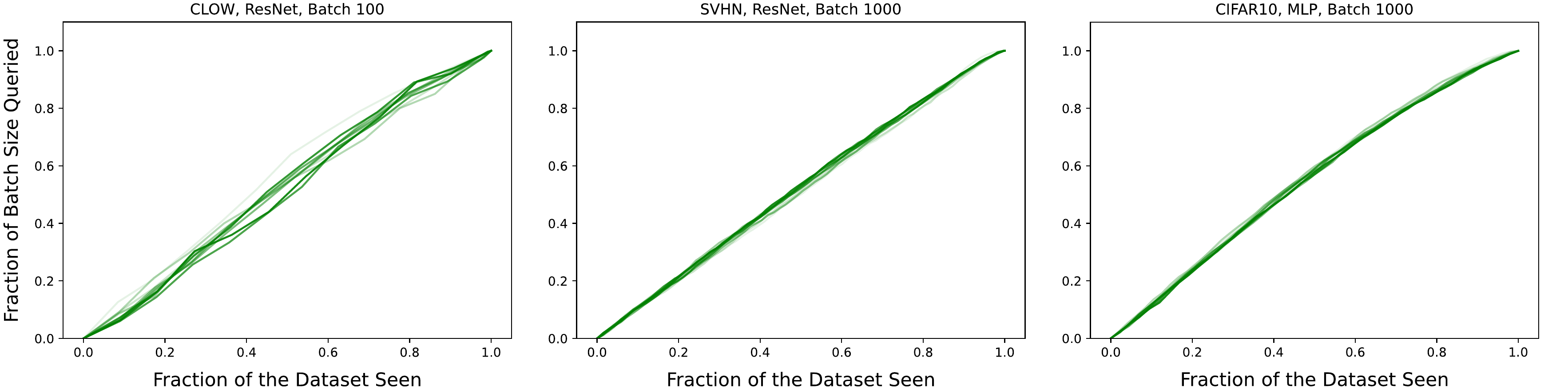}
    \caption{VeSSAL distributes its labeling budget equitably across different non-I.I.D. data streams. Here we show several different datasets, acquisition batch sizes, and model architectures, and all datasets (excluding \texttt{CLOW}, which is already non-I.I.D.) are sorted by their first principal component.}
\end{figure*}

\end{document}